# Reliability of Claude Code as an Agentic AI Grading System for Handwritten University Exams

**Rayed AlGhamdi**
Department of Information Technology
Faculty of Computing and Information Technology
King Abdulaziz University, Jeddah, Saudi Arabia
P.O.Box 80221 Jeddah 21589 KSA
raalghamdi8@kau.edu.sa
ORCID ID: 0000-0002-6277-2124

## Abstract

This study addressed four research questions about the reliability of Claude Code, an agentic deployment of the Claude Opus 4.6 large language model, as a grading system for handwritten university exam papers. RQ1 asked to what extent AI-assigned grades agree with expert human grades at the overall score level. RQ2 asked how that agreement varies across five question types, ranging from multiple choice through evaluative judgement. RQ3 asked how consistent AI grading is across two independent grading sessions, one week apart. RQ4 asked how accurate the AI-generated feedback content is.

Sixty consenting students from a Saudi Arabian computer graphics course had their 20-mark mid-term exam papers graded by Claude Code in a single agentic grading session, with three structured rubric files providing the marking scheme, the grading rules, and the output format. Reliability was estimated on disjoint samples: a 20-paper agreement sample for RQ1, RQ2, and RQ4, and a 15-paper consistency sample for RQ3. Statistical procedures included Pearson correlation, mean absolute error, Bland-Altman analysis, weighted Cohen's kappa, intraclass correlation, and a four-category content coding scheme that distinguished diagnostic feedback from confirmation feedback. The author performed human grading as the sole subject-matter expert for the course; the design therefore evaluates within-rater benchmark reliability against a domain-expert reference, not inter-rater reliability across multiple human graders.

AI grading achieved strong agreement with human grading on total scores ($r = 0.97$; MAE = 0.85 marks) with a small but statistically detectable AI-positive bias of 0.68 marks, excellent test-retest consistency (ICC = 0.96), and high feedback accuracy (92.8% diagnostic on lost mark items; 97.8% confirmation on full credit items). Drawing items were the clear outlier on three of four reliability measures. Three named failure modes recurred across multiple papers: keyword matching without conceptual depth, selection state ambiguity on multiple-choice corrections, and region of attention drift on hand-drawn diagrams.

The findings suggest that AI grading at this level of capability is ready to assist instructors on objective and short-form items under structural safeguards but is not yet ready to replace instructor judgment at boundary cases or on drawing items. The combination of conditions tested, namely the agentic deployment configuration, the handwritten Arabic-English mixed-format input, and the dual function feedback accuracy scheme, has not previously appeared in the peer-reviewed reliability literature.

# 1. Introduction

The rise of large language models (LLMs) has sparked serious interest in using artificial intelligence (AI) as an assistant for grading student work. Recent meta-analyses have begun to map the empirical evidence on AI in education broadly, including ChatGPT's effects on learning outcomes (Deng et al., 2025). Teachers spend substantial time marking exams and writing feedback. In a course with sixty to one hundred students, grading a mid-term or final exam can take several days, during which the instructor must hold back other academic responsibilities. If AI can produce a reliable first pass at grading, that time can be redirected to teaching and research. Yet the central question for any educator considering this shift is whether the AI grade is actually trustworthy. A paper about reliability must first answer: reliable in what sense, and under what conditions? This paper addresses both questions in the context of a real university exam graded by Claude Code, Anthropic's agentic coding tool, on handwritten answers written in a mix of Arabic and English by students at King Abdulaziz University.

Reliability in measurement theory has two distinct components. The first is *agreement* between the AI and an expert human grader on the same papers: do the two graders assign similar scores? The second is *consistency* across independent runs of the AI on the same papers: if the same paper is graded twice by the AI in separate sessions, do the scores match? Both properties must hold for an AI grading system to be considered reliable. An agreement without consistency means the AI sometimes matches the human and sometimes does not, depending on the run. Consistency without agreement means the AI reproduces its own mistakes perfectly. Both are necessary; neither alone is sufficient. Reliability is itself necessary but not sufficient for the validity of AI grading: an AI system that grades reliably may still fail to measure the construct the rubric specifies (construct validity) or may produce consequences for student learning that the reliability data cannot evaluate (consequential validity). The present paper is bounded to the reliability question; validity questions are addressed in the companion papers described in Section 6.

The paper addresses four research questions. RQ1 asks to what extent AI-assigned grades agree with expert human grades at the overall score level. RQ2 asks how that agreement varies across five question types ranging from multiple choice through short written answer, drawing interpretation, analytical reasoning, and evaluative judgement. RQ3 asks how consistent AI grading is when the same papers are graded in two independent sessions, one week apart. RQ4 asks how accurate the feedback content is as evaluated by an expert reviewer. Together, these four questions form a reliability profile for the AI grading system.

The study took place in CPIT285 Computer Graphics, a 300-level undergraduate course that I teach. The course is accredited by the Accreditation Board for Engineering and Technology (ABET) and covers foundational computer graphics topics. The mid-term exam consists of seven questions distributed across four sections, totalling twenty marks. The questions span Bloom's taxonomy levels one through five and require students to recall facts, apply concepts, interpret hand-drawn output, analyse scenarios, and evaluate competing arguments. Seventy-seven students were enrolled in the course and seventy-seven attended the mid-term exam on 16 April 2026. Of the seventy-six who attended, sixty consented to AI grading, thirteen declined, and three did not submit the consent form. The sixty consenting students form the analytical sample for this paper. Within this sample, two disjoint subsets supported the reliability analyses (an agreement subset of twenty papers and a consistency subset of fifteen papers, detailed in Section 3). The author performed human grading on the agreement subset as the sole subject-matter expert for the course; the resulting analysis is best understood as within-rater benchmark reliability against a domain-expert reference rather than as inter-rater reliability against a multi-grader human consensus.

The AI grading was performed using Claude Code with the Claude Opus 4.6 model, initiated from a single user instruction in a fresh session on 22 April 2026. Rather than grading the sixty papers one by one, the system autonomously dispatched ten parallel sub-agents, each grading a batch of five to six students. Each sub-agent was an independent instance of Claude Opus 4.6 with its own context window and no cross-agent communication. The total wall-clock time for grading all sixty papers was approximately 45 minutes. This agentic deployment is distinct from the chat and direct API deployments used in all prior peer-reviewed Claude grading studies. To the best of the author's knowledge, Claude Code has not been previously studied for grading in the published peer-reviewed literature.

The study contributes to four simultaneous dimensions that no prior published study has combined. First, it evaluates Claude Code in an agentic coding deployment, not chat or direct API. Second, it grades handwritten responses with mixed question types, including MCQ, short written answer, drawing interpretation, and evaluative judgement. Third, the student responses are in a mix of Arabic and English, with Arabic handwriting and right-to-left script presenting challenges that prior grading studies have not addressed. Fourth, the grading is applied to a real university mid-term exam where the grades count toward course outcomes, not to an experimental vignette or low-stakes assignment. Anthropic's Pratham International case study (Anthropic, 2026) describes a commercial Claude-powered handwritten grading deployment in India, but that deployment has not been published in peer-reviewed literature and does not report the reliability measures that the present study provides.

The remainder of the paper is organized as follows. Section 2 reviews the literature on LLM grading, covering agreement and consistency across question types and response modalities, and positioning the present study relative to existing work on Claude, GPT, and other LLM families. Section 3 describes the methodology, including the exam instrument, the Claude Code grading protocol, the blind human grading protocol, the test-retest consistency design, the analytical procedures, and the ethical framework. Section 4 presents the results addressing all four research questions. Section 5 discusses the findings against the literature and the limitations of the study. Section 6 concludes.

## 2. Literature Review

The literature on LLM grading has grown quickly in the past two years. Early studies focused on whether GPT or Claude could produce grades that correlated with human graders on typed essay responses. More recent work has extended the investigation to handwritten responses, to non-English languages, to multi-part questions with partial credit, and to feedback quality in addition to score accuracy. This section organises the existing peer-reviewed evidence under four themes: the reliability question itself, how agreement varies with conditions, how modality and time affect consistency, and how feedback content accuracy is measured.

### 2.1 The reliability question

Emirtekin (2025) published a systematic review of forty-nine peer-reviewed studies on LLM-powered automated assessment up to mid-2025. The review reports quadratic weighted kappa values ranging from 0.45 to 0.99 and correlations with human graders ranging from 0.38 to 0.95 across the included studies. This spread indicates that LLM grading reliability varies systematically with the LLM family, the prompt design, the response modality, the rubric specificity, and the cultural-linguistic context of the study. Emirtekin observes that reliability varies systematically with these conditions, and recommends that future studies report reliability profiles rather than single summary statistics. The present study follows this recommendation by reporting separate reliability measures for agreement, consistency, and feedback accuracy, with each measure broken down by question type where relevant.

Reliability in measurement theory comprises two distinct properties. *Agreement* measures whether the AI and a human expert assign similar scores to the same papers. *Consistency* measures whether the AI reproduces its own scores when regrading the same papers in a separate session. Shyr et al. (2025) articulated this distinction clearly in the context of LLM diagnostic reasoning on medical examination questions, adapting the classical repeatability-reproducibility framework from regulatory measurement science. The framework is general enough to apply beyond the medical-reasoning context, including to LLM grading reliability. Shyr et al. further observed that accuracy against a human benchmark and self-consistency across runs are conceptually distinct properties and can diverge empirically. A model can be consistently wrong in the same way across runs, or can agree with humans on average while showing run-to-run instability. The present study measures both properties on disjoint samples of papers to avoid conflating them.

## 2.2 How agreement varies

Flodén (2025) provides the anchor comparator for the present study. Flodén had ChatGPT 3.5 grade 463 Master's-level open-response answers three times each (n = 1,389 gradings) against expert human graders. The deployment used the chat interface, with a fresh chat opened for each grading. Across these gradings, 70% fell within ±10% of the teacher's score and 31% within ±5%. On categorical letter-grade matching, Cohen's κ was 0.139 (slight agreement) and the weighted κ was 0.41 (borderline fair-to-moderate). The dominant failure mode was ChatGPT accepting off-topic student answers that addressed a related but different question, a failure mode that expert humans did not exhibit. This is directly relevant to the CPIT285 mid-term exam, which contains one question (Q7) requiring students to adjudicate between two competing explanations without conflating them with a related but different concept. The pattern suggests Q7 will be one of the harder items for the AI grader and warrants close examination.

Zhang et al. (2024) tested ChatGPT 3.5, ChatGPT 4.0, and Claude 3 Haiku on 31 IELTS academic writing samples using a three-stage incremental prompt design. The authors measured interrater absolute agreement and interrater consistency against IELTS examiners for all three models, and intrarater (test-retest) reliability across fresh chat sessions for ChatGPT 3.5. With the most detailed prompt, ChatGPT 3.5 achieved interrater absolute agreement of ICC = 0.61 [95% CI 0.02, 0.85], interrater consistency of ICC = 0.77 [0.57, 0.88], and test-retest reliability of ICC = 0.89 [0.76, 0.94]. Claude 3 Haiku, with the same prompt, showed interrater absolute agreement of ICC = 0.68 [0.44, 0.83] and interrater consistency of ICC = 0.72 [0.50, 0.86]. The authors also reported that ChatGPT 3.5 graded significantly lower than IELTS examiners (Cohen's d = 0.71). Within the literature reviewed here, this is the earliest peer-reviewed work to evaluate Claude on rubric-based grading and to use a fresh-chat-session protocol to prevent chat-history contamination. The present study extends the fresh-session protocol from typed IELTS essays to handwritten mixed-format exam responses, and from ChatGPT to Claude Opus 4.6 in an agentic deployment.

Xue, Xiao, Liu, and Wilson (2026) tested Claude 3.7 Sonnet alongside GPT-4o, Gemini 2.0, DeepSeek-V3, and Qwen-3 on constructed-response items. The authors sampled each LLM's output five times per response at two temperatures and measured intra-model consistency using Fleiss's kappa. Claude 3.7 Sonnet showed a Fleiss kappa of 0.985 at temperature 0 and 0.750 at temperature 1. The authors also report that intra-consistency does not correlate with accuracy against human graders. A model can be highly consistent with itself and still be inaccurate. This finding underwrites the decision in the present study to estimate agreement and consistency on disjoint samples.

Three further peer-reviewed studies have evaluated Claude models against alternative LLMs and human raters. Wang et al. (2025) compared two Claude models (Claude-3.5-Haiku and Claude-3.5-Sonnet) against two OpenAI models (GPT-4o-mini and GPT-4o) and human raters using generalizability theory and many-facet Rasch modelling on 4,315 essays drawn from three Automated Student Assessment Prize (ASAP) essay sets totalling 1,787, 1,805, and 723 samples, respectively. The authors reported that LLMs aligned with human raters in ranking tasks but diverged in absolute scoring: humans were generally more lenient than LLMs (i.e., LLMs assigned lower scores than humans), with GPT models the most stringent of the LLMs tested. Claude models exhibited superior scoring stability compared to GPT models, approaching perfect reliability in key-point scoring. Başaran (2026) tested several generative LLMs (including Claude and a locally fine-tuned Gemma 3 model) on a German as a Foreign Language midterm exam at Anadolu University, finding Claude Opus 4 to be the best-performing AI and the fine-tuned Gemma 3 to perform at a low-to-mid range relative to human evaluators. Ibáñez et al. (2025) tested four multimodal LLMs (ChatGPT-4, Gemini, Amazon AI, Claude 3.5 Sonnet) on 34 students' UML class diagrams and reported moderate agreement with three expert instructors.

Across these Claude-specific studies, the deployment condition is a direct API or chat interface, and the response modality is typed text or image-based diagrams rather than handwritten mixed-format responses. To the best of the author's knowledge, no peer-reviewed study has tested a Claude model in an agentic coding deployment for grading handwritten exam papers. The present study fills that specific configuration gap.

Comparable reliability evidence from non-Claude LLMs and other languages further contextualises the present study. Grévisse (2024) tested GPT-4 and Gemini 1.0 Pro on 2,288 medical student answers across twelve courses in three languages, finding moderate agreement between the LLM and expert raters. Morjaria et al. (2024) tested ChatGPT-4 on 60 short-answer formative assessment responses to clinical vignettes in an undergraduate medical program. Across four prompting conditions, Spearman correlations with human graders ranged from 0.60 to 0.73 (all $p < 0.001$). A question-specific rubric eliminated a systematic positive bias: ChatGPT scored significantly higher than human graders without one (mean differences of 0.37 and 0.47 on a 5-point scale), but the difference disappeared with one. Disagreements of one or more points occurred in 65–80% of paired comparisons; disagreements of two or more points were less frequent at 20–38%. Yiğiter and Boduroğlu (2025) tested ChatGPT-4o on handwritten Turkish undergraduate responses to open-ended assessment items, comparing AI scoring with and without standardised criteria, with low agreement in the no-criteria condition and high agreement in the standardised-criteria condition. Pack et al. (2024) tested PaLM 2, Claude 2, GPT-3.5, and GPT-4 on 119 English language learner essays assessed twice on separate occasions and found GPT-4 to have the best intra-rater reliability. These studies collectively establish that LLM grading reliability depends on the LLM, the specificity of the rubric, the language, and the level of specialisation in the content domain.

## 2.3 Modality and time

Handwritten responses are harder than typed responses for any LLM. The peer-reviewed evidence base on handwritten LLM grading consists of a small but growing set of studies that are directly comparable to the present work.

Perš et al. (2026) graded handwritten engineering quiz responses that included hand-drawn circuit schematics in Slovenian, using a multi-stage pipeline with state-of-the-art multimodal LLM backends (GPT-5.2 and Gemini-3 Pro). The pipeline achieved approximately an 8-point mean absolute difference to lecturer grades on a 100-point scale, with low bias and an estimated manual-review trigger rate of approximately 17%. Two related studies from ETH Zürich extend this evidence base to STEM exam grading. Kortemeyer, Nöhl, and Onishchuk (2024) graded 252 high-stakes thermodynamics exam papers using GPT-4-32k for grading and GPT-4V plus MathPix for OCR transcription, finding that grading hand-drawn diagrams was less reliable than grading mathematical derivations and that the model's bookkeeping capacity limited fine-grained rubric application to no more than a handful of items at a time. The authors also reported that AI grading showed high precision ($\approx 0.95$) for identifying passing students but low recall ($\approx 0.50$), indicating that human review remains necessary for failing-grade decisions. Kortemeyer, Caspar, and Horica (2025) extended this work to GPT-5 on the open-ended portions of a year-long calculus exam at ETH Zürich (matched analytical dataset of 342 students, from 349 scans), with topics in linear algebra, differential equations, polar-coordinate integration, and Green's theorem. The authors introduced an Item Response Theory-based human-in-the-loop filter that combines a partial-credit threshold and a risk-based threshold derived from a two-parameter logistic model. Unfiltered AI-TA agreement was moderate ($R^2 = 0.85$). Under conservative confidence thresholds, agreement rose to $R^2 = 0.95$ but at the cost of leaving approximately 70% of items to human grading.

Başaran (2026) tested eight grading configurations: five general-purpose LLMs (including Claude Opus 4) accessed via API, one locally fine-tuned Gemma 3 27B Instruct, and two human grading methods; on a scanned multiple-choice German exam (n = 10 papers, 10 MCQ items each), reporting low-to-mid-range performance for the fine-tuned Gemma 3 relative to a four-evaluator human consensus. Caraeni et al. (2024) tested GPT-4o on 18 handwritten probability theory responses using a single-stage direct multimodal grading approach, and reported that even in the best condition, the AI grade deviated by 7.66% on average. Liu et al. (2024) tested GPT-4 on 54 mock-exam responses from a university-level mathematics course at ETH Zürich, using a multi-stage pipeline that combined optical character recognition, LaTeX transcription, and grading; the authors also explored confidence estimation by sampling multiple grading runs.

Together, these six studies (Perš et al., 2026; Kortemeyer, Nöhl & Onishchuk, 2024; Kortemeyer, Caspar & Horica, 2025; Başaran, 2026; Caraeni et al., 2024; Liu et al., 2024) establish that handwritten grading is possible but difficult, that diagram interpretation degrades reliability, and that the choice of LLM matters considerably. This evidence base supports two predictions for the present study: drawing items (Q4) and diagram-interpretation items will show measurably weaker reliability than text-only items, and the choice of Claude Opus 4.6 over a smaller or local model will matter for the handwritten modality.

For languages outside English, Arabic handwriting poses unique challenges for any current OCR system. Bhatia et al. (2024) reported a word error rate of 0.80% on handwriting recognition with the Qalam Arabic foundation model, built on a SwinV2 encoder and RoBERTa decoder architecture. The QARI-OCR system, built on Qwen2-VL, reports a word error rate of 16.0% and a character error rate of 6.1% (0.061) on diacritically-rich Arabic texts (Elsafty et al., 2025). These benchmarks indicate that Arabic OCR is technically readable to current multimodal systems on within-task evaluations, though the two studies were tested on different datasets under different conditions and do not constitute a direct head-to-head comparison. Neither study applied the recognition technology to grading; the present study extends the demonstration from OCR to actual grading of Arabic-English mixed responses.

Multilingual grading evidence comes from European studies. Grévisse (2024) graded 2,288 short-answer responses from 12 undergraduate medical courses at the University of Luxembourg using GPT-4 and Gemini 1.0 Pro. The study is the largest published multilingual investigation of LLM grading to date, with answers in French 62%, English 37%, and German 1%. Grévisse found a weak correlation between grade and answer language ($\eta^2 = 0.02$–$0.04$ across graders) and between grade and answer length, suggesting that language alone is not a primary driver of LLM grading reliability across three closely related Western European languages. GPT-4 graded significantly lower than the human evaluator overall (mean 0.65 versus 0.68, $p < 0.001$), while Gemini was not significantly different from the human grader; this indicates that bias direction can vary substantially even within the same study and rubric.

Test-retest consistency has been examined in four peer-reviewed LLM grading studies, each using a different temporal design. Pack et al. (2024) graded 119 English language placement test essays twice on two separate occasions with a minimum interim of 90 days (mean = 134.3 days, SD = 29.3) using PaLM 2, Claude 2, GPT-3.5, and GPT-4; this long interval was designed to detect drift across substantial time gaps. Zhang et al. (2024) used fresh chat sessions to grade 31 IELTS essays and reported test-retest reliability of ICC = 0.89 for ChatGPT 3.5; the authors did not separately compute test-retest reliability for ChatGPT 4.0 or Claude 3 Haiku. Xue et al. (2026) used a five-trial sampling design at two temperatures with Claude 3.7 Sonnet, reporting near-perfect intra-model consistency at temperature zero (Fleiss $\kappa = 0.99$) that decayed at higher temperatures. Flodén (2025) used an immediate-succession repeated grading design on ChatGPT 3.5 and observed 40-45% letter-grade drift across successive runs, which is the worst consistency result in the peer-reviewed literature and reflects the confounding effect of running the same prompt multiple times in the same chat history. The present study uses a one-week interval between sessions in fresh Claude Code sessions, which is shorter than Pack's drift-detection design and longer than Flodén's immediate-succession design. The one-week interval is designed to isolate within-version consistency: long enough to avoid chat-history contamination (following Zhang's session-independence principle) but short enough that model-update drift is unlikely to confound the test-retest result (in contrast to Pack's longer interval, where model versions may have changed silently). The empirical measurement of consistency is itself necessary because configuration-level controls are insufficient: Blackwell et al. (2024) showed that even temperature zero with a fixed random seed does not guarantee determinism in LLM output, due to non-deterministic behaviour in attention kernels and sampling.

Figure 1 positions the present study against these prior works. The horizontal axis shows LLM deployment condition (chat interface, direct API, agentic coding environment). The vertical axis shows response modality complexity (typed text, handwritten single-format, handwritten mixed-format). The fifteen existing studies cluster in the bottom-left and top-left quadrants, with none in the top-right. The present study occupies the empty top-right quadrant, which reflects the four-dimensional gap the study fills: agentic deployment, handwritten mixed-format responses, Arabic-English bilingual script, and real university stakes.

To the best of the author's knowledge, no peer-reviewed study has combined the four conditions of the present work: Claude in an agentic coding deployment, handwritten mixed-format responses including drawings, Arabic-English bilingual script, and real university stakes. A commercial precedent does exist: Anthropic's Pratham International case study (Anthropic, 2026) describes a Claude deployment for handwritten grading in India, where iterative prompt engineering against expert-reviewed benchmarks reportedly raised grading accuracy from approximately 30% to 80%. The case study is not peer-reviewed and does not report the reliability measures (correlation, κ, ICC, Bland-Altman analysis) required for methodological evaluation.

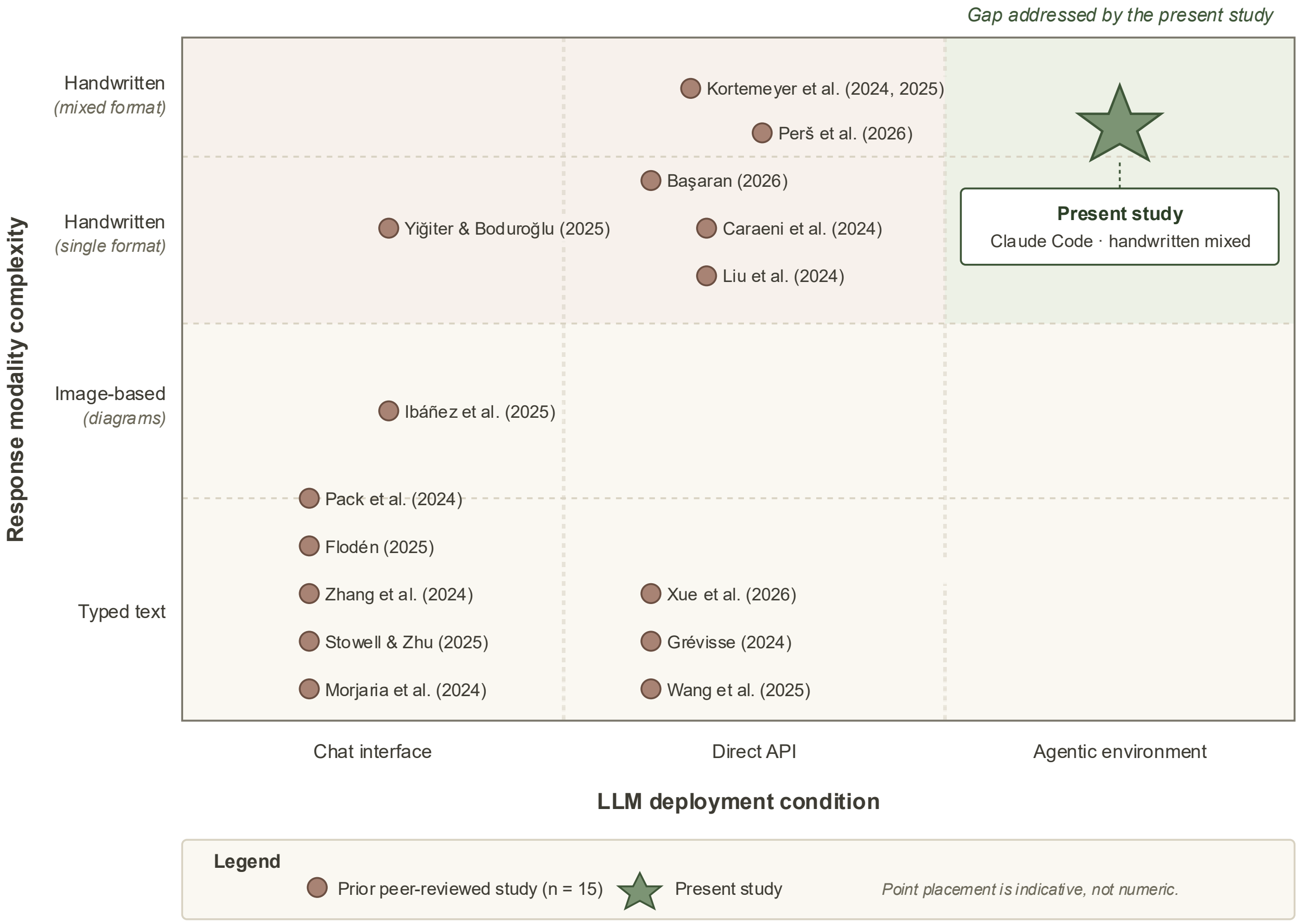

Figure 1: Peer-reviewed LLM grading studies by deployment condition and response modality

## 2.4 Feedback content accuracy

Score accuracy is only half of what makes a grade useful. The feedback that accompanies the score must also be accurate. If a student reads their grade report and the feedback is substantively wrong about what they did or did not understand, the feedback is worse than useless. It misleads.

Research on feedback content accuracy in LLM grading is recent. Scarlatos et al. (2024) proposed a three-category rubric for AI-generated feedback covering correctness, diagnostic value, and positivity. Seo et al. (2025) used a five-criterion framework to verify LLM evaluator consistency and accuracy in education. Both frameworks share a correctness dimension, which asks whether the feedback accurately identifies what the student got right or wrong and whether the explanation is substantively correct. The present study adapts this shared correctness dimension into a four-category coding scheme for expert review of the AI feedback content.

Other work has focused on feedback usefulness rather than accuracy. Pathak et al. (2025) showed that question-specific rubrics produce more useful LLM feedback on programming code. Mahinpei et al. (2026) compared LLM-generated feedback against teaching assistant feedback on proof-based computer science assignments and found that LLM feedback helps in some cases and hurts in others. Er et al. (2025) surveyed students who had received both AI and instructor feedback and found that students perceived instructor feedback as more useful overall. Askari (2025) implemented a generative AI-supported rubric model in a postgraduate business economics course in which ChatGPT-generated feedback was refined by instructors before delivery to students, and reported internal consistency, fairness, and pedagogical alignment of the resulting feedback. These studies address usefulness as perceived by students, the workflow of instructor-refined AI feedback, and related practical concerns, which are separate constructs from the factual accuracy of the feedback content. Factual accuracy is measured by an expert reviewer who compares the feedback against the ground truth of the marking scheme and the student's actual answer.

In prior work (AlGhamdi, 2024), I studied 111 computing students at King Abdulaziz University who received blinded ChatGPT-generated feedback on technical writing assignments. That study measured student perception of the feedback quality and found mixed reception. The present study extends this line of work by measuring the factual accuracy of the feedback content rather than student perception, and by using Claude Opus 4.6 in an agentic deployment rather than ChatGPT through the chat interface.

# 3. Methodology

## 3.1 Study context and participants

The study took place in CPIT285 Computer Graphics, a 300-level undergraduate course I teach in the Information Technology Department at the Faculty of Computing and Information Technology, King Abdulaziz University, during the Spring 2026 semester. Seventy-seven students were enrolled in the course, and seventy-six attended the mid-term examination on 16 April 2026. Table 1 summarises the four participant groups defined by consent decision and exam attendance.

Table 1. Study participants by consent and attendance status

| Group | Consent decision | Exam attendance | n | Grading procedure |
|---|---|---|---|---|
| A | Consented | Attended | 60 | Claude Code (AI) |
| B | Declined | Attended | 13 | Author (manual) |
| C | Did not submit the form | Attended | 3 | Author (manual) |
| D | Did not submit the form | Absent | 1 | Not graded |
| **Total** | | | **77** | |

Consent was obtained through a bilingual Arabic and English online form distributed approximately two weeks before the examination and closed immediately before the exam began. The consent form described the grading procedure, noted that Claude Code would grade the consenting students' papers, and guaranteed each consenting student the right to request a full human re-evaluation of their grade at any point during or after the course without justification or penalty. A reassurance panel preceded the consent decision itself, specifically addressing concerns about being locked into an AI grade. See *Appendix D* for the consent form. Students who declined consent and students who did not submit a consent form were treated identically, consistent with the principle that absence of explicit consent cannot be interpreted as consent. Only Group A, the consenting cohort of sixty students, forms the analytical sample for this paper.

Consented papers were sorted alphabetically by student identifier and assigned anonymised codes S001 through S060, which were written on each paper's cover sheet. The cover sheet bearing student's name and identification number was physically masked with an adhesive cover before scanning, and only the five answer pages were scanned into the "Exam Papers" folder, where Claude Code reads. Claude Code, therefore, received no identifying information. Human grading of the 35-paper blind sample (detailed in Section 3.4) was performed on papers bearing only the S001–S060 code; I did not link anonymised codes to student identities until after both blind human grading and Claude Code Run 1 were complete, at the point of preparing the delivery list matching each student's email address to their grade report.

## 3.2 Exam instrument

I designed the CPIT285 mid-term examination to assess both the foundational concepts and higher-order reasoning covered in the first half of the course. Figure 2 visualizes the exam structure against Bloom's taxonomy levels and Student Outcome mapping, and Table 2 specifies the structure in tabular form.

Table 2. Structure of the CPIT285 mid-term exam (20 marks total)

| Section | Questions | Marks | Question type | Bloom level | SO mapping |
|---|---|---|---|---|---|
| **A** | Q1.1–Q1.5 | 5 | Multiple choice | L1–L2 | SO1 |
| **A** | Q2 | 1 | Short written | L2 | SO1 |
| **B** | Q3 | 3 | Short written | L3 | SO2 |
| **B** | Q4 | 3 | Drawing interpretation | L3 | SO2 |
| **C** | Q5 | 2 | Analytical reasoning | L4 | SO1 |
| **C** | Q6 | 2 | Analytical reasoning | L4 | SO2 |
| **D** | Q7 | 4 | Evaluative judgement | L5 | SO1 |
| **Total** | **7 items** | **20** | **5 types** | | **SO1 = 12, SO2 = 8** |

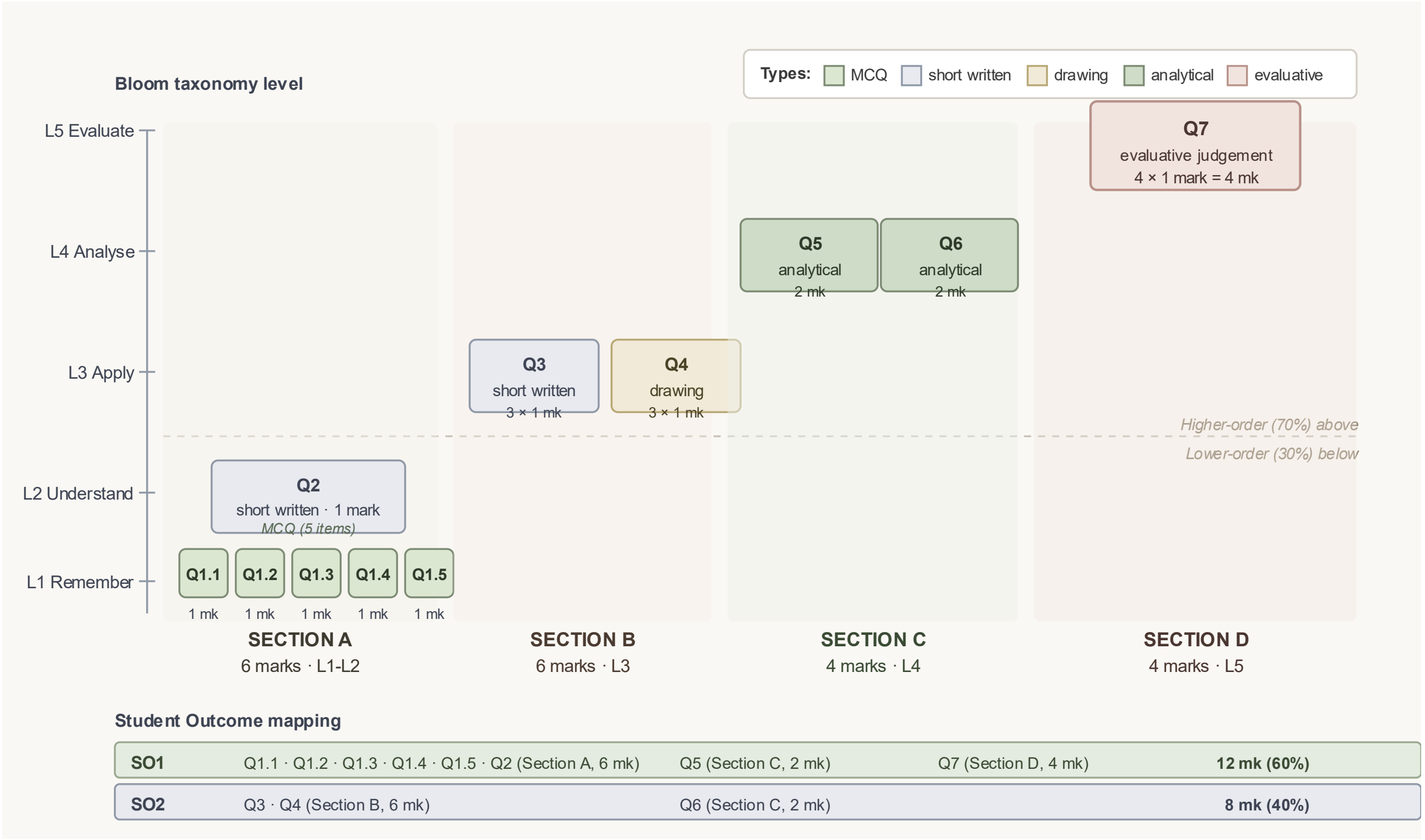

Figure 2: CPIT285 mid-term exam structure - 7 questions, 5 question types, Bloom L1-L5

The exam distributes 30% of marks to lower-order items (Section A, Bloom L1–L2) and 70% to higher-order items (Sections B, C, D; Bloom L3–L5). This distribution is appropriate for a 300-level computing course and ensures that the grading reliability analysis is not dominated by easy-to-grade multiple-choice items. Students wrote in a mix of Arabic and English. Most students used English for technical terminology and multiple-choice answers, and some switched to Arabic for extended written explanations in Q2, Q3, Q5, Q6, and Q7. The graphical elements of Q4 required students to reproduce the visual output of three different OpenGL primitive drawing modes (GL_LINES, GL_LINE_LOOP, GL_TRIANGLES) from a set of five supplied vertex coordinates, evaluated through reference-image comparison rather than textual answer matching. The complete exam instrument is reproduced in *Appendix A*, the marking scheme is provided in *Appendices B*, and Q4 reference images are available in the study's *OSF repository* (refer to the data availability statement for the access link).

### 3.3 Claude Code grading protocol

Grading was conducted using Claude Code version 2.1.117 with Claude Opus 4.6, accessed through the command-line interface on my local development environment. Claude Code is an agentic coding tool that provides the underlying model with direct file system access, execution of shell commands, and tool calls through the Model Context Protocol. Access to Claude Opus 4.6 was provided through a Claude Max subscription, which supplies elevated usage limits sufficient for parallel sub-agent execution without quota-triggered model fallback. The grading took place on 22 April 2026 on a host running Windows 10 Enterprise with Claude Code executed inside a Windows Subsystem for Linux (WSL 2) environment running Ubuntu 22.04 LTS.

The model was pinned via a project-local configuration file (`.claude/settings.json`) that restricted the available model set to claude-opus-4-6 only, preventing mid-session model switching. The effort level was set to `xhigh`, one step below the maximum on the five-level speed-intelligence scale (low, medium, high, xhigh, max), providing extended reasoning depth for multi-step partial-credit grading decisions. Auto-updates were disabled for the duration of the study to ensure identical client behaviour across grading sessions. Model identity, effort level, and client version were verified via the `/status` command at the start of the session, with outputs recorded in an environment log (`run_environment_log.md`) included in the study's *OSF repository*.

Figure 3 presents the grading pipeline architecture. The workflow was initiated from a single user instruction issued in a fresh Claude Code session:

```
Read all three files in /01_Rubric_and_AnswerKey/ completely before
proceeding, then grade all papers except S001 in /02_ExamPapers/ following
the rules and producing outputs exactly as specified in output_schema.md.
```

No further human instructions were given during the grading of individual papers. All decisions about batching, parallelisation, score extraction, and CSV compilation were made by the model based on this single initial instruction and the rules encoded in the three rubric files.

The grading workflow relied on three authored files in the rubric folder. First, `grading_rules.md` specifies the behavioural constitution of the grading system, including language-handling rules for Arabic-English mixed responses, handwriting interpretation rules, confidence thresholds, flagging criteria for ambiguous cases, and feedback quality standards. Second, `marking_scheme.md` specifies the complete answer key: correct answers for multiple-choice items, key points required for full or partial credit on each written question, award and zero-credit criteria, and the reference image comparison protocol for Q4. Third, `output_schema.md` specifies the exact output format for grade reports, flag reports, and the master CSV. All the rubric folder files are included in the study's *OSF repository*.

Student papers were organised into two versions per student. The seven-page full version (S001_full.pdf through S060_full.pdf), stored in /06_FullExamPapers/, contained all pages, including the cover sheet and assessment mapping sheet, and was used only at the final delivery step. The five-page grading version (S001.pdf through S060.pdf), stored in /02_ExamPapers/, contained only the answer pages and was the file Claude Code actually read during grading.

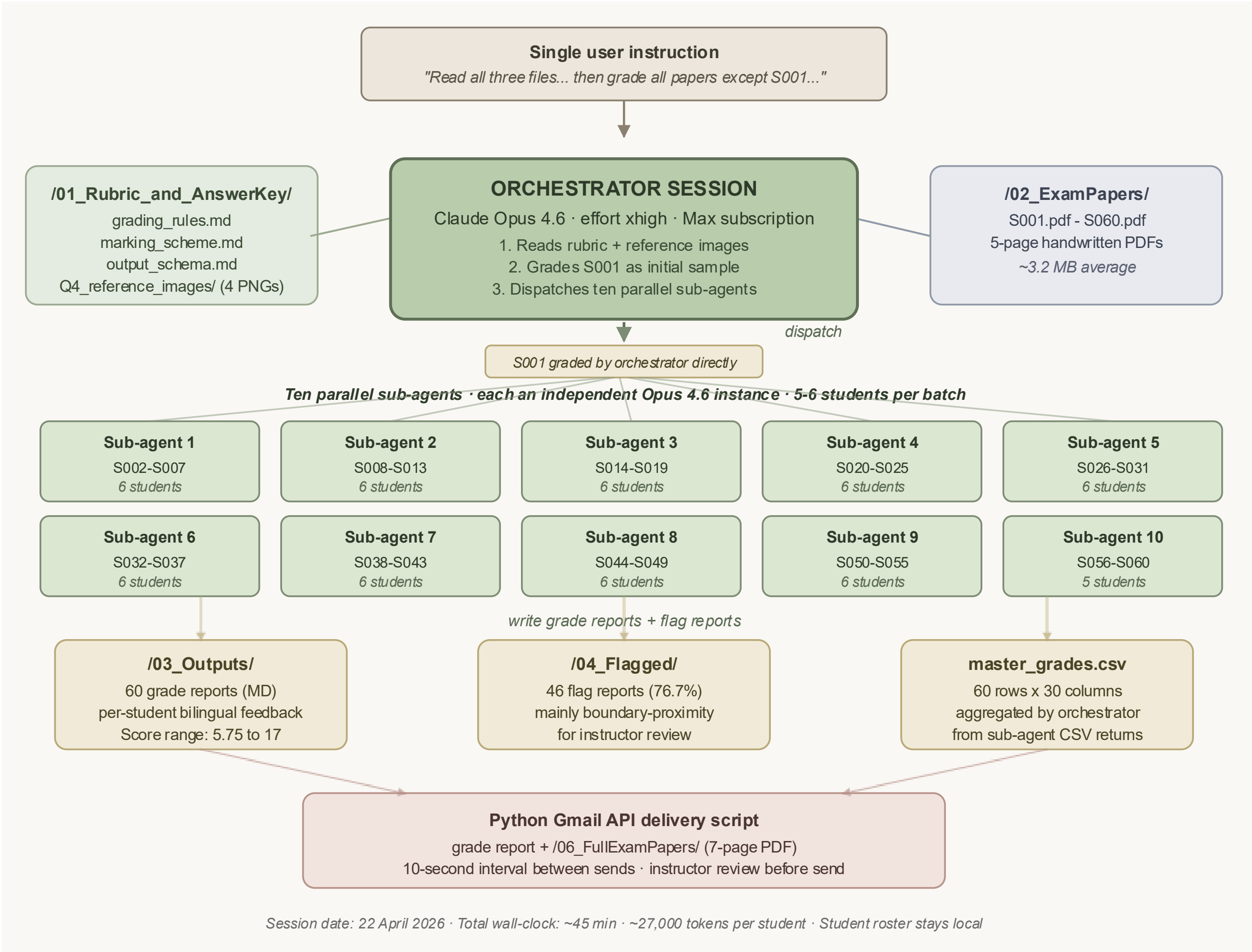

Figure 3: Claude Code grading pipeline - orchestrator with ten parallel sub-agents

The orchestrator session began by reading all rubric materials and the four Q4 reference images in full. S001 was then graded first by the orchestrator directly as an initial sample, enabling me to verify that the output format matched the specification in `output_schema.md` before the remaining fifty-nine papers were processed. No changes to any rubric file, instruction, or configuration setting were made between the S001 output and the grading of the remaining papers, so S001 was graded under conditions identical to those applied to S002 through S060.

After S001 was verified, the orchestrator distributed the remaining fifty-nine papers across ten parallel sub-agents using Claude Code's Agent tool. Each sub-agent was an independent instance of Claude Opus 4.6 with its own context window, inheriting the model configuration of the parent session. Sub-agents were assigned batches of five or six students each. Each sub-agent received a self-contained prompt specifying its role, the full paths to the three rubric files and four Q4 reference images, an explicit instruction to use the Read tool (rather than Bash-based PDF conversion) for both PDFs and reference images, the MCQ answer key embedded directly in the prompt, condensed grading criteria, output file paths, and a CSV return format specification. Each sub-agent operated in isolation, without awareness of or communication with other sub-agents, ensuring that each batch was graded from a clean context with no carry-over between students. After all sub-agents completed, the orchestrator aggregated the CSV summaries into a master CSV and verified that all sixty grade reports were present, that no score exceeded its maximum, and that all section totals were internally consistent.

The parallel sub-agent architecture reduced total wall-clock time for the sixty-paper grading to approximately 45 minutes. Token consumption was approximately 27,000 tokens per student, driven primarily by the multimodal reading of the five-page handwritten PDF and the four Q4 reference images.

Across the sixty papers, 46 grade reports (76.7%) triggered at least one flag for human review. The flag rate was driven primarily by the score-boundary-proximity rule specified in `grading_rules.md`, which flags any paper whose total falls within two marks of a grade boundary (10 for passing, 14 for B, 17 for A). The distribution of scores in the cohort clustered between 10 and 15, which placed the majority of papers within two marks of at least one boundary. The high flag rate, therefore, reflects the score distribution of the cohort rather than a high rate of AI grading uncertainty, and means that instructor review under the four structural safeguards described in Section 3.7 covered the majority of the cohort rather than an exception subset.

Markdown grade reports generated by the sub-agents were then converted to styled bilingual PDFs using a Python pipeline based on fpdf2, arabic_reshaper, and python-bidi, with per-sentence bidirectional processing applied to ensure correct rendering of mixed Arabic-English content. The PDF grade reports were delivered to students via an automated Python script interfacing with the Gmail API, with a ten-second interval between sends to comply with rate limits. Each student received two attachments: the PDF grade report and the seven-page full scan of their original exam paper.

Neither the student roster linking anonymised codes to real names nor the blind human grading spreadsheet was ever stored in the project folder or accessible to Claude Code or to any sub-agent. Both files remained local to my own device throughout the study. A complete procedural record of the grading session, including the exact sub-agent prompts, tool usage, timing data, per-student output file listing, and the end-to-end replication guide, is provided as methodology appendices (`AI_grading_methodology.md` and `Appendix_Replication_Guide.md`) in the study's *OSF repository*.

### 3.4 Blind human grading protocol

The most important methodological safeguard in this study is the blind grading protocol, which ensures independence between the AI and human grader on the comparison sample. The protocol follows three rules and an identity-blinding requirement, described in turn below:

1. Human grading is performed before any Claude Code output is inspected for the comparison-sample papers.
2. I do not view any Claude Code output for any paper before completing human grading.
3. Human and AI scores are recorded in separate spreadsheets and compared only after both sets of scores are finalised.

Human grading was performed by me as the sole subject-matter expert for CPIT285 at this institution, on papers bearing only the anonymised S001–S060 code, with student names and identification numbers physically masked throughout the blind grading window (19–20 April 2026). Anonymised codes were linked to student identities only after both human grading and Claude Code Run 1 were complete. The single-rater design is acknowledged as a methodological feature throughout: the human grade serves as a domain-expert reference standard, not a multi-grader consensus, and the agreement statistics in Section 4 should be read as within-rater benchmark reliability against that reference. Figure 4 shows the temporal ordering of grading activities and their independence guarantees.

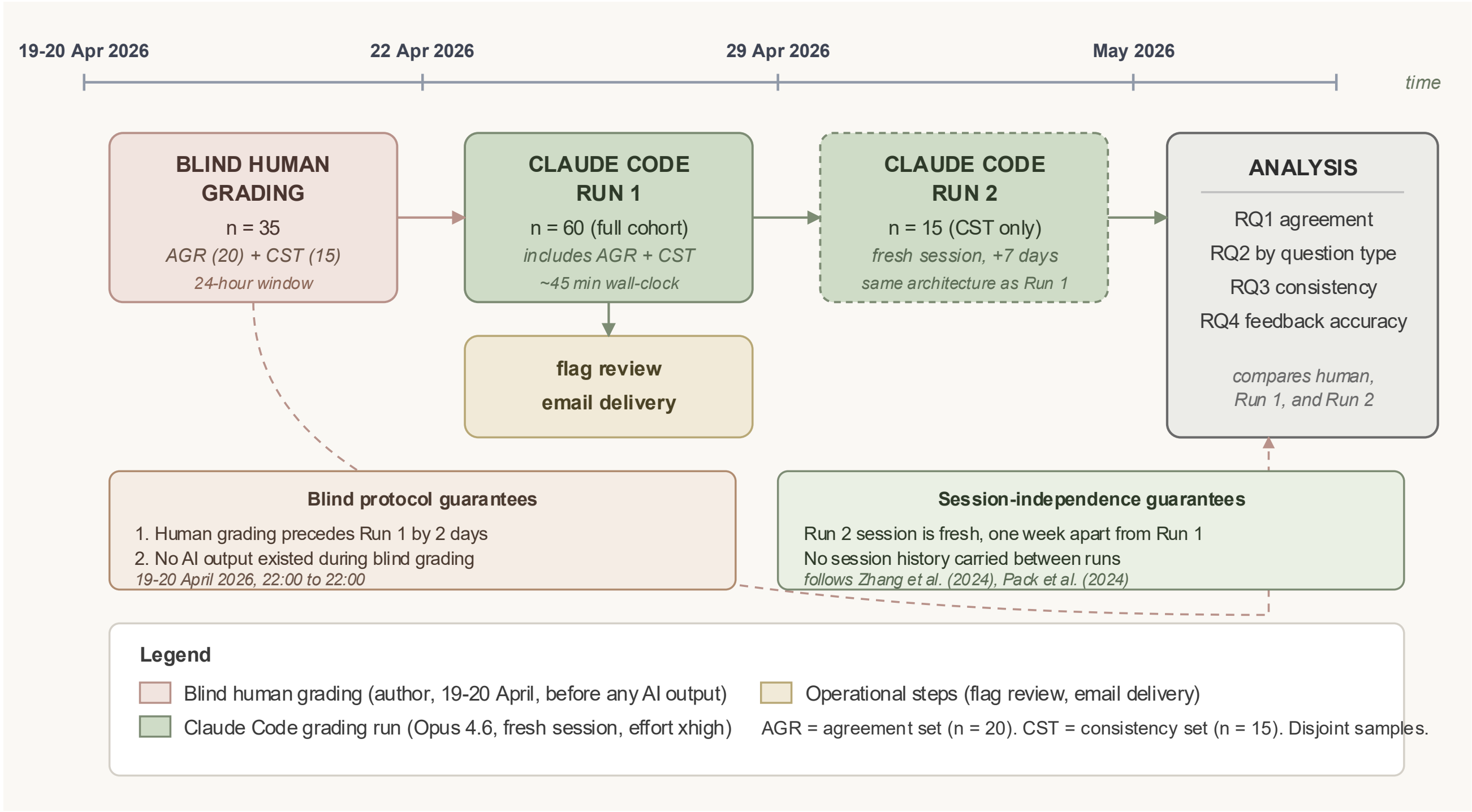

Figure 4: Temporal protocol - blind human grading precedes all AI grading runs

The agreement sample $S_{AGR}$ consists of twenty papers selected by choosing every third paper from the consenting cohort:

$$S_{AGR} = \{ S_{1+3k}: k = 0,1,2, \dots ,19 \} = \{S_{001}, S_{004}, S_{007}, \dots , S_{058}\}$$

The every-third-paper selection rule was defined before the consent form closed, eliminating any possibility of selection bias based on knowledge of which students consented or of paper content. The consistency sample $S_{CST}$ consists of fifteen papers drawn from the remaining forty consented papers:

$$S_{CST} = \{S_{002}, S_{003}, S_{005}, S_{006}, S_{008}, S_{009}, S_{011}, S_{012}, S_{014}, S_{015}, S_{017}, S_{018}, S_{020}, S_{021}, S_{023}\}$$

By construction $S_{AGR} \cap S_{CST} = \emptyset$, ensuring that agreement and consistency are estimated on disjoint samples. Blind human grading was performed for both samples ($S_{AGR} \cup S_{CST}$, n = 35). I graded all thirty-five papers in two sittings within a forty-eight-hour window to minimise drift in grading standards across the sample, using a printed copy of the same answer key that Claude Code received. Per-question scores were recorded on paper first and transferred to blind_grading_scores.xlsx afterwards. The remaining twenty-five consented papers were graded by Claude Code only.

### 3.5 Test-retest consistency protocol

The fifteen papers in S_CST were graded twice by Claude Code on two separate occasions, one week apart. Both runs used identical inputs: the same three rubric files, the same student paper files, the same start instruction, and the same Claude Code configuration pinned through .claude/settings.json. Each run was conducted in a fresh Claude Code session with the same orchestrator-plus-parallel-sub-agents architecture as Run 1, although Run 2 used fewer sub-agents because it processed only fifteen papers rather than sixty.

The two runs were independent by design. No session history from the first run was carried into the second, and no information about first-run outputs was provided to the orchestrator or to any sub-agent in Run 2. The two total scores produced for each paper, one from each run, can therefore be treated as repeated independent measurements for intraclass correlation analysis.

This session-independence design follows the methodological principle established by Zhang et al. (2024), who used fresh chat sessions to prevent chat history contamination, and extends that principle from typed IELTS essays to handwritten mixed-format exam responses. The one-week interval distinguishes the present design from both Flodén's (2025) immediate-succession repeated grading at one extreme and Pack et al.'s (2024) minimum-90-day interval (mean = 134 days) at the other. The shorter Pack-style interval risks model-update drift confounding the consistency estimate; the longer Flodén-style design within the same chat session risks chat-history contamination. The one-week interval is short enough to keep within a single Claude Opus 4.6 model version and long enough to require a fresh session, addressing both confounds simultaneously.

### 3.6 Analytical procedures

Each research question is addressed by a specific statistical procedure chosen to match the scale of the data, the independence of the graders, and the reliability property being measured. Table 3 summarises the mapping.

Table 3. Mapping of research questions to analytical procedures

| RQ | Construct | Sample | n | Statistic | Source for threshold |
|---|---|---|---|---|---|
| RQ1 | Overall agreement | $S_{AGR}$ | 20 | $r$, MAE, Bland-Altman | Bland & Altman (1999) |
| RQ2 (primary) | Agreement by type | $S_{AGR}$ | 20 | Weighted $\kappa_w$ per type (k = 5) | Viera & Garrett (2005) |
| RQ2 (secondary) | Agreement by question | $S_{AGR}$ | 20 | Weighted $\kappa_w$ per question (k = 20) | Viera & Garrett (2005) |
| RQ3 (primary) | Test-retest consistency | $S_{CST}$ | 15 | ICC(2,1), 95% CI | Koo & Li (2016) |
| RQ3 (secondary) | Consistency by type | $S_{CST}$ | 15 | ICC(2,1) per type | Koo & Li (2016) |
| RQ3 (tertiary) | Accuracy on CST papers | $S_{CST}$ | 15 | $r$, MAE vs. human | — |
| RQ4 (primary) | Diagnostic feedback accuracy | $S_{AGR}$ | 20 | C1 / (C1+C3+C4) on lost-mark items | Seo et al. (2025) |
| RQ4 (secondary) | Confirmation feedback accuracy | $S_{AGR}$ | 20 | C2 / (C2+C3+C4) on full-credit items | Seo et al. (2025) |

**RQ1. Overall agreement:** Let $X_j^H$ denote the human total score and $X_j^{AI}$ denote the Claude Code total score for paper $j \in S_{AGR}$, both on the twenty-point scale. The primary analysis reports three quantities:

$$r = \frac{\sum_{j=1}^{20}(X_j^{AI} - \bar{X}^{AI})(X_j^H - \bar{X}^H)}{\sqrt{\sum_{j=1}^{20}(X_j^{AI} - \bar{X}^{AI})^2 \sum_{j=1}^{20}(X_j^H - \bar{X}^H)^2}} \quad (1)$$

$$\text{MAE} = \frac{1}{20}\sum_{j=1}^{20} | X_j^{AI} - X_j^H | \quad (2)$$

Bland-Altman analysis plots the difference $d_j = X_j^{AI} - X_j^H$ against the mean $\bar{X}_j = (X_j^{AI} + X_j^H)/2$ for each paper, with limits of agreement calculated as:

$$\text{LoA} = \bar{d} \pm 1.96\, \hat{\sigma}_d \quad (3)$$

where $\bar{d}$ is the mean difference (interpreted as systematic bias) and $\hat{\sigma}_d$ is the standard deviation of differences.

**RQ2. Agreement by question type**: Let $Q$ denote the set of five question types: MCQ (Q1.1–Q1.5), short written (Q2, Q3), drawing (Q4), analytical (Q5, Q6), and evaluative (Q7). For each type $q \in Q$, Cohen's weighted kappa is computed on the per-question scores:

$$\kappa_w = 1 - \frac{\sum_{i=1}^{k}\sum_{j=1}^{k} w_{ij}\ p_{o,ij}}{\sum_{i=1}^{k}\sum_{j=1}^{k} w_{ij}\ p_{e,ij}} \qquad (4)$$

where $p_{o,ij}$ is the observed proportion of paired scores in cell $(i,j)$, $p_{e,ij}$ is the expected proportion under chance agreement, and $w_{ij}$ is the quadratic weight:

$$w_{ij} = \frac{(i-j)^2}{(k-1)^2}$$

Quadratic weighting penalises larger disagreements more heavily than smaller ones and is appropriate for partial-credit ordinal scales (Cohen, 1968). Interpretation follows Viera and Garrett (2005): $\kappa_w < 0.20$ slight, 0.21–0.40 fair, 0.41–0.60 moderate, 0.61–0.80 substantial, $> 0.80$ almost perfect. The secondary analysis reports $\kappa_w$ for each of the seven individual questions as a supplementary table.

**RQ3. Test-retest consistency**: The primary analysis uses the intraclass correlation coefficient of type ICC(2,1) under the Shrout and Fleiss (1979) nomenclature, also designated as the two-way random-effects absolute-agreement single-measurement model in Koo and Li (2016). For total scores $X_j^{(1)}, X_j^{(2)}$ across the two runs on paper $j \in S_{CST}$:

$$\mathrm{ICC}(2,1) = \frac{\mathrm{MS}_R - \mathrm{MS}_E}{\mathrm{MS}_R + (k-1)\,\mathrm{MS}_E + \frac{k}{n}(\mathrm{MS}_C - \mathrm{MS}_E)} \qquad (6)$$

where $\mathrm{MS}_R$ is the mean square for rows (papers), $\mathrm{MS}_C$ is the mean square for columns (runs), $\mathrm{MS}_E$ is the residual mean square, $n = 15$ is the number of papers, and $k = 2$ is the number of runs. The 95% confidence interval is reported alongside the point estimate. Interpretation follows Koo and Li (2016): ICC $<0.50$ poor, 0.50–0.75 moderate, 0.75–0.90 good, $> 0.90$ excellent.

The secondary analysis computes ICC(2,1) on a per-question-type basis to examine whether consistency varies across types in the same way that agreement does. The tertiary analysis compares each of the two Claude Code runs against the blind human grades on the fifteen consistency-test papers:

$$r_1 = \mathrm{cor}\left(X^{(1)}, X^H\right), r_2 = \mathrm{cor}\left(X^{(2)}, X^H\right) \qquad (7)$$

$$\mathrm{MAE}_1 = \frac{1}{15}\sum_{j \in S_{CST}} \mid X_j^{(1)} - X_j^H \mid, \mathrm{MAE}_2 = \frac{1}{15}\sum_{j \in S_{CST}} \mid X_j^{(2)} - X_j^H \mid \qquad (8)$$

The tertiary analysis is what Shyr et al. (2025) describe as jointly estimating repeatability (run 1 vs. run 2) and accuracy (either run vs. human), noting that repeatability and accuracy are conceptually distinct and can diverge empirically as Xue et al. (2026) demonstrated on constructed-response items.

**RQ4. Feedback accuracy.** The feedback content of every grade report produced by Claude Code on the twenty papers in $S_{AGR}$ is coded by me as the subject-matter expert. The coding scheme is adapted from educational feedback evaluation frameworks discussed by Scarlatos et al. (2024) and Seo et al. (2025). Scarlatos et al. proposed three feedback categories: correct, diagnostic, and positive. The "correct" category refers to feedback that is accurate and relevant to the question and the student's response, and "diagnostic" refers to feedback that accurately identifies errors or misconceptions in the student's answer. Seo et al. used a five-criterion framework to verify LLM evaluator consistency and accuracy in education. The present study draws specifically on the dimension common to both frameworks: whether the feedback substantively reflects what the student got right or wrong.

Each feedback statement is assigned to one of four mutually exclusive categories. *C1, accurate diagnostic*, applies when the student lost marks on the item, the AI correctly identifies what went wrong, and the explanation is substantively correct. *C2, accurate confirmation*, applies when the student earned full marks on the item and the AI's confirmation correctly characterises the student's reasoning, providing substantive elaboration of why the answer is correct rather than mere acknowledgement. *C3, partially accurate*, applies to either function (diagnostic on lost-mark items or confirmation on full-credit items) when the AI identifies the right area of concern, but the explanation contains a substantive error. *C4, inaccurate*, applies to either function when the AI identifies the wrong area, the explanation is substantively wrong, or both.

The decision flowchart for assigning a code is sequential. I first determine whether the student earned full marks on the question. If full marks were earned, the AI feedback is evaluated against C2, C3, or C4, since C1 is not valid on a full-credit item by definition. If the student lost any marks, the AI feedback is evaluated against C1, C3, or C4, since C2 is not valid on a lost-marks item by definition.

Let $f_{jq} \in \{C_1, C_2, C_3, C_4\}$denote the coded category for feedback on paper $j$, question $q$, where $X_{jq}$is the human-graded score and $M_q$is the maximum mark for question $q$. The analysis reports three complementary measures.

Diagnostic accuracy is computed on the subset of items where the student lost marks:

$$\mathrm{Acc}_{\mathrm{diag}} = \frac{\sum_{(j,q):X_{jq}<M_q} \mathbb{1}[f_{jq} = C_1]}{\sum_{(j,q):X_{jq}<M_q} 1} \qquad (9)$$

Confirmation accuracy is computed on the subset of items where the student earned full credit:

$$\mathrm{Acc}_{\mathrm{conf}} = \frac{\sum_{(j,q):X_{jq}=M_q} \mathbb{1}[f_{jq} = C_2]}{\sum_{(j,q):X_{jq}=M_q} 1} \qquad (10)$$

The overall four-code distribution is also reported across all items for transparency. Diagnostic accuracy is the primary RQ4 measure, since explaining errors is the harder cognitive task and the more deployment-relevant capability for instructors evaluating AI grading tools. Confirmation accuracy is reported as a secondary finding.

The single-coder design is a methodological limitation acknowledged in Section 5. It is defensible in this study on three grounds: I am the sole subject-matter expert for the course, no colleague in the department teaches the same CPIT285 material, and the coding judgments are constrained by established correct answers rather than by interpretive preference.

Post-grading review and grade reconciliation. Following Run 1 of Claude Code grading and the completion of blind human grading, the two sets of grades were brought together for the first time. I compared each AI grade with the corresponding human grade on the thirty-five papers in S_AGR and S_CST and re-examined per-question disagreements of 0.5 marks or more against the marking scheme. Where re-examination showed that one grade was in error, that grade was revised; all other disagreements were retained unchanged and remain in the agreement statistics. In the agreement sample, 116 of the 440 graded cells (counting the three parts of Q4 separately) differed by 0.5 marks or more, and nineteen were adjusted through four types of decision. Eight human grades were revised upward to match the AI grade on partial-credit cases (S007 Q6.1 and Q7.1, S040 Q5.1 and Q5.2, S058 Q3.1, Q3.3, Q4B and Q4C). Three human grades were revised independently of the AI grade on one drawing in which the correct shape had been drawn in the wrong plane (S034 Q4A, Q4B and Q4C). Seven AI grades were overridden where the AI had credited keyword presence without conceptual depth, misread a crossed-out MCQ selection, or misinterpreted a hand-drawn diagram (S001 Q3.1, S016 Q1.5 and Q4A, S022 Q3.1 and Q3.3, S046 Q3.1 and Q3.3). One human grade was confirmed against an AI overestimate (S007 Q4A). No grade was adjusted in the consistency sample. Every adjustment is documented in *Appendix F*. All statistics in Section 4 are computed on the reconciled grades.

### 3.7 Ethical considerations

The study was conducted in accordance with the Law of Ethics of Research on Living Creatures (Royal Decree No. M/59 of 14/09/1431 H) and its Implementing Regulations (National Committee of Bioethics, Kingdom of Saudi Arabia). Article 10.33 of the Implementing Regulations recognises research on educational tests, surveys, interviews, or public behaviour monitoring in which information is recorded without revealing participant identity and which presents no more than minimal risk. The present study falls within this category: it measures the reliability of an AI grading procedure applied to an educational assessment, all records were anonymised at the point of collection (as detailed in sections 3.1 and 3.3), and the grading protocol carried no risk of academic disadvantage beyond a conventional instructor-graded exam.

Four structural safeguards, established during protocol development and applied uniformly to all sixty consenting students, together form the ethical foundation of the design.

1. **No AI-assigned grade entered the official record before instructor review.** I examined every grade report, inspected all flag reports generated by the grading system, and spot-checked a minimum of one paper per question type before any grade was entered into the university's grade management system. Given the 76.7% flag rate driven by the boundary-proximity rule, this instructor review covered the majority of the cohort rather than an exception set. The supervision principle is consistent with Askari's (2025) "reliable but supervised" model of AI-assisted assessment in higher education, in which AI-generated feedback and grades are refined by an instructor before being delivered to students.
2. **Every grade report was delivered to the student for review before academic consequences followed.** Each consenting student received their grade report and a scan of their complete original exam paper by email, with a window of no fewer than five working days between delivery and grade finalisation, allowing genuine review.
3. **Students retained an unconditional right to request regrading by the instructor.** The consent form and the grade report email both stated that any student could request full manual regrading of any or all questions at any time, without justification and without burden of proof that the AI grade was wrong. Regrading, when requested, was performed on the original paper and the manual grade replaced the AI grade in the official record. The unconditional regrading right operationalises the principle that AI assistance should preserve rather than reduce student agency in assessment (Darvishi et al., 2024).
4. **Non-consent carried no observable penalty or distinction.** The sixteen papers from students who declined or did not submit the consent form (Table 1, Groups B and C) were graded manually by me using the same marking scheme and delivered through the same institutional channel on the same schedule. No student could identify from their grade report, from their grade in the university system, or from any communication with me whether their paper had been graded by AI or by me.

The combined effect is that no student could experience academic harm as a consequence of consenting to participate: every AI-assigned grade passed through human review, every student could review and contest their grade, regrading was unconditional, and non-consent produced no disadvantage.

Data are protected under a two-tier scheme. The anonymised rubric files, configuration file, folder structure, delivery and PDF conversion scripts, grading outputs, R analysis scripts, run environment log, and procedural and replication records are openly available (see Data Availability Statement). The student roster, the scanned papers, and the blind human grading spreadsheet remained local to my device throughout the study and are available only by request to me, subject to institutional approval.

# 4. Results

This section reports the four reliability findings of the study. Section 4.1 presents overall agreement between the human and AI graders on total scores (RQ1). Section 4.2 reports agreement decomposed by question type (RQ2). Section 4.3 reports test-retest consistency across two independent grading sessions (RQ3). Section 4.4 reports the accuracy of the AI-generated feedback (RQ4). Throughout, all statistics use the corrected grades described in Section 3.6, in which seven AI-assigned grades and eleven human grades were revised following the post-grading review.

## 4.1 Overall Agreement (RQ1)

The agreement sample comprised 20 students drawn systematically from the 60 consenting papers. Both graders produced an exam total out of 20 marks for each student. The descriptive statistics for the two graders are reported in Table 4. The means differed by less than one mark, and the score ranges overlapped substantially.

Table 4. Descriptive statistics for the agreement sample (n = 20)

| Grader | Mean | SD | Min | Max | Median |
|---|---|---|---|---|---|
| **Human** | 11.45 | 2.79 | 6.75 | 17.00 | 11.13 |
| **Claude Code** | 12.13 | 2.74 | 7.75 | 16.50 | 12.13 |

*Total scores out of 20 marks. The agreement sample comprised every third paper from the 60 consenting students, sorted alphabetically (S001, S004, S007 ... S058).*

The two sets of total scores were strongly correlated, and the magnitude of disagreement was small. Table 5 reports the full set of agreement statistics. The Pearson correlation was 0.97 with a bootstrap 95% confidence interval of [0.94, 0.99] across 10,000 resamples ($p < 0.001$). The Spearman rank correlation, included as a robustness check, was 0.95 ($p < 0.001$). The Mean Absolute Error between the two graders was 0.85 marks out of 20.

Table 5. Overall agreement statistics (RQ1) and comparison with prior LLM grading studies.

| Statistic | Value | 95% CI / test | Comparison with literature |
|---|---|---|---|
| Pearson r | 0.97 | [0.94, 0.99] | Higher than typical reported correlations: Pack et al. (2024) reported r = 0.73 for Claude 2 on ELL essays; Stowell and Zhu (2025) reported ICCs of 0.21 to 0.81 for ChatGPT in psychology essays; Yavuz (2025) reported moderate correlations for ChatGPT and Bard in EFL essays. |
| Spearman ρ | 0.95 | $p < 0.001$ | Robustness check; non-parametric form. |
| MAE | 0.85 | — | In line with the deviation magnitudes reported by Caraeni et al. (2024) for GPT-4o on handwritten college mathematics, who observed a 7.66% deviation in their best condition. |
| Mean bias (H − AI) | −0.68 | $t(19) = -4.18$, $p < 0.001$ | Direction consistent with Flodén (2025), who observed that ChatGPT 3.5 tended to give marginally higher scores. The present study quantifies this bias and tests it inferentially. |
| SD of differences | 0.68 | — | — |
| 95% Limits of Agreement | [-2.01, +0.66] | — | — |

*95% CI for Pearson r computed via 10,000 bootstrap resamples (percentile method). LoA = limits of agreement (Bland and Altman, 1999).*

A Bland-Altman analysis was conducted to characterise the pattern of agreement. The mean difference between human and AI scores was −0.68 marks, indicating that the AI grader awarded slightly higher totals than the human grader on average. A one-sample t-test on the differences confirmed that this systematic shift was statistically distinguishable from zero ($t(19) = -4.18$, $p < 0.001$). The 95% limits of agreement, computed as the bias plus or minus 1.96 standard deviations of the differences, were [−2.01, +0.66]. The Shapiro-Wilk test on the differences ($W = 0.94$, $p = 0.21$) supported the assumption of approximate normality required for parametric limits of agreement. The Bland-Altman plot is shown in Figure 5.

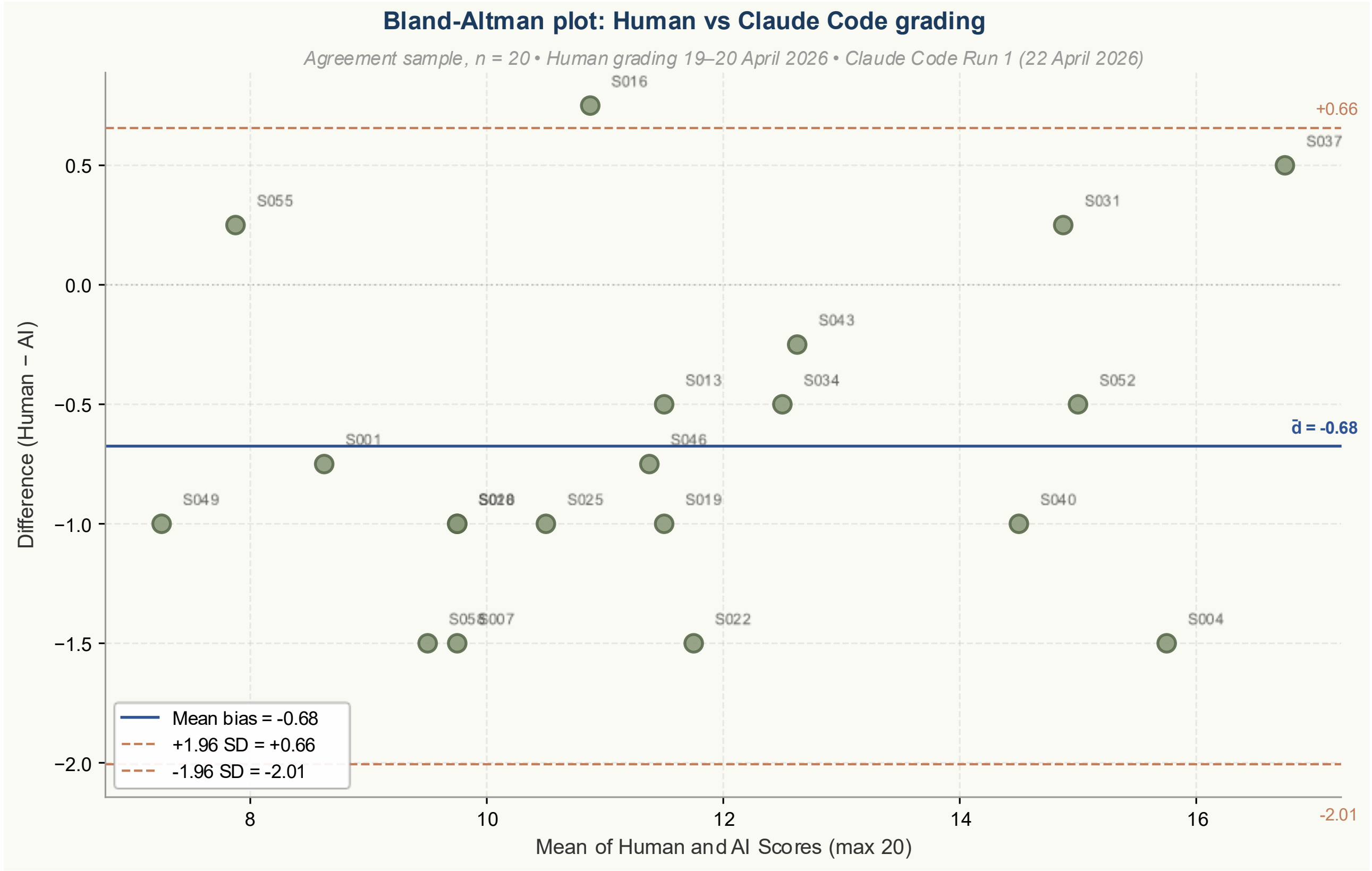

Figure 5: Bland-Altman plot - Human vs Claude Code grading

The strong correlation, low MAE, and modest systematic bias together describe a reliability profile in which the two graders rank papers in nearly the same order but the AI tends to assign slightly higher marks. The bias of approximately two-thirds of a mark on a twenty-mark exam is small in absolute terms, but it is consistent in direction and statistically detectable at the sample size used here.

## 4.2 Agreement by Question Type (RQ2)

To examine whether agreement varied across the five question types in the exam, weighted Cohen's kappa was computed for each type using quadratic weights, treating each per-question grade as an ordinal category derived from the smallest grading increment used for that item. The primary results are reported in Table 6.

Table 6. Agreement by question type, weighted Cohen's kappa with comparison to literature (RQ2).

| Question Type | Items | n ratings | Weighted κ | MAE | Exact agreement | Interpretation | Comparison with literature |
|---|---|---|---|---|---|---|---|
| MCQ | Q1.1–Q1.5 | 100 | 0.98 | 0.01 | 99/100 (99%) | Almost perfect | At the upper end of the QWK = 0.99 ceiling reported in the LLM grading systematic review by Emirtekin (2025). |
| Written | Q2, Q3.1–Q3.3 | 80 | 0.73 | 0.16 | 52/80 (65%) | Substantial | Comparable to QWK ranges reported across the Emirtekin (2025) review (0.45 to 0.99) for short-answer items. |
| Drawing | Q4 | 20 | 0.74 | 0.33 | 11/20 (55%) | Substantial | Drawing performance falls within the same range, but with the lowest exact-agreement rate; consistent with the diagram-item difficulty reported by Kortemeyer et al. (2024). |
| Analysis | Q5.1–Q5.2, Q6.1–Q6.4 | 120 | 0.67 | 0.15 | 75/120 (62%) | Substantial | Within the moderate-to-substantial range reported for analytical short-answer items. |
| Evaluation | Q7.1–Q7.4 | 80 | 0.65 | 0.21 | 47/80 (59%) | Substantial | Within the range observed for evaluation items in Flodén (2025). |

*Weighted Cohen's kappa computed with quadratic weights using the irr package in R. Interpretation per Landis and Koch (1977) and Viera and Garrett (2005). MAE expressed in marks for each question's specific scoring scale.*

Agreement varied across question types. The MCQ items achieved the highest weighted kappa (0.98) and were within "almost perfect" range. The four other question types each achieved kappa values in the "substantial" range (0.65 to 0.74). The pattern is consistent with the prediction that AI grading would be most reliable on items with discrete correct answers and less reliable on items requiring evaluative judgement or visual interpretation. The Drawing item shows the lowest exact agreement at 55%, which reflects the difficulty of interpreting hand-drawn diagrams from scanned images.

To examine whether the pattern was driven by specific items within each type, per-question kappa values were also computed. The results are reported in Table 7 as supplementary detail.

Table 7. Per-question agreement statistics (supplementary, RQ2)

| Question | Max marks | n | Weighted κ | MAE | Exact agreement |
|---|---|---|---|---|---|
| Q1.1 | 1 | 20 | 1.00 | 0.00 | 100% |
| Q1.2 | 1 | 20 | 0.95 | 0.03 | 95% |
| Q1.3 | 1 | 20 | 1.00 | 0.00 | 100% |
| Q1.4 | 1 | 20 | 0.92 | 0.05 | 95% |
| Q1.5 | 1 | 20 | 0.95 | 0.08 | 95% |
| Q2 | 1 | 20 | 0.79 | 0.10 | 75% |
| Q3.1 | 1 | 20 | 0.62 | 0.23 | 60% |
| Q3.2 | 1 | 20 | 0.75 | 0.15 | 70% |
| Q3.3 | 1 | 20 | 0.70 | 0.18 | 65% |
| Q4 | 3 | 20 | 0.74 | 0.33 | 55% |
| Q5.1 | 1 | 20 | 0.26 | 0.28 | 55% |
| Q5.2 | 1 | 20 | 0.15 | 0.25 | 60% |
| Q6.1 | 0.5 | 20 | 0.78 | 0.06 | 80% |
| Q6.2 | 0.5 | 20 | 0.69 | 0.09 | 75% |
| Q6.3 | 0.5 | 20 | 0.82 | 0.04 | 85% |
| Q6.4 | 0.5 | 20 | 0.73 | 0.08 | 80% |
| Q7.1 | 1 | 20 | 0.65 | 0.23 | 60% |
| Q7.2 | 1 | 20 | 0.71 | 0.20 | 60% |
| Q7.3 | 1 | 20 | 0.62 | 0.23 | 55% |
| Q7.4 | 1 | 20 | 0.63 | 0.20 | 60% |

*Two questions (Q5.1, Q5.2) on the analysis of refresh rate and frame rate fell below the moderate-agreement threshold and are discussed in Section 5.*

At the individual question level, two items showed weighted kappa values below 0.40. Q5.1 reached 0.26 and Q5.2 reached 0.15, both relating to the analysis of refresh rate and frame rate. Closer inspection of these cases suggested that the AI grader tended to award marks for definitions of Hertz and FPS that were technically present in the student's response but did not address the friend's misleading advice as required by the rubric. This pattern resembles the keyword-matching failure mode described in Section 5.3.

### 4.3 Test-Retest Consistency (RQ3)

The consistency sample comprised 15 students disjoint from the agreement sample. The same 15 papers were graded twice by Claude Code, with one week separating the two sessions. Both sessions used an identical configuration: Claude Opus 4.6 invoked through Claude Code at the xhigh effort level, the same three rubric files, the same start instruction, and a fresh session with no carry-over from the first run.

The two AI grading runs produced descriptively similar totals. Run 1 had a mean total of 11.82 (SD = 3.68, range 5.75 to 17.00), and Run 2 had a mean total of 12.00 (SD = 3.85, range 4.75 to 17.50). The mean shift from Run 1 to Run 2 was −0.18 marks, and the largest single-paper difference was 2.00 marks. No paper crossed a grade boundary between the two runs.

The intraclass correlation coefficient ICC(2,1) for absolute agreement on total scores is reported in Table 8. ICC(2,1) was 0.96 with a 95% confidence interval of [0.89, 0.99], $F(14, 14) = 52.58$, $p < 0.001$. Per Koo and Li (2016), ICC values above 0.90 fall in the "excellent" range.

Table 8. Test-retest consistency on the consistency sample with comparison to literature (RQ3, n = 15)

| Measure | Value | 95% CI / test | Comparison with literature |
|---|---|---|---|
| ICC(2,1) absolute agreement | 0.96 | [0.89, 0.99]; $F(14, 14) = 52.58$; $p < 0.001$ | At the upper end of reported AI grading consistency. Pack et al. (2024) reported ICC = 0.93 for Claude 2; Stowell and Zhu (2025) reported ICCs of 0.21 to 0.81 for ChatGPT on essays. |
| Pearson $r$ (R1 vs R2) | 0.96 | $p < 0.001$ | — |
| MAE (R1 vs R2) | 0.78 marks | — | — |
| Mean shift (R1 − R2) | −0.18 marks | — | Smaller than the 40% to 45% letter-grade drift across repeated gradings reported by Flodén (2025) for ChatGPT 3.5. |
| Maximum absolute difference | R1 − R2 | | 2.00 marks |

*Note: ICC interpretation per Koo and Li (2016). Run 1 conducted on 22 April 2026; Run 2 conducted approximately one week later. Both runs used identical Claude Code configuration: Claude Opus 4.6 at xhigh effort level, identical rubric files, identical start instruction, fresh session with no Run 1 context carried over.*

When the ICC analysis was decomposed by question type, the rank order observed for human-AI agreement in Section 4.2 was reproduced. The MCQ items reached an ICC of 0.92 with 96% exact agreement across runs. The Drawing item reached 0.52, with only 5 of 15 ratings in exact agreement and a wide 95% confidence interval of [0.01, 0.80]. The width of this confidence interval, which spans the full poor-to-good range, reflects the small per-type sample of 15 ratings and means the Drawing point estimate should be treated as suggestive rather than definitive. The remaining three types fell between 0.78 and 0.83. Per-type ICC results are reported in Table 9.

Table 9. Per-question-type ICC(2,1) (RQ3)

| Question Type | n ratings | ICC(2,1) | 95% CI | MAE | Exact agreement | Interpretation |
|---|---|---|---|---|---|---|
| MCQ | 75 | 0.92 | [0.87, 0.95] | 0.04 | 72/75 (96%) | Excellent |
| Written | 60 | 0.78 | [0.67, 0.87] | 0.13 | 46/60 (77%) | Good |
| Drawing | 15 | 0.52 | [0.01, 0.80] | 0.50 | 5/15 (33%) | Moderate |
| Analysis | 90 | 0.82 | [0.74, 0.88] | 0.06 | 74/90 (82%) | Good |
| Evaluation | 60 | 0.83 | [0.73, 0.90] | 0.08 | 50/60 (83%) | Good |

*Interpretation per Koo and Li (2016). The Drawing item is the only question type with a confidence interval crossing the 0.50 threshold below which ICC is interpreted as poor. The per-type rank order replicates the pattern observed in RQ2 (Table 6).*

A tertiary analysis was conducted following the framework of Shyr et al. (2025), which distinguishes repeatability (consistency of an instrument with itself) from accuracy (agreement of an instrument with a reference standard). On the same 15 papers, both AI runs were compared against the human grades produced during the blind grading session. The results are reported in Table 10. The AI runs showed slightly tighter agreement with each other ($r = 0.96$) than with the human grader (Run 1: $r = 0.95$; Run 2: $r = 0.93$), and both runs reproduced the systematic positive bias observed in the agreement sample at Section 4.1. The repeatability-versus-accuracy divergence reported by Xue et al. (2026) is empirically supported by this pattern, although the divergence in the present study is small.

Table 10. Tertiary analysis: repeatability vs accuracy (RQ3, n = 15)

| Comparison | Pearson r | MAE | Mean difference | Type |
|---|---|---|---|---|
| Run 1 vs Run 2 | 0.96 | 0.78 | −0.18 | Repeatability |
| Run 1 vs Human | 0.95 | 1.45 | +1.18 | Accuracy (Run 1) |
| Run 2 vs Human | 0.93 | 1.63 | +1.37 | Accuracy (Run 2) |

*Following the framework of Shyr et al. (2025), repeatability and accuracy are reported separately. Both AI runs reproduced the systematic positive bias relative to the human grader observed in the agreement sample (Section 4.1). The pattern in which intra-LLM consistency exceeds AI-human accuracy is consistent with the divergence predicted by Xue et al. (2026).*

The three-panel visualisation of these comparisons is shown in Figure 6, and the score distributions for the human grader and both AI runs on the consistency sample are shown in Figure 7. The three distributions overlap substantially, and the rightward shift of the AI distributions relative to the human distribution visualises the systematic positive bias quantified in Tables 8 and 10.

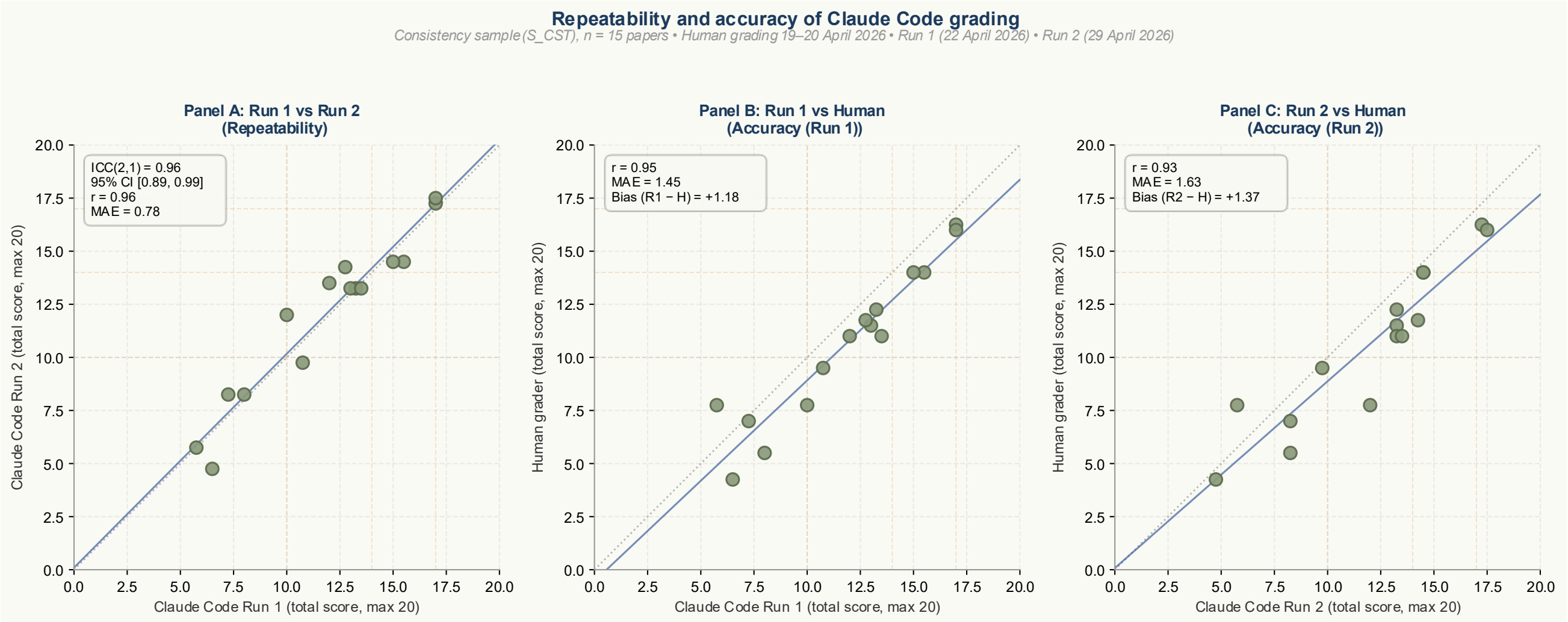

Figure 6: Repeatability and accuracy of Claude Code grading

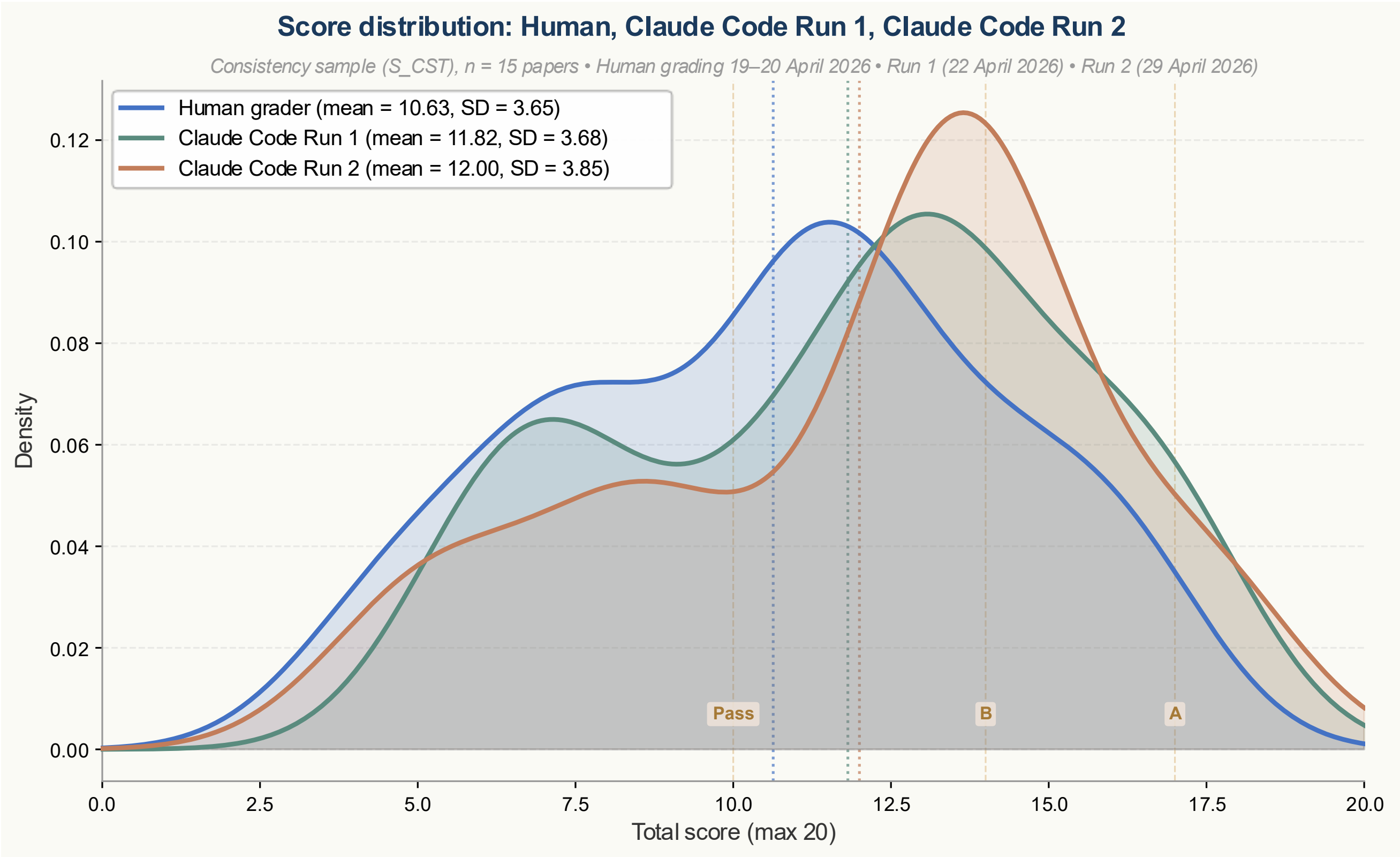

Figure 7: Score distribution - Human, Claude Code Run 1, Claude Code Run 2

### 4.4 Feedback Accuracy (RQ4)

I coded the AI-generated feedback for the 20 papers in the agreement sample using a four-category scheme adapted from Seo et al. (2025) and Scarlatos et al. (2024). The accuracy categories were partitioned by feedback function: C1 codes accurate diagnostic feedback on items where the student lost marks, C2 codes accurate confirmation feedback on items where the student earned full marks, C3 codes partially accurate feedback regardless of function, and C4 codes inaccurate feedback regardless of function. The coding was performed by me alone as the subject-matter expert for the course. This single-coder design is acknowledged as a limitation in Section 5.

The full coding produced 400 coded items (20 students × 20 questions). The overall four-code distribution and the two reporting layers (diagnostic accuracy and confirmation accuracy) are reported in Table 11.

Table 11. Feedback accuracy by reporting layer (RQ4).

| **Measure** | **n items** | **C1 (accurate diagnostic)** | **C2 (accurate confirmation)** | **C3 (partially accurate)** | **C4 (inaccurate)** |
|---|---|---|---|---|---|
| Overall (all items) | 400 | 205 (51.2%) | 175 (43.8%) | 13 (3.2%) | 7 (1.8%) |
| Diagnostic items (lost marks) | 221 | 205 (92.8%) | — | 11 (5.0%) | 5 (2.3%) |
| Confirmation items (full credit) | 179 | — | 175 (97.8%) | 2 (1.1%) | 2 (1.1%) |

*Coding scheme adapted from Seo et al. (2025) and Scarlatos et al. (2024). C1 applies only to lost-mark items; C2 applies only to full-credit items; C3 and C4 apply to either function.*

Diagnostic accuracy, computed as the proportion of C1 codes among items where the student lost marks, was 92.8%. Confirmation accuracy, computed as the proportion of C2 codes among items where the student earned full marks, was 97.8%. Together, accurate feedback in either function accounted for 95.0% of all coded items.

When diagnostic accuracy was disaggregated by question type, the same gradient observed in Sections 4.2 and 4.3 emerged. The results are reported in Table 12. MCQ diagnostic accuracy was 91.7%, Written 93.1%, Analysis 98.5%, and Evaluation 100.0%. The Drawing item again stood out as the weakest case, with a diagnostic accuracy of 27.3%. Of the 11 diagnostic Drawing items, 36.4% were coded as partially accurate and 36.4% as inaccurate. Confirmation accuracy showed the same outlier pattern: 100.0% across the other four question types, but 55.6% on Drawing. Figure 8 visualises both the diagnostic and confirmation accuracy values across the five question types, with the Drawing item visibly separated from the other four on both measures.

Table 12. Feedback accuracy by question type with comparison to literature (RQ4)

| **Question Type** | **Diagnostic items (n)** | **C1 %** | **C3 %** | **C4 %** | **Confirmation items (n)** | **C2 %** | **C3 %** | **C4 %** | **Comparison with literature** |
|---|---|---|---|---|---|---|---|---|---|
| MCQ | 36 | 91.7 | 5.6 | 2.8 | 64 | 100.0 | 0.0 | 0.0 | Higher than the moderate diagnostic feedback ratings reported by Er et al. (2025) for AI feedback in higher education. |
| Written | 58 | 93.1 | 6.9 | 0.0 | 22 | 100.0 | 0.0 | 0.0 | Comparable to the AI feedback accuracy levels described by Pack et al. (2024) and Stowell and Zhu (2025) for short written items. |
| Drawing | 11 | 27.3 | 36.4 | 36.4 | 9 | 55.6 | 22.2 | 22.2 | No comparator available; the literature on AI feedback accuracy for handwritten diagrammatic items is sparse. |
| Analysis | 68 | 98.5 | 1.5 | 0.0 | 52 | 100.0 | 0.0 | 0.0 | Consistent with the upper end of reported feedback accuracy for analytical items. |
| Evaluation | 48 | 100.0 | 0.0 | 0.0 | 32 | 100.0 | 0.0 | 0.0 | At the upper end of values reported across the literature. |

*The Drawing question (Q4) is the only category where feedback accuracy on either function falls below 90%. The pattern matches the consistency results reported in Table 9.*

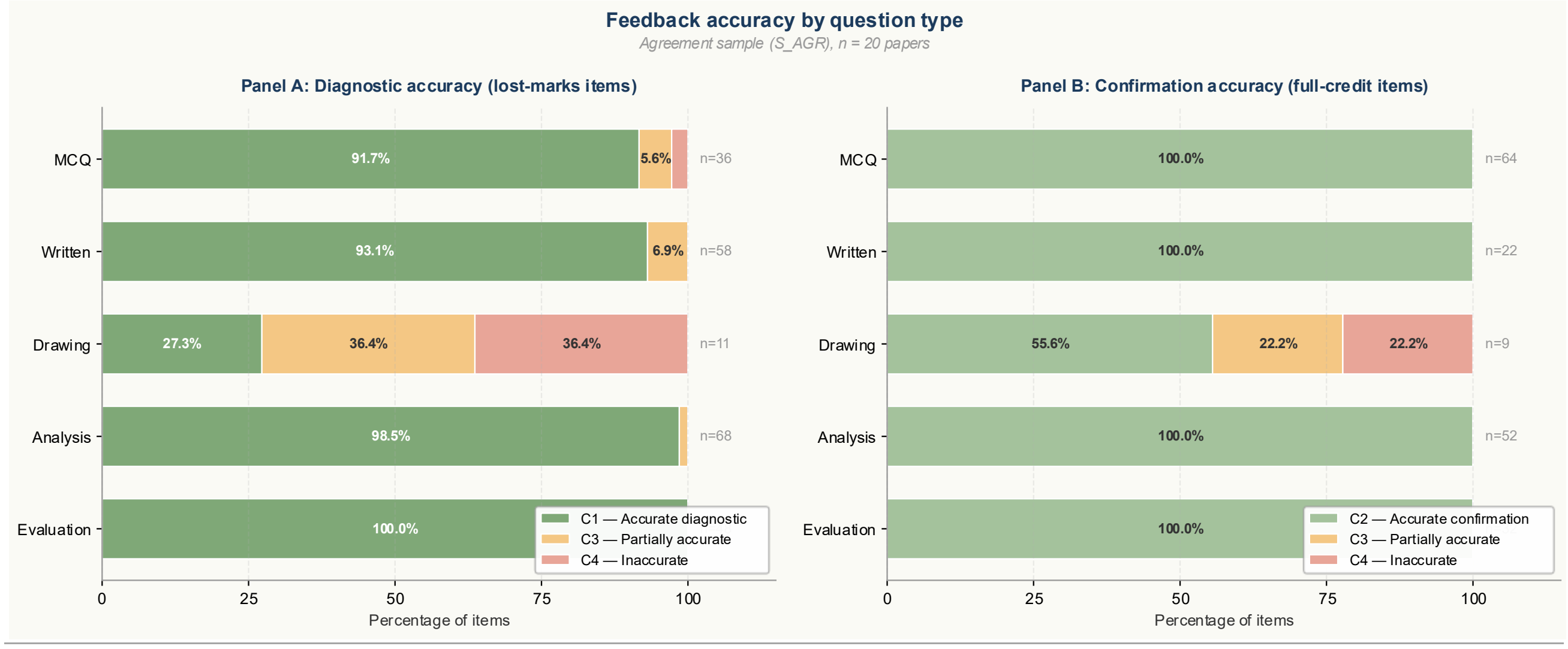

Figure 8: Feedback accuracy by question type

The qualitative review of feedback strings during coding identified three named failure modes that recurred across multiple papers and are consistent with the patterns observed in the quantitative analyses. First, on Q3.1 and Q3.3, the AI awarded credit and provided diagnostic feedback acknowledging keyword presence without verifying conceptual understanding (five overridden cases across S001, S022, and S046). This pattern is related to, but distinct from, the failure mode Flodén (2025) reported for ChatGPT 3.5, which credited answers addressing a related but different question (Section 5.3). Second, on Q1.4 and Q1.5, the AI inconsistently interpreted student corrections, sometimes reading the original (crossed-out) selection rather than the final corrected selection (S016 and S025 cases). Third, on Q4 drawings, the AI's feedback was at times influenced by student annotations on the page rather than the drawing itself (S046 cases). These observations are consistent with the substantial gap between Drawing reliability and the other question types observed across all four research questions.

### 4.5 Summary of Reliability Findings

Across the four research questions, four convergent observations emerge. First, AI grading achieved strong agreement with human grading on total scores ($r = 0.97$, MAE = 0.85 marks), with a small but statistically detectable positive bias of approximately two-thirds of a mark. Second, AI grading was excellent on test-retest consistency (ICC = 0.96), with no paper crossing a grade boundary between runs. Third, the AI feedback was accurate on 95% of the coded items, with diagnostic accuracy of 92.8% on items where students lost marks and confirmation accuracy of 97.8% on full-credit items. Fourth, all four reliability measures showed a consistent gradient across question types: MCQ items were strongest on every measure, and drawing items were the clear outlier on three of four measures. On the kappa measure, drawing items fell within the same cluster as the written, analytical, and evaluative items. Figure 9 visualises this gradient by plotting weighted kappa, ICC, diagnostic accuracy, and confirmation accuracy on a common scale for each question type.

The pattern in Figure 9 is consistent across all four reliability dimensions. MCQ items reach or approach the upper bound on every measure ($\kappa = 0.98$, ICC = 0.92, diagnostic 0.92, confirmation 1.00). Written, Analysis, and Evaluation items fall in a similar middle band, with values between 0.65 and 1.00 across the four measures. The Drawing item is the only category that falls below the 0.50 threshold on any measure (ICC = 0.52) and the only category with diagnostic accuracy below 0.50 (0.27). On three of the four reliability measures, the gap between Drawing and the next-weakest format is wider than the gap between any other pair of types. On the agreement measure (kappa), Drawing falls within the same cluster as Written, Analysis, and Evaluation. The pattern across measures supports the conclusion that hand-drawn diagrammatic items are a structurally distinct case for the AI grading system, especially on measures that capture the absolute level of grading errors rather than category-level agreement.

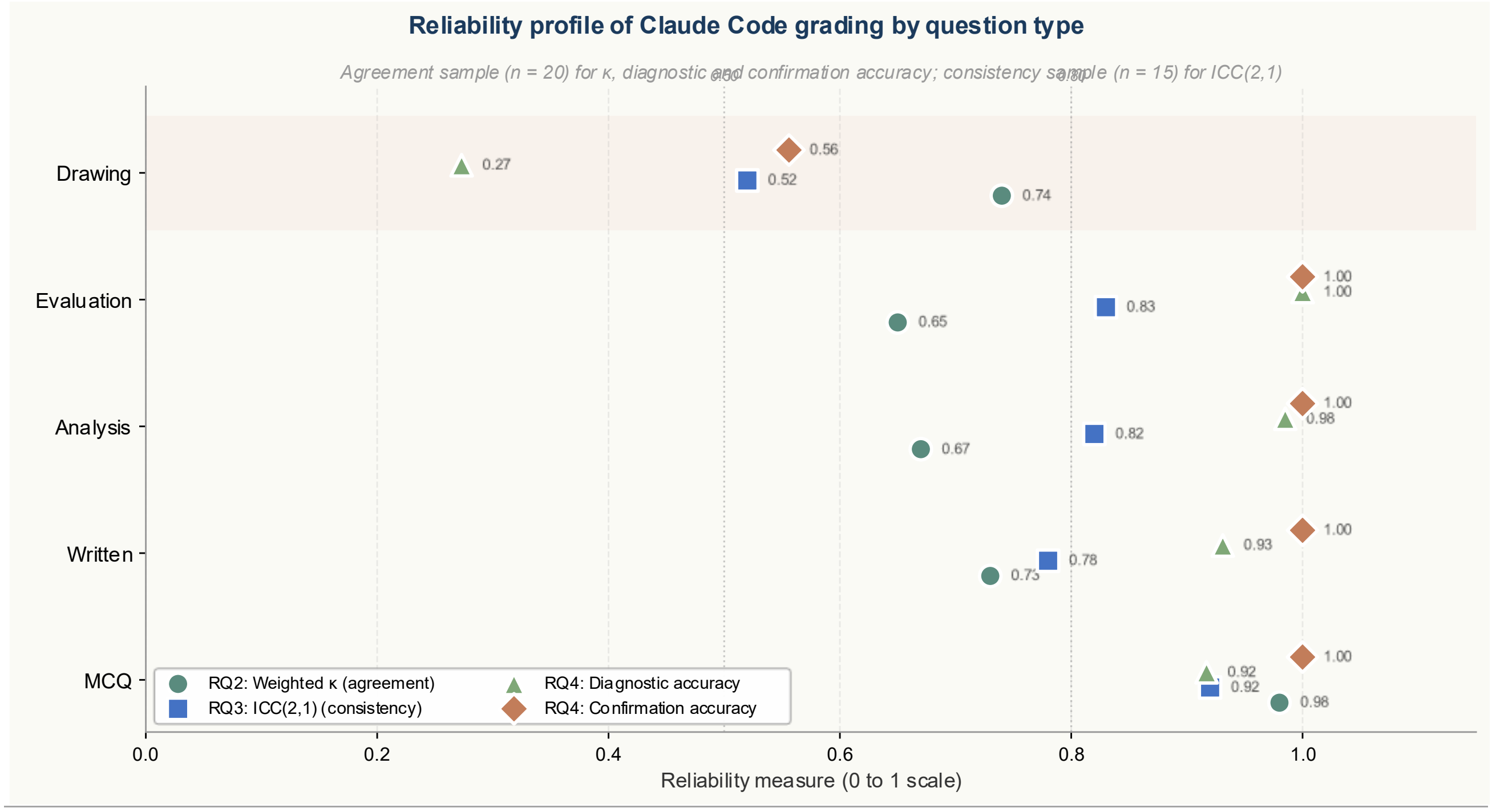

Figure 9: Reliability profile of Claude Code grading by question type

## 5. Discussion

The four research questions in this study were chosen because they together constitute the minimum evidence base an instructor would need before deciding whether AI grading is ready for deployment in their classroom. An instructor wants to know whether the AI agrees with them on average (RQ1), whether the agreement holds across the kinds of questions they actually ask (RQ2), whether the AI is stable enough to give similar grades on similar work over time (RQ3), and whether the explanations it offers students are accurate (RQ4). The findings reported in Section 4 answer each of these questions, and the answers point in the same direction. AI grading by Claude Code in an agentic configuration was strong on objective items, relatively weaker on items requiring open-ended reasoning, and visibly limited on items requiring visual interpretation. The systematic positive bias of 0.68 marks (about two-thirds of a mark on the 20-mark scale), the divergence between intra-AI consistency and AI-human accuracy, and the named failure modes identified through qualitative review each contribute a different perspective on what that pattern means. This section discusses each in turn, situates the findings within the published literature, and offers an interpretation of what AI grading can and cannot reliably do at present.

### 5.1 Strong agreement, but with a small bias that has practical consequences

The Pearson correlation of 0.97 between human and AI total scores is, on its own, a reassuring number. It places this study at the upper end of correlations reported across recent peer-reviewed LLM grading work. Pack et al. (2024) reported a correlation of 0.73 between Claude 2 and human raters on 119 English-language-learner essays graded twice across two separate occasions. Zhang et al. (2024) found an interrater consistency ICC of 0.72 between Claude 3 Haiku and IELTS examiners on 31 IELTS academic writing samples. In a study of ChatGPT-graded psychology essays with retrieval-augmented generation, Stowell and Zhu (2025) observed intraclass correlations ranging from 0.21 to 0.81 across grading dimensions. The figure reported here sits above these comparators. The interpretation that I draw is not that Claude Code is uniformly more accurate than the LLMs tested in those studies, but that the agentic configuration may have reduced prompt-by-prompt variation that researchers in chat-based studies must manage through prompt engineering. One plausible reason the agreement here sits at the upper end of the published range is that the marking scheme, answer key, and output format are stored in persistent files read at the start of every session, rather than reconstructed in each prompt.

The more interesting finding from RQ1 is the bias. The mean human-minus-AI difference was −0.68 marks, with a 95% confidence interval that excluded zero ($t(19) = -4.18$, $p < 0.001$); equivalently, the AI grader assigned a mean 0.68 marks higher than the human grader (95% confidence interval [0.35, 1.01]). The magnitude corresponds to approximately 3.4% of the 20-mark scale, which is comparable to the 3.1 percentage points reported by Flodén (2025) on a different cohort, model, and modality. The direction is consistent with four prior studies. Flodén (2025) reported that ChatGPT 3.5 tended to give marginally higher scores than human teachers (3.1 percentage points on average). Morjaria et al. (2024) found that ChatGPT-4 systematically scored higher than human assessors of medical short-answer responses when no question-specific rubric was provided, with mean differences of 0.37 and 0.47 on a 5-point scale (7.4 and 9.4% of the scale); the bias became non-significant once a rubric was included. Kortemeyer et al. (2024) similarly observed that AI graded a German-language handwritten thermodynamics exam more leniently than human teaching assistants on the same fine-grained rubric. The follow-up study by Kortemeyer et al. (2025) using GPT-5 on calculus open-ended responses also reported AI total scores systematically higher than TA total scores (7.89 vs 7.35). The direction contrasts with Zhang et al. (2024), who found that ChatGPT 3.5 graded IELTS essays significantly lower than official examiners (Cohen's d = 0.71), and with Wang et al. (2025), who reported that humans were generally more lenient than LLMs across 4,315 essays in three genres. The direction of LLM bias relative to human grader is not a fixed property of LLMs; it appears to depend on the specific model, the rubric, the assessment context, and the language of the responses. Grévisse (2024) tested three Western European languages (French, English, German) on typed responses and found only a weak correlation between language and LLM grade. To the best of the author's knowledge, the Arabic-English bilingual handwritten setting of the present study has not been systematically investigated in prior work. In the present study, three structured rubric files were provided to Claude Code, yet a positive AI-minus-human bias persisted at 0.68 marks; this suggests that rubric provision alone may not eliminate bias in agentic LLM grading, even though Morjaria et al. (2024) found rubric provision sufficient to render the bias non-significant for ChatGPT-4 in chat-based grading. The difference between studies may reflect the deeper integration of rubric content in agentic grading (where rubric files are read once at session start and remain in context), the larger absolute scale (20 marks rather than 5), or genuine differences between Claude Opus 4.6 and ChatGPT-4. This study extends Flodén's (2025) observation in three respects. First, the bias is tested inferentially against zero rather than reported descriptively. Second, it is observed in handwritten Arabic-English bilingual responses rather than typed monolingual ones. Third, the bias is replicated across two independent grading sessions separated by one week, in contrast to Flodén's within-session three-run design, which rules out short-term session noise as the explanation. The bias is small, but it is real, and it is reproducible.

For an instructor considering AI grading, the practical question is whether a bias of approximately two-thirds of a mark matters. The answer depends on where the bias falls relative to the grade boundaries. The 95% limits of agreement, reported in Section 4.1, were [−2.01, +0.66]. This means that for most papers, the AI score will fall within approximately two marks below the human score and just under one mark above it. On a twenty-mark exam, two marks are enough to move a student between grade categories. The students at the boundaries are therefore the students most affected by AI–human disagreement, and these are exactly the students for whom grading judgements are most consequential. The four structural safeguards described in Section 3.7, and in particular the requirement that every grade pass through human review before entering the official record, are the protection against this concern. They translate the statistical finding into a deployment recommendation: AI grading is reliable enough to assist instructor grading, but not reliable enough to replace it for students whose totals fall near grade boundaries.

### 5.2 The question-type gradient is the actionable finding

If the bias is the headline number, the question-type gradient is what an instructor designing an AI-graded exam will most need. Section 4.2 reported weighted Cohen's kappa values ranging from 0.98 on multiple-choice items to 0.65 on evaluation items. On test-retest consistency, diagnostic feedback accuracy, and confirmation feedback accuracy, Drawing was the clear outlier at the weak end of the gradient. On the kappa measure, Drawing fell within the same substantial-range cluster as Written, Analysis, and Evaluation ($\kappa$ = 0.65–0.74), with Drawing's value ($\kappa$ = 0.74) actually the highest of the four. The convergence is therefore strong on three of four reliability measures, while the kappa measure flattens the gradient at the lower end and obscures Drawing's distinctness. The structural separation of drawing from the other formats is most visible in measures sensitive to absolute scale (ICC, diagnostic accuracy) rather than in measures of category agreement. Figure 9 visualises this convergence across four reliability properties on a single coordinate system. The pattern is robust to the choice of reliability measure, which suggests that the gradient reflects a property of the AI grading system rather than an artefact of any one analytical procedure.

The mechanism behind the gradient is intuitive once stated. Multiple-choice items have discrete correct answers and clean partial-credit boundaries. The AI's task on these items is reduced to recognising which option the student selected, which is essentially a perception task that the model performs well when handwriting is legible. Written and analytical items require the AI to assess whether the student's prose communicates a particular concept, which is a comprehension task that the model performs reasonably well when the rubric specifies the conceptual targets clearly. Drawing items require the AI to interpret a hand-drawn diagram against a reference image, which is a perceptual task on a visual representation that handwritten student diagrams are unlikely to closely match. Evaluation items require the AI to distinguish a correct argument from a related but incorrect one (the failure mode Flodén (2025) identified as dominant for ChatGPT 3.5), and the lower $\kappa$ on Q7 in the present study ($\kappa$ = 0.65) is consistent with that pattern. Kortemeyer et al. (2024) reported a similar pattern in their study of 252 handwritten thermodynamics exams, observing that diagram items were less reliable than text items. At the meta-analytic level, Emirtekin's (2025) systematic review of 49 LLM-grading studies reported quadratic weighted kappa values ranging from 0.45 to 0.99 and correlations from 0.38 to 0.95 across the included studies, confirming that reliability varies widely depending on task and instrument. The present study extends both lines of evidence by characterising the gradient across five question types within a single assessment instrument, which allows differences between formats to be attributed to the formats themselves rather than to differences between studies.

What this means for institutional deployment is that AI grading is not equally appropriate for all assessment formats. An instructor designing an exam with AI grading in mind would do well to weigh more heavily the formats where the AI performs reliably, and to plan for human review of the formats where it does not. The findings here suggest that MCQ-heavy exams are good candidates for AI grading with light human oversight, that written, analytical, and evaluation items are acceptable candidates with normal review, and that drawing items require human review of every paper. The convergent evidence in Figure 9 is what supports this differentiated recommendation: where four reliability properties point in the same direction across question types, the deployment implication can be drawn with greater confidence than from any single property alone.

### 5.3 Three named failure modes ground the quantitative pattern in real cases

The numbers in Section 4 describe what the AI got right and wrong in aggregate. They do not, on their own, explain why. During the qualitative review accompanying the RQ4 coding, three failure modes emerged that recurred across multiple papers and offered specific mechanisms for the gradient observed in Section 4.2.

The first I have called keyword matching without conceptual depth. On Q3.1 and Q3.3, both of which asked students to explain how greyscale values in physically-based rendering maps relate to material properties, the AI awarded credit on five question-instances across three papers (S001, S022, S046) for answers that contained the surface vocabulary of the topic (smooth, rough, map, material) but did not articulate the underlying mechanism. A representative case is S001 Q3.1, where the student wrote a single sentence describing the appearance of the maps without explaining what the values mean to the renderer. The AI's diagnostic feedback acknowledged the keyword presence and treated it as evidence of understanding. The human grader, on review, treated the keyword presence as evidence only of vocabulary recall. The pattern recurred on two further papers (S022, S046), each on both Q3.1 and Q3.3. The cluster on Q3.1 and Q3.3 specifically, rather than across the rubric uniformly, suggests that the AI is most susceptible to this failure mode on questions where the topic vocabulary is distinctive enough to substitute for the explanation. This pattern is related to but distinct from the failure mode Flodén (2025) identified in ChatGPT 3.5, where the AI accepted off-topic answers addressing a related question. Both modes share an underlying weakness, namely that the model rewards surface relatedness without verifying that the response addresses the actual demand of the question, and the present study extends Flodén's observation from typed Master's-level responses in monolingual English to handwritten undergraduate responses in Arabic-English.

The second failure mode I have called selection-state ambiguity on MCQ. On Q1.4 and Q1.5, several students selected an option, then crossed it out and selected a different one. The AI did not handle these corrections consistently. The clearest case is S016 Q1.5, where the student initially circled the correct answer, crossed it out, and circled an incorrect one. The AI read the original (crossed-out) circle as the selection and awarded the mark, even though the student's final selection was incorrect. On Q1.3 of the same paper, with the same correction pattern, the AI correctly tracked the student's final selection. The within-paper inconsistency is what makes this finding interesting. The model is not systematically biased toward reading either the first or the final selection; it appears to make this decision case-by-case in ways that the rubric files cannot constrain. This is a failure mode specific to the handwritten modality. Chat-based AI grading studies do not encounter it because the input is typed and the selection is unambiguous. Caraeni et al. (2024) examined GPT-4o on handwritten college mathematics responses and reported a 7.66% grading deviation in their best condition, attributing part of the residual error to the demands of interpreting handwritten input. Their finding establishes that handwritten-modality challenges are real but does not extend to the specific case of student self-corrections on multiple-choice items. To the best of the author's knowledge, no published study has reported the selection-state ambiguity pattern described here. As handwritten-modality LLM grading expands, this failure mode is one that future work and rubric design should address explicitly.

The third failure mode I have called region-of-attention drift on drawings. On Q4, the AI's evaluation was at times influenced by content on the page that lay outside the drawing region itself. The clearest illustration is S046 Q4A, where the student linked V4 to V5 in violation of the GL_LINES specification (which requires only the segments V1–V2 and V3–V4, with V5 discarded) and annotated the drawing with the words 'not stripes' as a private note. The AI weighted the annotation heavily and awarded full marks. The human grader saw the linking error and reduced the mark. A related sub-pattern, observable in S010 and S034, is that the AI evaluated the drawn shape correctly but missed systematic conceptual errors elsewhere on the page, such as drawings constructed on the wrong coordinate axes or starting from the wrong origin. The shape was correct; the coordinate system was wrong; the AI did not detect the second error. This sub-pattern parallels the first failure mode in a different sensory channel: the AI evaluates what is visually salient and misses what is conceptually critical, just as it earlier rewarded keyword presence and missed conceptual depth.

These three failure modes do not exhaust the cases observed in the agreement sample, but they account for the bulk of the disagreements that led to AI grade overrides during the post-grading review. Importantly, they all point toward a general principle: the AI grader performs well when the assessment task reduces to recognising surface features (a circled option, a key concept word, a drawn shape) and performs less well when the task requires distinguishing between surface presence and conceptual depth, or between the answer region and the surrounding page.

### 5.4 Repeatability and accuracy are not the same thing

Section 4.3 examined two related but distinct properties of the AI grading system: how closely the AI agrees with itself across runs, and how closely either run agrees with the human grader. By the criteria of Koo and Li (2016), an ICC point estimate above 0.90 falls in the excellent range, with the lower bound of the 95% confidence interval [0.89] in the present study sitting at the boundary between good and excellent. The same fifteen papers, graded by the same model under identical configuration one week apart, produced almost identical totals. The mean shift from Run 1 to Run 2 was 0.18 marks (Run 2 marginally higher than Run 1), and no paper crossed a grade boundary between the two runs. This is the kind of stability that institutional adopters would want to see, and it suggests that for the purposes of routine deployment, results from a single Claude Code grading session can be expected to replicate closely on a second pass.

But the tertiary analysis comparing each run against the human grader on the same fifteen papers tells a different story. Run 1 had a Pearson correlation of 0.95 with the human grader and a mean bias of +1.18 marks; Run 2 had a Pearson correlation of 0.93 and a mean bias of +1.37 marks. Both biases are larger in magnitude than the +0.68 bias observed on the agreement sample (Section 4.1) but consistent with it in direction. The two values are not directly comparable: the agreement sample (S_AGR) and the consistency sample (S_CST) are disjoint by construction (Section 3.4), and the two samples have different score distributions. The pattern that holds across both samples is the direction of the bias: the AI grader is positively biased relative to the human grader in both. The two AI runs agreed with each other slightly more closely than either agreed with the human grader, and both runs reproduced the systematic positive bias observed in the agreement sample. Shyr et al. (2025) made the conceptual case for distinguishing these properties in their repeatability-versus-reproducibility framework, arguing that an instrument's consistency with itself is logically distinct from its agreement with a reference standard. Xue et al. (2026) demonstrated the divergence empirically, showing that intra-LLM consistency at low temperature did not correlate with accuracy across three large language models. The present results provide an empirical instantiation of that distinction in a real exam setting, with the added feature that consistency and accuracy are estimated on disjoint samples rather than on the same sample.

For deployment, this matters more than it may appear. An instructor running Claude Code twice would conclude, correctly, that the system is stable. An instructor comparing Claude Code against their own grading would conclude, also correctly, that the system runs about a mark and a half higher than they do (1.18 to 1.37 marks across the two runs). Both observations are true; both are consequences of the same data. The implication is that test-retest reliability cannot be the only validation step for AI grading deployment. An institution adopting an AI grading tool also needs the human-side calibration step provided by structural safeguards of the kind described in Section 3.7. Stability is not accuracy, and a system can be reliably wrong.

### 5.5 Feedback is more accurate than grading is consistent

The RQ4 results add a layer that complicates the broader pattern. While the question-type gradient appears in agreement, consistency, and diagnostic feedback accuracy, confirmation feedback accuracy reached 100% on every question type except Drawing (55.6%). The AI almost always characterises correct answers accurately.

The four-code accuracy scheme used here was adapted from the correctness dimensions of Scarlatos et al. (2024) and Seo et al. (2025), and refined during coding to distinguish two distinct functions of accurate feedback: diagnostic feedback on items where students lost marks, and confirmation feedback on items where students earned full credit. I refined the scheme after observing during initial review that Claude Code's confirmation feedback on full-credit items was not mere acknowledgement but substantive elaboration of the student's reasoning. Confirmation feedback, therefore, needed its own accuracy criterion if the coding was to characterise what the AI was actually doing. The full reasoning for this refinement, and the reconciliation rule applied to codes that did not match item credit status, are documented in Section 3.6.

The diagnostic-confirmation split is therefore not just an analytical refinement but a substantive empirical finding: the same AI system that exhibits a question-type reliability gradient on grades does not exhibit the same gradient on confirmation feedback accuracy. The mechanism behind the diagnostic-confirmation difference is again intuitive. Confirmation is an easier cognitive task than diagnosis. To confirm a correct answer, the AI needs only to verify that the student's reasoning matches the marking scheme, which is mostly a comparison task. To diagnose an incorrect answer, the AI needs to identify what is wrong, why it is wrong, and how the correct response would differ, which is a generative explanatory task. The 92.8% diagnostic accuracy is the more deployment-relevant figure than the 97.8% confirmation accuracy, since the function of feedback in formative assessment is primarily to help students understand their errors and the gap between current and desired performance (Nicol & Macfarlane-Dick, 2006). Er et al. (2025) reported that students perceived instructor feedback as more useful than AI feedback. The present finding offers a partial qualifier: AI feedback can be substantively accurate at high rates on most question types in this exam. Whether students perceive it as useful is a separate question, addressed in a companion paper (AlGhamdi, in preparation).

The qualitative review of feedback strings during coding produced one observation worth noting alongside the quantitative result. On full-credit items, Claude Code's feedback consistently extended beyond confirming correctness. It restated the student's reasoning in clearer terms, named the concepts the student had implicitly used, and elaborated on the relationship between the answer and the broader course content. A representative case from S031 Q7.2 reads: "You correctly explained that polygon count controls 'details' of the shape, and distinguished that zoom quality 'depends on pixel, not polygon' and is related to the raster image. This clearly separates what polygons control (geometric detail) from what controls sharpness when zooming (raster/texture resolution). Well done." This is not a generic acknowledgement; it is a substantive elaboration that may serve as informal scaffolding for student learning. Whether such elaboration has measurable pedagogical effects is a question for future research.

### 5.6 Limitations

Six limitations of this study warrant acknowledgement.

First, the sample sizes are small. The agreement sample comprised twenty papers, and the consistency sample fifteen. While these samples were sufficient to detect the systematic bias ($p < 0.001$) and to characterise the question-type gradient on most measures, the Drawing ICC confidence interval was wide [0.01, 0.80], reflecting the $n = 15$ consistency-sample design and the rarity of Drawing items per paper. Larger samples would tighten such intervals and allow finer subgroup analyses. Sample sizes in this range are typical of recent peer-reviewed work on AI grading of handwritten or short-form responses, including the $n = 37$ sample in Stowell and Zhu (2025), the $n = 60$ sample in Morjaria et al. (2024), and the $n = 18$ handwritten mathematics evaluation in Caraeni et al. (2024). The constraint reflects the availability of consenting students in a single course at a single institution.

Second, the human grading was performed by a single rater, the author. The agreement statistics reported in Section 4, therefore, measure within-rater benchmark reliability against a domain-expert reference, not inter-rater reliability against a multi-grader human consensus. A two-rater design would have allowed inter-rater reliability between human graders to be reported alongside human-AI agreement, providing a conventional reliability benchmark and supporting stronger claims about the generalisability of the human-AI agreement finding. The single-rater design follows from the fact that the author is the sole subject-matter expert teaching CPIT285 at this institution; no colleague at the institution teaches the same material with sufficient depth to grade independently. External multi-grader validation, ideally including raters from outside the home institution, is a necessary next step for institutional generalisation of the agreement findings reported here.

Third, the RQ4 feedback accuracy coding was likewise performed by a single coder, and the same defensibility argument applies. The coding judgements are constrained by established correct answers rather than by interpretive preference, but a second coder would have allowed a kappa estimate of inter-coder agreement on the four-category scheme.

Fourth, the analysis involved a reconciliation procedure that warrants transparent acknowledgement. Fourteen grade adjustments were made during the comparison phase between human and AI grades on the agreement and consistency samples: seven cases where the AI grade was overridden as incorrect on review, six cases where the human grade was revised after re-examination, and one case where the original human grade was retained after re-examination despite AI disagreement. The reconciliation procedure is recorded in full in *Appendix F*. While reconciliation reflects standard practice in coded comparative analysis, readers should treat the reported reliability statistics as post-reconciliation values, not as the raw output of either human or AI grading.

Fifth, the Claude Code grading sessions used Claude Opus 4.6 at the xhigh effort level. Other Claude Code configurations (e.g., lower effort levels, different models in the Claude family, smaller fine-tuned alternatives) and other agentic deployments would likely produce different reliability profiles, with weaker configurations expected to show wider error distributions. The technical configuration documented in Section 3.3 should be treated as the empirical test condition rather than as a recommended general configuration.

Sixth, this study reports reliability findings only and does not establish the validity of AI grading. An AI system that achieves strong reliability against a domain-expert reference may nevertheless score the wrong construct (a construct-validity question) or have unintended consequences for student learning, motivation, or progression (a consequential-validity question). Construct validity in this context would require independent evidence that the AI's grading judgements track the same conceptual targets that the rubric encodes, beyond surface agreement on numerical scores. Consequential validity would require evidence about how AI grading affects learner outcomes downstream of the grade itself. Both questions are addressed in the companion papers described in Section 6. The findings reported here should not be taken to support institutional deployment decisions on their own; reliability is one input to the validity argument, not its substitute

Taken together, these limitations situate the present study as evidence from a single Saudi Arabian institution, a single course in computer graphics, and an Arabic-English mixed-language student population. These features should be tested in other contexts before institutions generalise the findings to their own assessment settings.

## 6. Conclusion

This study examined the reliability of Claude Code as an agentic grading system for handwritten university exam papers in a Saudi Arabian computer graphics course. Across four research questions and three reliability dimensions, the findings describe a system whose measured reliability sits at the upper end of the published range for LLM grading, while remaining visibly limited on items requiring visual interpretation. AI grading produced strong agreement with human grading on total scores ($r = 0.97$), excellent test-retest consistency across two grading sessions one week apart ($ICC = 0.96$), and high accuracy on AI-generated feedback (92.8% on diagnostic feedback for items where students lost marks, and 97.8% on confirmation feedback for items where students earned full credit). A consistent gradient appeared across reliability dimensions: multiple-choice items were strongest on every measure, and drawing items were the clear outlier on three of four reliability measures (test-retest consistency, diagnostic feedback accuracy, and confirmation feedback accuracy), with the kappa measure flattening this gradient because it is less sensitive to absolute scale than the other three measures. A small but statistically detectable systematic bias of 0.68 marks (3.4% of the 20-mark scale), with the AI grading higher than the human grader, was reproduced across two independent grading sessions.

The methodological contribution of the study lies in the combination of conditions tested. To the best of the author's knowledge, the agentic Claude Code deployment with three structured rubric files and a single start instruction is not a configuration that prior peer-reviewed reliability studies have examined. The handwritten Arabic-English mixed-format input is closer to the conditions of a real university exam than the typed-text inputs that account for most of the published evidence base. The dual-function feedback accuracy coding scheme, distinguishing diagnostic from confirmation feedback, made visible a capability of the system (substantive elaboration of correct reasoning) that simpler accuracy-only schemes would have collapsed into a single number. The four structural safeguards described in Section 3.7 illustrate one workable design for deploying AI grading without exposing students to academic harm.

For practice, the question-type gradient is the most actionable finding. Multiple-choice items appear ready for AI grading with light human oversight on boundary cases. Written, analytical, and evaluation items are acceptable candidates with normal review. Drawing items require human review of every paper at the present state of the technology. The systematic bias means that human-side calibration is necessary for institutional deployment, and the divergence between intra-AI consistency and AI-human accuracy means that test-retest reliability alone cannot be the only validation step. AI grading at this level of capability is reliable enough to assist instructors but not reliable enough to replace them, particularly at the boundaries where grading decisions are most consequential for students.

Three companion papers will extend the present findings into the validity domain that this paper does not addressThe first builds an argument-based validity case on the same grading data, asking whether AI-human agreement varies with cognitive demand and whether AI credit tracks the wording of answers beyond the human grade. The second reports on student acceptance of AI grading, using pre- and post-survey instruments grounded in established acceptance, justice, and feedback-quality frameworks; this evidence speaks to consequential validity by examining how recipients perceive AI-graded outcomes. The third proposes the Recourse-Conditional Acceptance Model (RCAM) as a theoretical framework for AI acceptance in consequential algorithmic judgment contexts. The data and analysis from the present study will provide empirical grounding for each. Beyond this research programme, the findings suggest three directions for future work: cross-institutional replication of the agentic deployment configuration, evaluation of newer Claude versions and alternative agentic frameworks, and dedicated investigation of the drawing-item limitation through prompt engineering, fine-tuning, or alternative input pipelines for hand-drawn diagrams.

The broader question that this study addresses is whether AI grading is ready for deployment in higher education. The answer the data support is partial and conditional. It is ready on objective and short-form items, with structural safeguards in place and human review of papers near grade boundaries and of all drawing items, in a configuration like the one tested here. It is not ready as an unsupervised replacement for instructor judgment. The four structural safeguards make the difference between a deployment that protects students and one that does not. The decision facing institutional adopters is not whether to use AI grading but how to use it within a framework that preserves the instructor's responsibility to the student.

## Declarations

**Acknowledgements**

The author thanks the sixty students who consented to have their mid-term examination papers graded by Claude Code and whose participation made this reliability study possible. Their willingness to engage with a novel grading procedure, with full knowledge of their right to request manual regrading at any time, was the foundation of the study.

**Ethics Statement**

The study was conducted in accordance with the Law of Ethics of Research on Living Creatures (Royal Decree No. M/59 of 14/09/1431 H) and its Implementing Regulations issued by the National Committee of Bioethics, Kingdom of Saudi Arabia. The study falls within the minimal-risk category recognised under Article 10.33 of the Implementing Regulations, which covers research on educational tests in which information is recorded without revealing participant identity and which presents no more than minimal risk. Anonymisation procedures, the consent process, the unconditional regrading right, and the four structural safeguards that protected participating students from any academic disadvantage are described in full in Section 3.7. Written informed consent was obtained from all participating students through a bilingual Arabic–English consent form prior to any AI grading, and consent was treated as withdrawable at any time without justification or penalty. The bilingual consent form is reproduced in Appendix D.

**Competing Interests**

The author declares no financial competing interests. The author received no payment, gift, sponsorship, or consultancy from Anthropic (developer of Claude Opus 4.6 and Claude Code) or from any other commercial party with an interest in the outcome of this study. Use of Claude Code was on the standard commercial terms available to any subscriber. The author teaches the CPIT285 course in which the study was conducted; this dual role as instructor and investigator is acknowledged in Section 3.7, where the structural safeguards designed to prevent the dual role from affecting students' academic outcomes are described.

**Data Availability**

The materials supporting the findings of this study are available in two locations: the manuscript appendices and a public repository.

The manuscript appendices include the documents a reviewer or reader may consult while reading the paper. These comprise the blank exam paper (*Appendix A*), the complete marking scheme used by both the human and AI grader (*Appendix B*), a sample AI-generated grade report and the corresponding flag report for one consenting student (*Appendices C1 and C2*), the bilingual consent form (*Appendix D*), the per-question reliability statistics that supplement the per-type tables in Section 4 (*Appendix E*), and the full reconciliation log of the 19 grade adjustments documented in Section 3.6 (*Appendix F*).

Materials supporting full computational reproducibility are openly available at the Open Science Framework (OSF) repository at https://doi.org/10.17605/OSF.IO/75T6A under a Creative Commons Attribution 4.0 International licence (CC BY 4.0). The repository contains the three rubric files provided to Claude Code (`grading_rules.md`, `marking_scheme.md`, `output_schema.md`), the four Q4 reference images used in the drawing-comparison protocol, the Claude Code configuration file (`.claude/settings.json`), the procedural records of both grading sessions, the Python scripts used for grade-report delivery and PDF conversion, the anonymised grade reports and flag reports for all consenting students from Run 1 and for the consistency-test sample from Run 2, the master grades CSV files for both AI runs, the post-reconciliation human grades file used in the analysis, the correction log documenting all 14 grade adjustments, the feedback coding file used for the RQ4 analysis, the R analysis scripts (with a Python equivalent provided as an alternative), the data dictionary, the SVG source files for all figures, and the run environment logs documenting the technical configuration of each grading session.

The following data are not publicly available, in accordance with the institutional ethics framework described in Section 3.7 and the consent agreement signed by participating students: the student roster linking anonymised codes (S001 to S060) to real student identities, the original scanned exam papers in any form (because handwriting can enable re-identification), the per-paper blind grading record from the human grader, any free-text survey responses that could enable stylistic re-identification of individual students, and any communications between the author and individual students regarding regrading. These restricted data are retained locally by the author. Requests for controlled access to restricted data may be directed to the corresponding author and will be considered on a case-by-case basis, subject to institutional approval and a data-use agreement.

**Use of AI**

*AI as the object of study* - The study evaluates the reliability of Claude Code (Anthropic), an agentic deployment of the Claude Opus 4.6 large language model, as a grading system for university exam papers. The AI grading is the empirical phenomenon under investigation. Claude Code's role in the study, including the configuration parameters, the rubric files provided as input, the start instruction, and the timing of grading sessions, is documented in full in Section 3 (Methodology). Claude Code performed the grading of 60 consenting students' handwritten exam papers in Run 1 (22 April 2026) and the consistency-test grading of 15 papers in Run 2 (29 April 2026). All AI-assigned grades passed through the four structural safeguards described in Section 3.7 before entering the official record, including instructor review of every grade report, instructor review of all flagged papers, a five-working-day student review window, and unconditional access to manual regrading on request. The complete configuration files, rubric files, AI-generated grade reports, and analysis scripts are openly available at the Open Science Framework repository, documented in the Data Availability Statement.

*AI as a tool during manuscript preparation* - The author used Claude Code as a writing and editorial assistant during preparation of this manuscript, in the manner that researchers commonly use such tools alongside reference managers, statistical packages, and grammar checkers. The assistant was used to review and refine language, improve clarity and flow of expression, and produce final-draft figure renderings from author-specified designs. The author drafted the research questions, designed the methodology, conducted the analyses, interpreted the findings, and wrote the manuscript content. Statistical procedures were chosen by the author against established methodological literature; the analysis code was reviewed against the published statistical methods cited in Section 3.6, and outputs were cross-validated.

The AI assistant did not access raw student data, did not generate empirical findings, and did not contribute to the interpretation of results. The intellectual content, the methodological decisions, the interpretation of findings, and the conclusions drawn are the responsibility of the author. The author has reviewed every output produced with AI assistance and accepts full accountability for the accuracy, originality, and integrity of the manuscript.

Appendix A

# Full exam paper

blank version

Kingdom of Saudi Arabia
King Abdulaziz University
Faculty of Computing and Information Technology
Information Technology Department

# CPIT-285 Computer Graphics

Second Semester - 2025-2026

Type of Assessment: Midterm Exam

Instructor: Prof Rayed AlGhamdi

Total Marks for this Assessment: 20

---

| | |
|---|---|
| Student Name | |
| Student ID | |
| Section | |
| Program | Information Technology |

Total Obtained Marks ___________ out of 20 Duration 90 Minute(s)

## Student Outcomes (SO) – Maximum Marks Contributed by this Assessment

| Student Outcome | Maximum Marks |
|---|---|
| SO1 | 12.00 |
| SO2 | 8.00 |
| **Total (SO)** | **20** |

Assessment Date: Wednesday, April 15, 2026

## Difficulty distribution — Bloom's Taxonomy

Sections are ordered by cognitive level from remembering to evaluating.

| Level | Share | Marks |
|---|---|---|
| L1–L2 · Remember — Sec A | 30% | 6 marks · Q1.1–Q1.5, Q2 |
| L3 · Apply — Sec B | 30% | 6 marks · Q3, Q4 |
| L4 · Analyse — Sec C | 20% | 4 marks · Q5, Q6 |
| L5 · Evaluate — Sec D | 20% | 4 marks · Q7 |

| 6 | 6 | 4 | 4 |
|---|---|---|---|
| Sec A · L1–L2<br>Remember & Understand<br>Q1.1–Q1.5 · Q2 | Sec B · L3<br>Apply<br>Q3 · Q4 | Sec C · L4<br>Analyse<br>Q5 · Q6 | Sec D · L5<br>Evaluate<br>Q7 |

**Lower order thinking · 30%**
Section A · 6 marks
Recall of definitions and concepts from lectures and labs.

**Higher order thinking · 70%**
Sections B, C & D · 14 marks
Apply, analyse, and evaluate in practical contexts.

Lower order (L1–L2) · 6 marks · 30% — Higher order (L3–L5) · 14 marks · 70%

## Student outcome mapping

**SO1 — ANALYSE & APPLY** — 12 marks · 60%
Analyze a complex computing problem and apply computing principles to identify solutions.
Q1.1 Q1.2 Q1.3 Q1.4 Q1.5 Q2 Q5 Q7

**SO2 — DESIGN, IMPLEMENT & EVALUATE** — 8 marks · 40%
Design, implement, and evaluate a computing-based solution to meet given computing requirements.
Q3 Q4 Q6

SO1 · 12 marks · 60% | SO2 · 8 marks · 40%

● SO1 — Analyse & apply computing principles ● SO2 — Design, implement & evaluate solutions

| Question | Content | Marks | SO | Bloom's level | Rationale |
|---|---|---|---|---|---|
| Q1.1 A | Vector graphics composition | 1 | SO1 | L1 Remember | *Computing principle* |
| Q1.2 A | OpenGL polygon function | 1 | SO1 | L1 Remember | *Computing principle* |
| Q1.3 A | RGB colour depth | 1 | SO1 | L1 Remember | *Computing principle* |
| Q1.4 A | Roughness map reflection | 1 | SO1 | L3 Apply | *Apply PBR principles* |
| Q1.5 A | Rasterization in pipeline | 1 | SO1 | L2 Understand | *Pipeline concept* |
| Q2 A | UV seams — unwrapping | 1 | SO1 | L2 Understand | *Apply graphics concept* |
| Q3 B | PBR greyscale maps | 3 | SO2 | L3 Apply | *Evaluate design choices* |
| Q4 B | Draw OpenGL primitives | 3 | SO2 | L3 Apply | *Implement output* |
| Q5 C | Hz vs FPS | 2 | SO1 | L4 Analyse | *Analyse computing systems* |
| Q6 C | Baking — sniper rifle | 2 | SO2 | L4 Analyse | *Evaluate solution fit* |
| Q7 D | Polygon count vs resolution | 4 | SO1 | L5 Evaluate | *Identify correct solution* |

⚠ IMPORTANT ⚠
يُمنع منعًا باتًا إدخال أو استخدام أي أجهزة إلكترونية داخل قاعة الاختبار
وتُعد حيازتها أو استخدامها محاولة غش تُطبَّق بشأنها العقوبات وفق أنظمة الجامعة
The electronic device inside the examination hall is strictly prohibited
Possession or use will be considered cheating and subject to disciplinary action

## Section A — Remembering & Understanding Questions / 6 points

**Q1. Circle the correct answer** (1 point)

**Q1.1.** Vector graphics are composed of
- a. Pixels
- b. Polygons
- c. Palette
- d. All the above

**Q1.2.** Which OpenGL functions allow you to draw a polygon? (1 point)
- a. GL_POINTS
- b. GL_LINES
- c. GL_TRIANGLES
- d. GL_LINE_STRIP

**Q1.3.** RGB true color model has ______ color depth (1 point)
- a. 24-bit
- b. 32-bit
- c. 64-bit
- d. None of the above

**Q1.4.** The figure shows an unwrapped cube on UV layout with a roughness map applied. Each face (1–6) has a different color. Based on the map, which option correctly describes the faces' reflection behavior? (1 point)

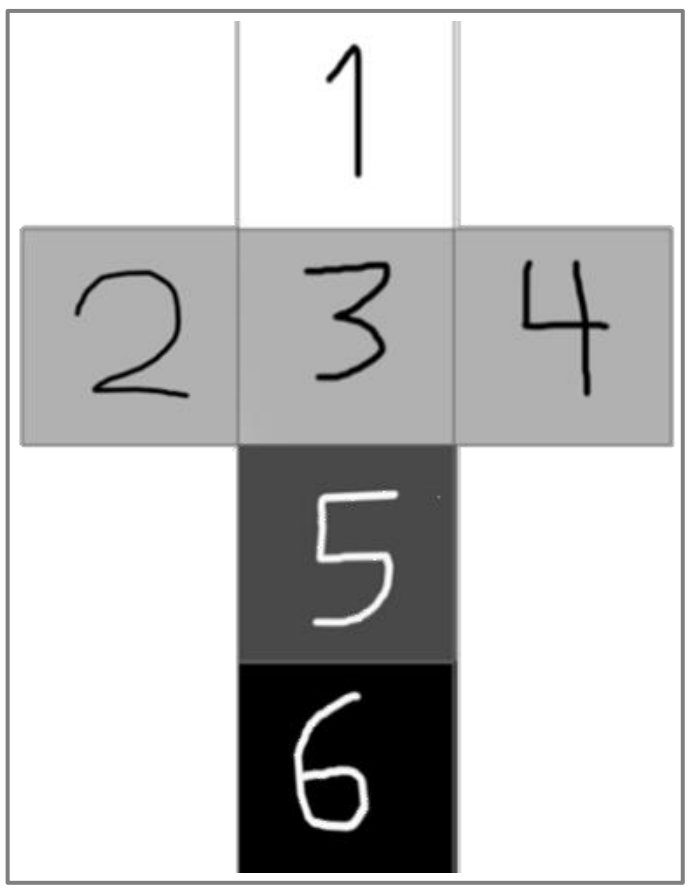

- a. Face (1) is least reflective; faces (2–3–4) are moderate; face (5) is high; face (6) is most reflective
- b. Face (1) is most reflective; faces (2–3–4) are high; face (5) is moderate; face (6) is least reflective
- c. Face (1) is most reflective; faces (2–3–4) are moderate; face (5) is high; face (6) is least reflective
- d. None of the above

**Q1.5.** In the graphics pipeline, rasterization is the process of: (1 point)
- a. Converting pixels into primitives
- b. Mapping 3D objects into 2D fragments for display
- c. Calculating lighting and shading effects
- d. Rendering textures directly onto models

---

**Q2. In Lab 4, you worked with UV unwrapping on the Thoub model.** Explain what a seam is in the context of UV unwrapping, and describe what role it plays in converting a 3D mesh into a flat 2D layout. (1 point)

*Answer using clear bullet points and key terms*

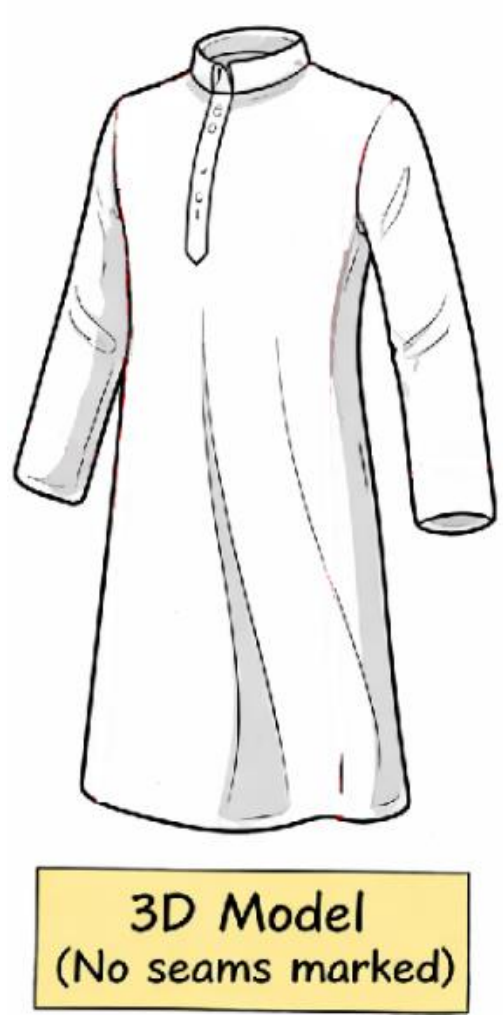

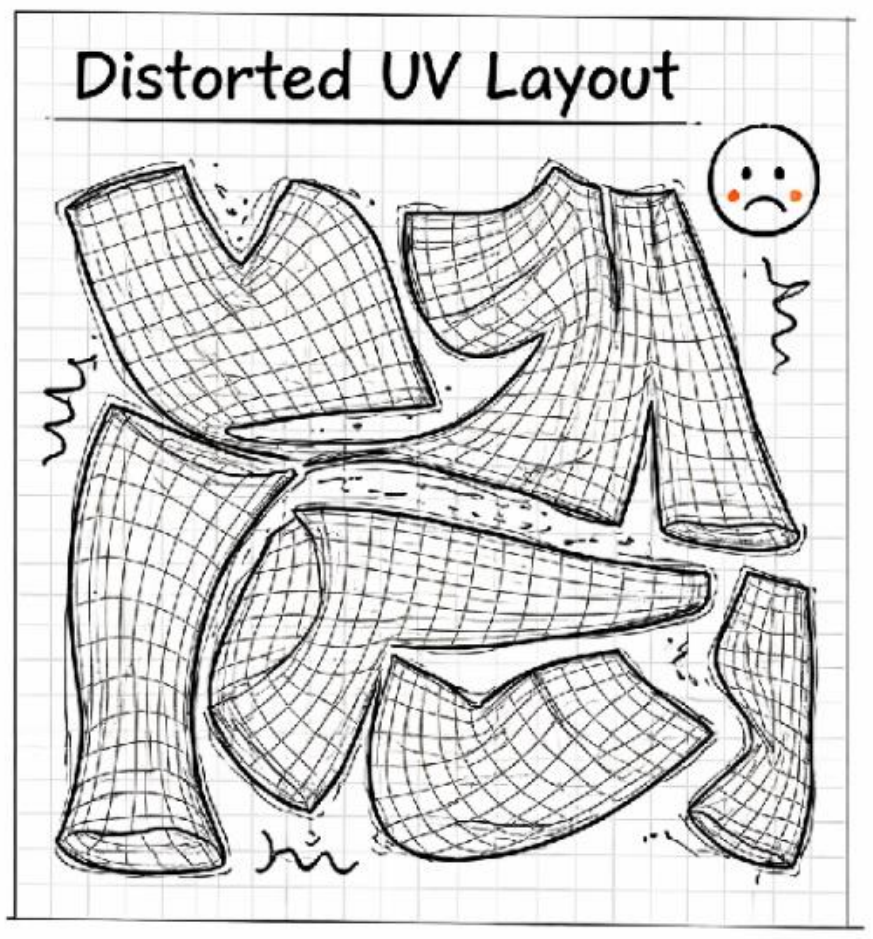

## Section B — Applying Questions / 6 points

**Q3.** PBR materials often use multiple texture maps, such as roughness, metallic, and height maps. Many of these maps appear as greyscale (black-and-white) images. (3 points total)

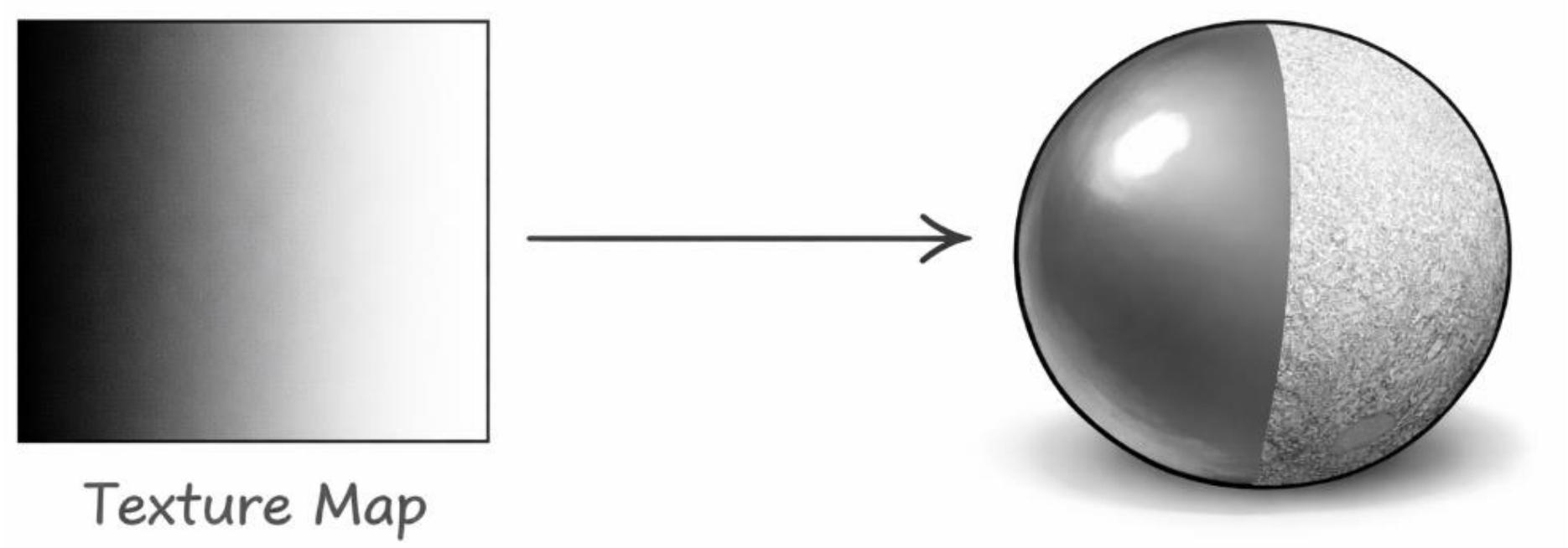

1. Explain what greyscale values represent in PBR maps (e.g., roughness or metallic). What do black and white values indicate? (1 point)
2. How does the renderer use these greyscale maps during shading to produce a realistic surface appearance? (1 point)
3. Why is using texture maps more effective than assigning a single constant value (e.g., one roughness value) to the entire object? (1 point)

*Answer using clear bullet points and key terms*

**Q4. Draw the output of running the following OpenGL functions.** Use the vertex coordinates below to draw the output of each OpenGL function in the box provided. (3 point)

V1=(0.5, 0.5, 0.0), V2=(0.2, 0.3, 0.0), V3=(0.2, 0.7, 0.0),

V4=(0.7, 0.7, 0.0), V5=(0.7, 0.3, 0.0)

| | |
|---|---|
| **(A) GL_LINES** | 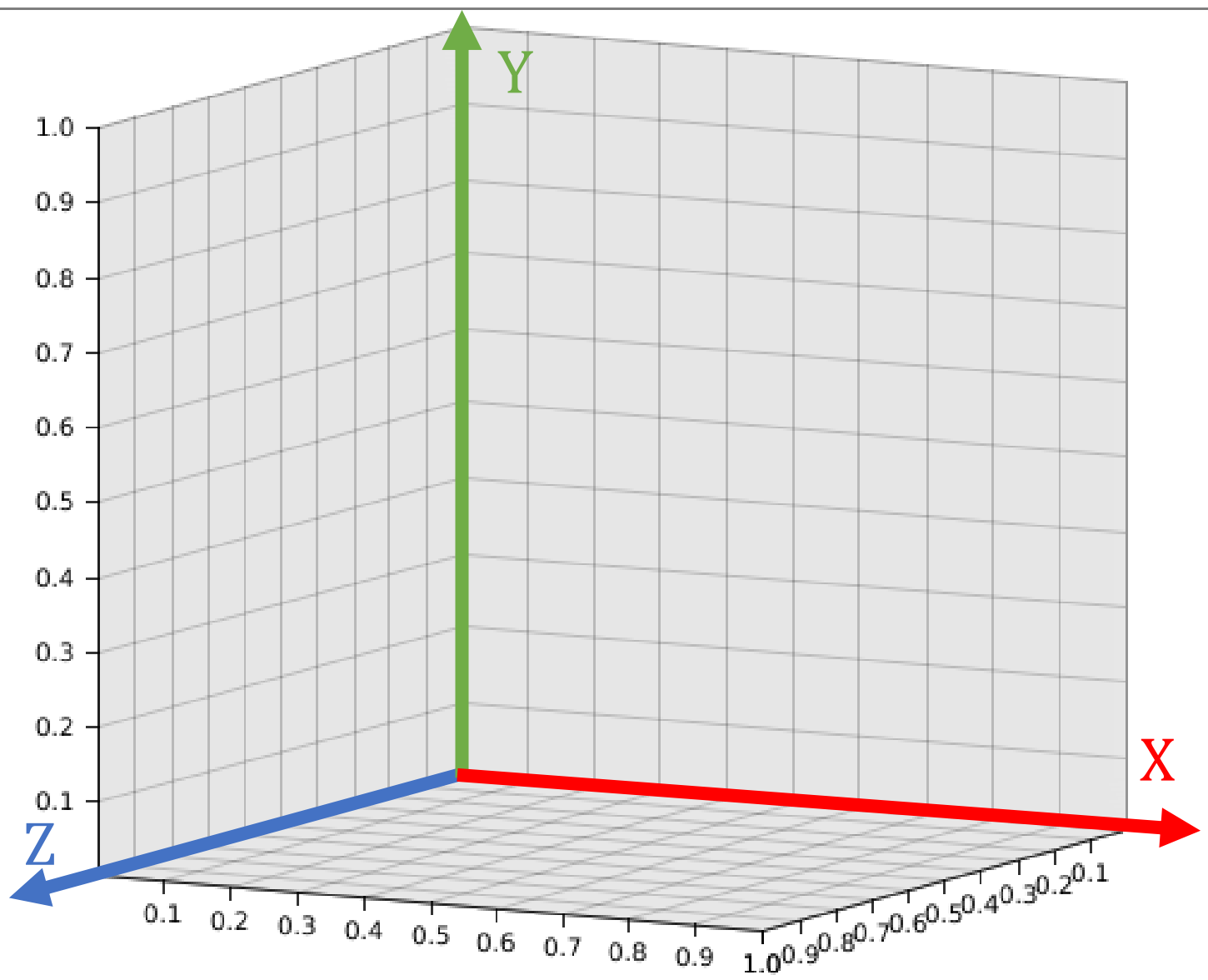 |
| **(B) GL_LINE_LOOP** | 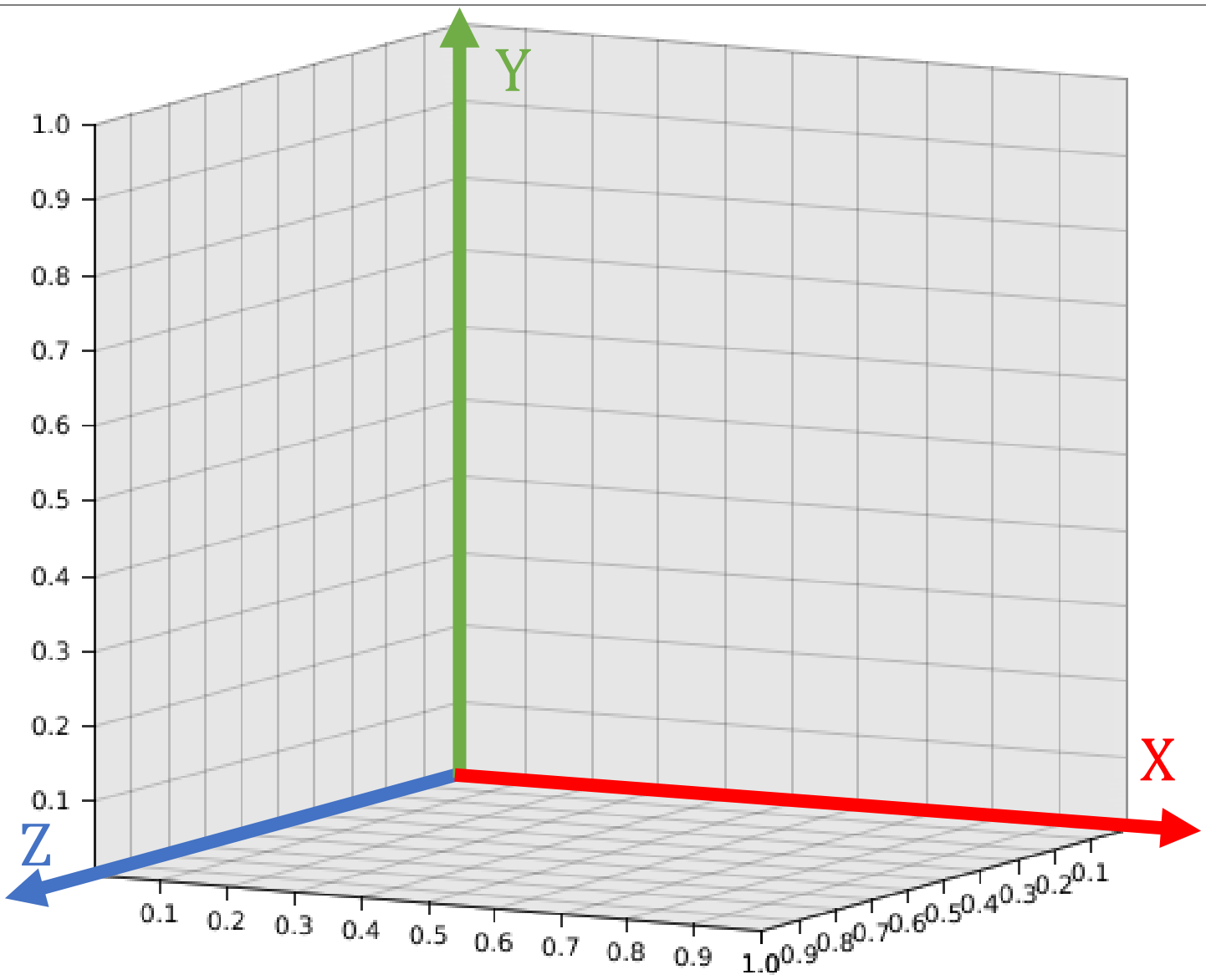 |
| **(C) GL_TRIANGLES** | 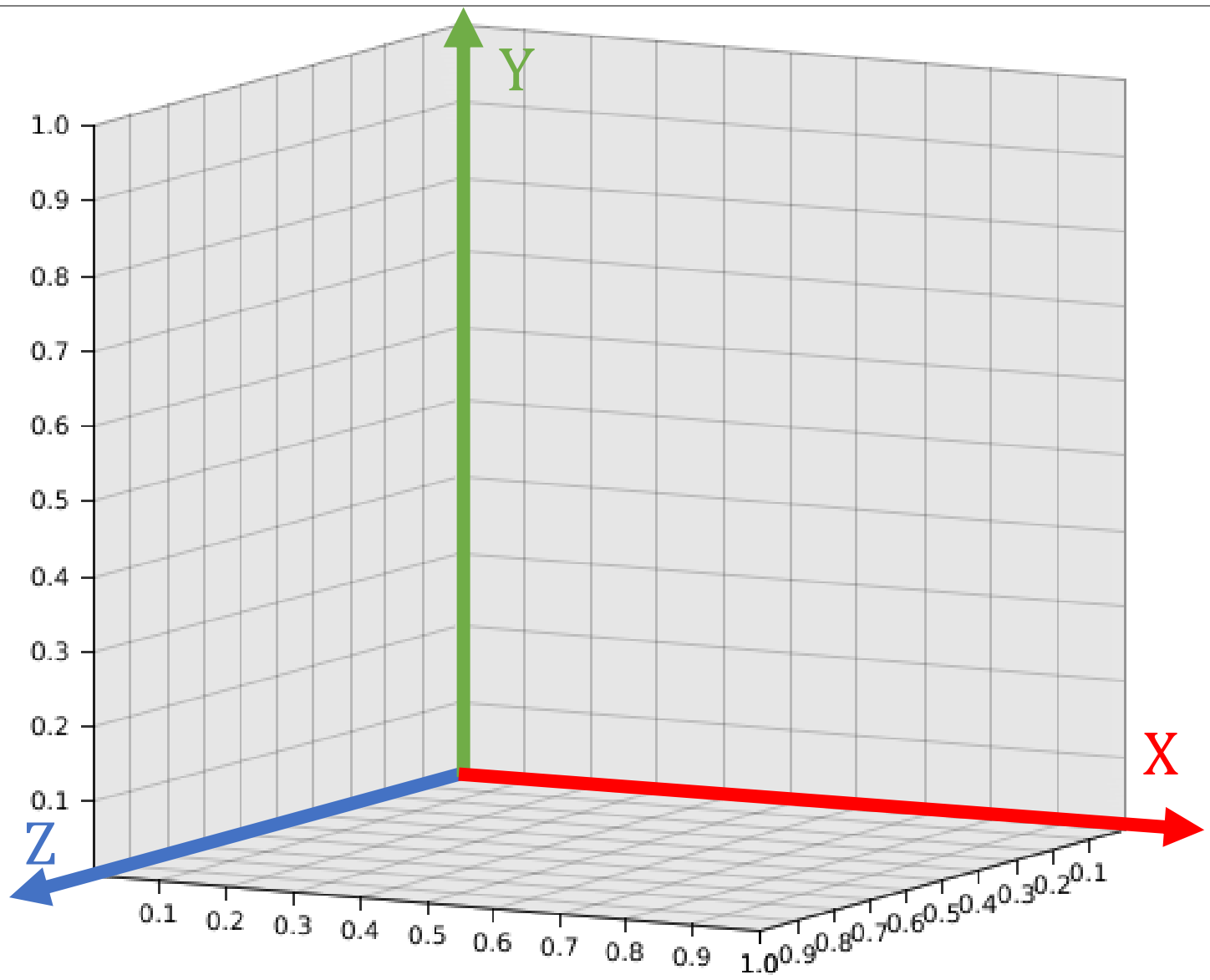 |

## Section C — Analyzing Questions / 4 points

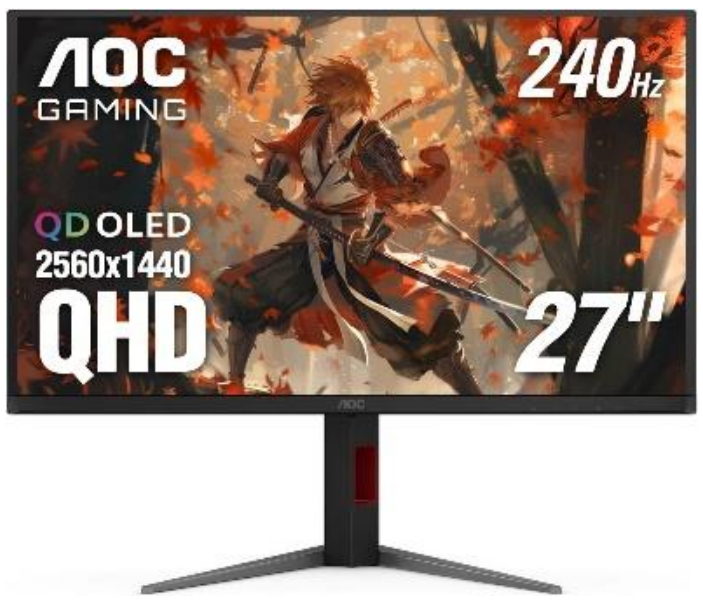

**Q5**. Ahmed buys a gaming monitor advertised as 240 Hz. His friend tells him he must also achieve 240 FPS in his game to "use the monitor fully." Explain whether this advice is correct. In your answer, clarify what Hz measures, what FPS measures, which component controls each, and what the relationship between the two values means for actual on-screen smoothness. (2 points)

*Answer using clear bullet points and key terms*

---

**Q6.** In the class, you were shown two versions of a sniper rifle model: a high-poly version with 373K polygons and a low-poly version with only 1542 polygons. **After baking**, **the low-poly sniper became visually almost identical to the high-poly one**. A student in the class says: *"They look the same, so baking just copied the shape and polygon detail from the high-poly into the low-poly."*

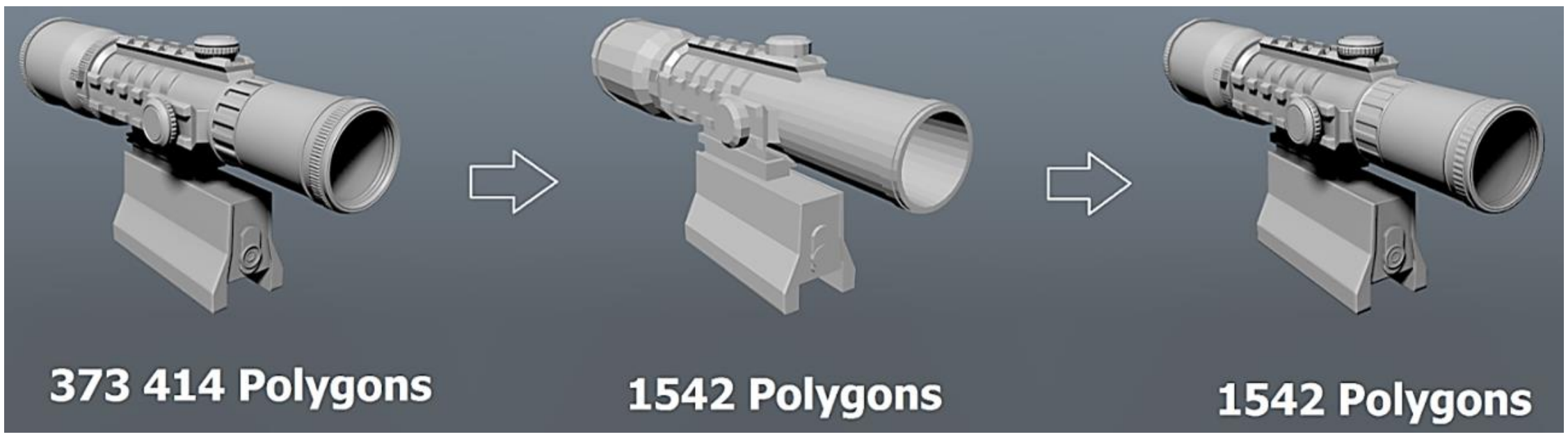

1. Explain precisely why this statement is wrong. (0.5 point)
2. What was actually transferred during baking? (0.5 point)
3. What remained fundamentally different between the two models? (0.5 point)
4. Why would a game engine struggle with the high-poly version? (0.5 point)

*Answer using clear bullet points and key terms*

## Section D — Evaluating Questions / 4 points

**Q7.** Two students are debating 3D objects made in Blender.

> **Student A** claims: *"You need a higher polygon count to get higher resolution. More polygons means better quality when you zoom in."*
>
> **Student B** disagrees. He opens a character model built from one million polygons and zooms into the face. The surface still looks blurry and low quality. Student B smiles and says: *"See? One million polygons, still blurry."* Student A cannot explain why.

**Answer all four parts:**

1. What point was Student B trying to prove? Was he correct? (1 point)
2. What does polygon count actually control — and why does it not affect sharpness when zooming in? (1 point)
3. Why did the one-million-polygon model still appear blurry? What was the actual limiting factor? (1 point)
4. Using the terms *vertices, edges, and faces*, explain in what sense 3D geometry behaves differently from a raster image, and which separate system does have a resolution limit. (1 point)

*Answer using clear bullet points and key terms*

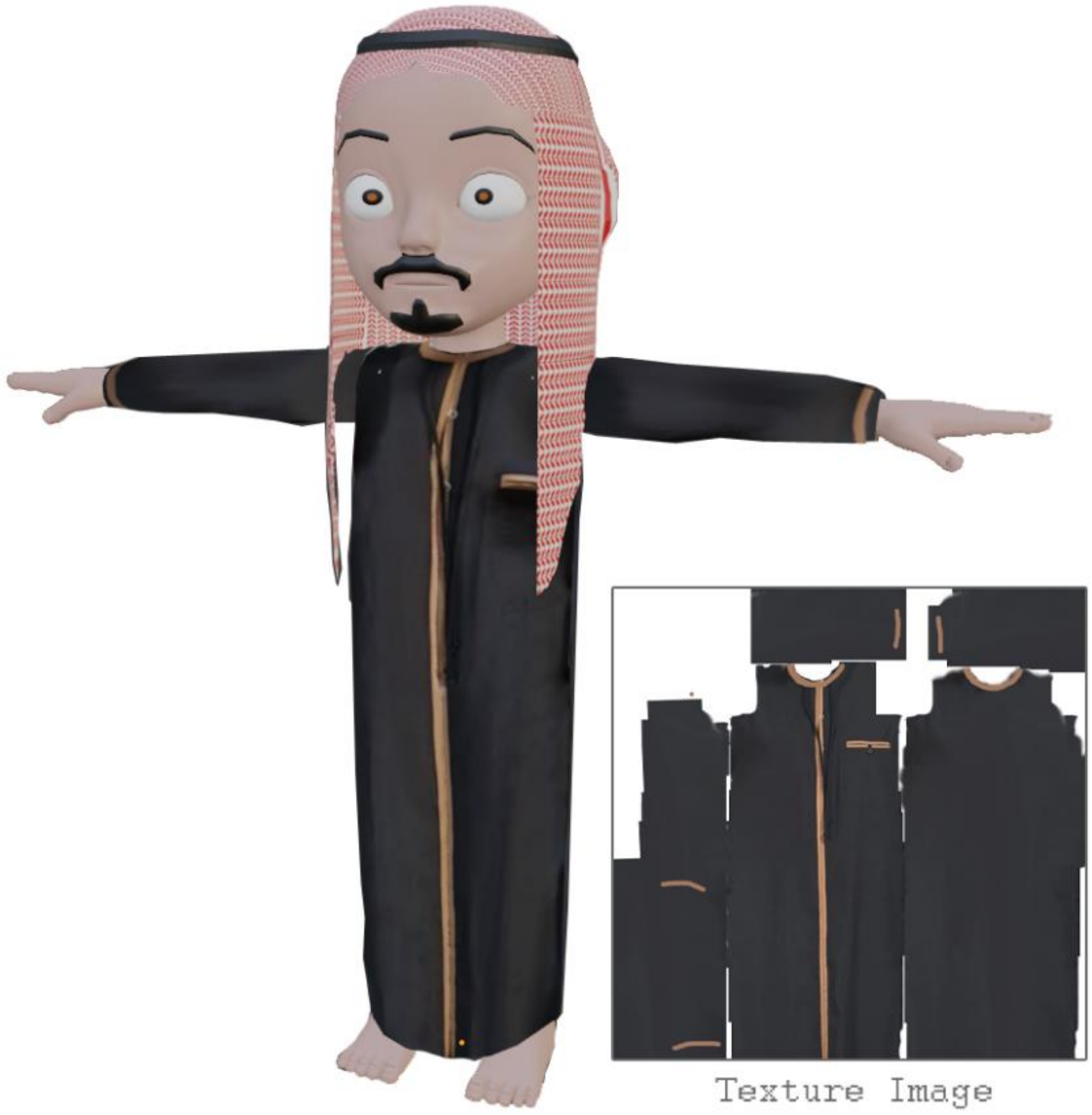

Appendix B

# Marking Scheme and Answer Key

CPIT285 Computer Graphics — Mid-Term Examination

# About This Appendix

This appendix presents the complete marking scheme and answer key used in the present study. The same document, in markdown format, was provided to Claude Code as one of the three rubric files at the start of every grading session, and the same answer key was used by the human grader during the blind grading protocol described in Section 3.4 of the manuscript.

The marking scheme specifies the model answer for each question, the partial-credit boundaries, the criteria for awarding zero, and accepted variant phrasings where relevant. It is presented here so that readers can evaluate the validity of the reliability findings reported in Section 4 against the assessment instrument actually used.

The exam totals 20 marks across four sections, aligned with Bloom's taxonomy levels L1 to L5 and the two computing student outcomes (SO1 and SO2) accredited by the Accreditation Board for Engineering and Technology.

## Exam Overview

| Section | Questions | Marks | Bloom's Level | Student Outcome |
|---|---|---|---|---|
| A — Remember and Understand | Q1.1–Q1.5, Q2 | 6 | L1–L2 | SO1 |
| B — Apply | Q3, Q4 | 6 | L3 | SO2 |
| C — Analyse | Q5, Q6 | 4 | L4 | SO1 / SO2 |
| D — Evaluate | Q7 | 4 | L5 | SO1 |
| TOTAL | Seven main questions | 20 | L1–L5 | SO1 = 12, SO2 = 8 |

The 30/70 distribution between lower-order (Section A) and higher-order (Sections B, C, D) thinking was deliberate, intended to balance recall of foundational graphics concepts with the analytical and evaluative reasoning expected at the 300-level course tier.

# Section A — Remember and Understand (6 marks)

Section A consists of five multiple-choice items (Q1.1 through Q1.5) and one short-written explanation (Q2). All items are aligned with Student Outcome 1 (computing principles) at Bloom's levels L1 (recall) and L2 (understanding).

### Q1.1 — Vector graphics composition (1 mark · SO1 · L1)

**Question**

Vector graphics are composed of: a) Pixels b) Polygons c) Palette d) All the above

**Correct answer: b) Polygons**

**Concept**

Vector graphics store shapes as mathematical descriptions, namely vertices, edges, and polygons (geometric primitives), rather than as pixel grids. This is the property that makes vector graphics resolution-independent.

**Common wrong answers**

- **a) Pixels —** describes raster or bitmap graphics, not vector.
- **c) Palette —** a colour lookup table, unrelated to graphic structure.
- **d) All the above —** incorrect; only polygons apply.

**Marking decision**

| | |
|---|---|
| **1** | Answer is b) |
| **0** | Any other answer |

*No partial marks on multiple-choice items.*

---

### Q1.2 — OpenGL polygon functions (1 mark · SO1 · L1)

**Question**

Which OpenGL function draws a filled polygon? a) GL_POINTS b) GL_LINES c) GL_TRIANGLES d) GL_LINE_STRIP

**Correct answer: c) GL_TRIANGLES**

**Concept**

Only GL_TRIANGLES produces a filled polygon primitive. GL_POINTS draws individual disconnected dots, GL_LINES draws disconnected line-segment pairs, and GL_LINE_STRIP draws connected but open and unfilled line segments. The word “filled” in the question is the key distinguisher; line-based functions produce outlines or segments, not filled shapes.

**Common wrong answer**

d) GL_LINE_STRIP — students may confuse a connected line sequence with a polygon. A polygon is a closed filled area, not an open line strip.

**Marking decision**

| | |
|---|---|
| **1** | Answer is c) |
| **0** | Any other answer |

## Q1.3 — RGB colour depth (1 mark · SO1 · L1)

### Question

The RGB true colour model has ________ colour depth: a) 24-bit b) 32-bit c) 64-bit d) None of the above

**Correct answer: a) 24-bit**

### Concept

RGB consists of three channels (Red, Green, Blue) at 8 bits per channel, totalling 24 bits. This permits $2^{24}$ = 16,777,216 distinct colours. Note that 32-bit includes an alpha (transparency) channel and is therefore RGBA, not pure RGB.

### Marking decision

| | |
|---|---|
| **1** | Answer is a) |
| **0** | Any other answer |

## Q1.4 — Roughness map reflection behaviour (1 mark · SO1 · L3)

### Question

The figure shows an unwrapped cube UV layout with a roughness map. Each face (1 to 6) has a different colour from brightest (face 1) to darkest (face 6). [The four answer options describe different combinations of which face is least and most reflective.]

**Correct answer: a) Face 1 is least reflective; faces 2-3-4 are moderate; face 5 is high; face 6 is most reflective**

### Core concept

On a roughness map, pixel values range from 0.0 (black/dark) to 1.0 (white/bright):

- Dark values (near 0) indicate low roughness, smooth surfaces, sharp reflections, and high reflectivity.
- Bright values (near 1) indicate high roughness, rough surfaces, diffuse light, and low reflectivity.

The mapping is therefore: darkest face (Face 6) = most reflective; brightest face (Face 1) = least reflective.

### Common errors

- Reversing the relationship (assuming white means reflective): award 0.
- Correct direction but wrong intermediate grouping: award 0 on multiple-choice items.

### Marking decision

| | |
|---|---|
| **1** | Answer is a) |
| **0** | Any other answer |

## Q1.5 — Rasterization in the graphics pipeline (1 mark · SO1 · L2)

**Question**

In the graphics pipeline, rasterization is the process of:

- a) Converting pixels into primitives
- b) Mapping 3D objects into 2D fragments for display
- c) Calculating lighting and shading effects
- d) Rendering textures directly onto models

**Correct answer: b) Mapping 3D objects into 2D fragments for display**

**Concept**

Rasterization is the pipeline stage that converts 3D geometric primitives (typically triangles) into 2D pixel fragments suitable for display on a screen. It comes after geometry processing and before fragment or pixel processing.

**Common wrong answers**

- **a)** reverses the direction; rasterization converts primitives to pixels, not pixels to primitives.
- **c)** lighting and shading are fragment processing tasks, not rasterization.
- **d)** texture mapping happens at fragment processing, not at rasterization.

**Marking decision**

| | |
|---|---|
| **1** | Answer is b) |
| **0** | Any other answer |

## Q2 — UV seams (1 mark · SO1 · L2)

**Question**

In Lab 4, you worked with UV unwrapping on the Thoub model. Explain what a seam is in the context of UV unwrapping, and describe what role it plays in converting a 3D mesh into a flat 2D layout.

**Model answer**

*A seam is a marked edge on a 3D mesh that defines where the surface is cut so it can be flattened into a 2D UV layout without overlapping. As a garment is cut along seams to lay the fabric flat, seams in 3D tell the software where to split the mesh. Without seams, the software has no guide for how to unfold the surface, resulting in a distorted or tangled UV layout that cannot be textured properly.*

**Key points required**

| Point | Worth | What to look for |
|---|---|---|
| A seam is a cut or split edge on the mesh | 0.5 | Student must describe it as a boundary, cut, or split, not just “a line” |
| It enables flattening of the 3D surface into 2D | 0.5 | Student must connect seams to the unwrapping or flattening purpose |

### Marking decision

| | |
|---|---|
| **1** | Both points clearly present |
| **0.5** | Only one point present, or both vaguely expressed |
| **0** | Only “seams mark where to texture” with no reference to flattening |

### Accepted phrasings

- ✓ *“Seams are where the mesh is cut”*
- ✓ *“Seams allow the 3D surface to be opened up flat”*
- ✓ *“Without seams the UV is distorted” (implies understanding of purpose)*
- ✓ *Arabic equivalent of any of the above*

### Not accepted

- ✗ *“Seams are lines on the model” (too vague, no connection to purpose)*
- ✗ *“Seams help with texturing” (does not explain the unwrapping role)*

---

# Section B — Apply (6 marks)

Section B consists of two questions: a three-part written explanation about physically-based rendering greyscale maps (Q3) and a three-part hand-drawn diagram task on OpenGL primitives (Q4). Both questions are aligned with Student Outcome 2 (design and implementation) at Bloom's level L3 (apply).

## Q3 — PBR greyscale maps (3 marks · SO2 · L3)

*Three sub-parts each worth 1 mark.*

### Q3.1 — What do greyscale values represent? (1 mark)

#### Model answer

*Greyscale values in PBR maps represent the intensity of a specific material property at each point on the surface. Black (value 0.0) means the minimum value of that property — for example, fully smooth on a roughness map, or fully non-metallic on a metallic map. White (value 1.0) means the maximum value — fully rough, or fully metallic. Each pixel independently controls the property at that exact location on the surface.*

#### Key points

- Black = minimum value of the property (0.0) — must be stated
- White = maximum value of the property (1.0) — must be stated
- Values control the property per pixel or per surface point — strengthens the answer

**Marking decision**

| | |
|---|---|
| **1** | Student correctly explains black AND white meaning for at least one map type |
| **0.5** | Student explains only one direction (black OR white but not both) |
| **0** | Student says “black = dark colour, white = light colour” without property context |

---

## Q3.2 — How does the renderer use greyscale maps during shading? (1 mark)

**Model answer**

*During rendering, the system samples the greyscale map at each surface point and reads the pixel value as a number between 0 and 1. This value is fed directly into the lighting equation for that specific point. A roughness value of 0.0 causes sharp, bright specular reflections; a value of 1.0 causes scattered, diffuse light. Different regions of the same surface therefore behave differently under the same light source, producing realistic variation.*

**Key points**

- Renderer samples the map per pixel or texel at each surface point — must be present
- Value feeds into the lighting or shading calculation at that point — must be present
- Different surface points behave differently under the same light — demonstrates understanding

**Marking decision**

| | |
|---|---|
| **1** | Student explains sampling AND effect on lighting calculation |
| **0.5** | Student explains sampling OR effect on lighting but not the connection |
| **0** | Student only says “the map changes how the surface looks” with no mechanism |

---

## Q3.3 — Why are maps more effective than a single constant value? (1 mark)

**Model answer**

*A single constant value applies the same property uniformly across the entire surface. Every part of the object behaves identically under light, which is unrealistic. Real objects have varying surface properties across different regions. Texture maps allow each pixel of the surface to have its own independent value, reproducing the natural variation of real-world materials. For example, human skin is shinier on the nose and rougher on the cheeks; this cannot be represented with one number.*

**Key points**

- Single value = uniform across entire surface = unrealistic — must be present

- Maps allow per-pixel or spatially varying values — must be present
- Any concrete real-world example — strengthens but not required for full mark

**Marking decision**

| | |
|---|---|
| **1** | Student explains uniformity problem AND per-pixel solution |
| **0.5** | Student explains only one side (the problem OR the solution) |
| **0** | Student says "maps look better" without explaining why |

## Q4 — Draw OpenGL primitives (3 marks · SO2 · L3)

*Three drawings each worth 1 mark.*

**Vertex coordinates**

V1 = (0.5, 0.5, 0.0)   V2 = (0.2, 0.3, 0.0)   V3 = (0.2, 0.7, 0.0)   V4 = (0.7, 0.7, 0.0)   V5 = (0.7, 0.3, 0.0)

All z-values are zero. Drawings are in the XY plane.

### Reference Image Comparison Protocol for Q4

Before grading any student drawing in Q4, the grader (human or AI) loads and studies four reference images stored in the rubric folder:

| File | What it shows |
|---|---|
| Q4_vertices_reference.png | All five vertices plotted at correct positions, no connections |
| Q4A_correct_GL_LINES.png | Correct output for GL_LINES |
| Q4B_correct_GL_LINE_LOOP.png | Correct output for GL_LINE_LOOP |
| Q4C_correct_GL_TRIANGLES.png | Correct output for GL_TRIANGLES |

These four reference images are reproduced in Appendix C.

**Grading procedure for each drawing**

1. Load the corresponding reference image.
2. Load the student's drawing from the exam paper scan.
3. Compare the two visually using the three criteria below.
4. Award marks based on the comparison.

**Three assessment criteria**

| Criterion | What to check |
|---|---|
| Vertex positions | Are the vertices plotted at approximately the correct locations on the grid? Compare to the vertices reference image. |
| Connections | Do the lines or shapes connect the correct vertices in the correct order? |

| Criterion | What to check |
|---|---|
| Primitive behaviour | Does the drawing demonstrate correct understanding of what this specific OpenGL function does? |

**Standard marking decision for Q4 drawings**

| | |
|---|---|
| **1** | All three criteria substantially correct compared to the reference image |
| **0.5** | Vertex positions correct but connections show misunderstanding of primitive behaviour |
| **0** | Primitive behaviour fundamentally misunderstood regardless of vertex positions |

*Scan quality flag rule: If the student's drawing area is cut off, partially obscured, or too faint to assess, the case is flagged for human review and the student is given the benefit of the doubt for anything partially visible. Scan quality is never a basis for penalising the student.*

## Q4(A) — GL_LINES (1 mark)

### What the correct drawing shows

- Two separate, disconnected line segments.
- Segment 1: V1 (0.5, 0.5) to V2 (0.2, 0.3).
- Segment 2: V3 (0.2, 0.7) to V4 (0.7, 0.7).
- V5 is plotted as a dot but not connected to anything; it is discarded.
- The two line segments do not touch each other.

### Why V5 is discarded

GL_LINES processes vertices in pairs. V1-V2 forms pair 1, V3-V4 forms pair 2. V5 is the fifth vertex with no partner and is therefore discarded by the primitive.

### Most common error

The student connects all five vertices in a continuous strip. This demonstrates confusion between GL_LINES (pairs) and GL_LINE_STRIP (sequential). If the student's drawing shows continuous connection, award 0.

## Q4(B) — GL_LINE_LOOP (1 mark)

### What the correct drawing shows

- All five vertices connected sequentially: V1 to V2 to V3 to V4 to V5.
- Plus a closing segment back to the first vertex: V5 to V1.
- Result: a closed five-sided outline shape (pentagon-like).
- No fill; outline only.
- The closing segment V5 to V1 must be clearly visible.

**Key distinction from GL_LINE_STRIP**

GL_LINE_STRIP draws V1 to V2 to V3 to V4 to V5 (open, no closing line).

GL_LINE_LOOP draws V1 to V2 to V3 to V4 to V5 to V1 (closed, includes closing line).

Compare to the reference image. If the closing line V5 to V1 is absent, award 0.5 rather than 0, because the student understood sequential connection but missed the closure.

---

### Q4(C) — GL_TRIANGLES (1 mark)

**What the correct drawing shows**

- One triangle only: V1 (0.5, 0.5), V2 (0.2, 0.3), V3 (0.2, 0.7).
- The triangle may be filled or outlined; both are acceptable.
- V4 and V5 are plotted as dots but not connected; they are discarded.
- No second triangle and no additional shape.

**Why V4 and V5 are discarded**

GL_TRIANGLES processes vertices in groups of three. V1-V2-V3 forms triangle 1. V4-V5 is only two vertices, an incomplete group, and is discarded by the primitive.

**Most common error**

The student uses all five vertices to draw a larger shape. Compare to the reference image. If V4 or V5 are connected to anything, award 0.5 maximum. If the student drew two separate triangles (misunderstanding the vertex count), award 0.

---

# Section C — Analyse (4 marks)

Section C consists of two analytical questions: a contrast between monitor refresh rate and frame rate (Q5) and a four-part analysis of the baking workflow used in real-time game asset preparation (Q6). Q5 is aligned with Student Outcome 1; Q6 is aligned with Student Outcome 2. Both are at Bloom's level L4 (analyse).

### Q5 — Hz vs FPS, Ahmed's monitor (2 marks · SO1 · L4)

**Question**

Ahmed buys a 240 Hz gaming monitor. His friend says he must achieve 240 FPS to “use the monitor fully.” Is this advice correct?

**Model answer**

*The friend's advice is partially correct but misleading. Hz measures how many times per second the monitor physically refreshes its display, a hardware property of the monitor controlled by the screen itself. FPS measures how many frames per second the GPU renders, a property controlled by the GPU and the game's graphical demands. To fully utilise a 240 Hz monitor, the GPU ideally needs to produce 240 or more frames per second. If FPS is below 240, the monitor repeats frames between refreshes, wasting some of its capability. If FPS exceeds 240, the extra frames are never displayed and screen tearing may occur. The two are complementary: Hz sets the ceiling of what can be shown; FPS determines how much of that ceiling is actually reached.*

*The advice is correct in direction but overstated, because the monitor still functions below 240 FPS, just not at full potential.*

**Marking — two points totalling 2 marks**

**Point 1 (1 mark): Correct and distinct definitions of both Hz and FPS**

- Hz = monitor refresh rate, how many times the screen redraws per second, a property of the monitor hardware.
- FPS = GPU render rate, how many frames the GPU produces per second, a property of the GPU and the game.

**Both must be correctly defined for 1 mark.**

| | |
|---|---|
| **1** | Both Hz and FPS correctly defined |
| **0.5** | Only one correctly defined |
| **0** | Student says “Hz and FPS are the same thing” |

**Point 2 (1 mark): Correct explanation of the relationship**

- FPS caps what the monitor can display: if FPS < Hz, potential is wasted.
- If FPS > Hz, extra frames are not shown and tearing may occur.
- The friend's advice is partially correct but overstated; the monitor still works below 240 FPS.

| | |
|---|---|
| **1** | Student captures the directional relationship AND notes the advice is not entirely correct |
| **0.5** | Student says the relationship is correct without nuance |
| **0** | Student says the advice is completely correct OR completely wrong |

---

## Q6 — Baking, sniper rifle scenario (2 marks · SO2 · L4)

*Four sub-parts each worth 0.5 marks.*

**Context provided to students**

A high-poly sniper model contains 373,414 polygons. After baking, the low-poly version contains 1,542 polygons. Both look visually identical. A student says: “Baking just copied the shape and polygon detail from the high-poly into the low-poly.”

### Q6.1 — Why is the statement wrong? (0.5 marks)

**Model answer**

*Baking did not copy any geometry or polygons. The low-poly sniper still has only 1,542 polygons after baking; the mesh structure is unchanged. The statement is wrong because it confuses visual appearance with geometric structure.*

| | |
|---|---|
| **0.5** | Student clearly states no geometry/polygons were copied; mesh unchanged |
| **0** | Student does not explain why; just says “it is wrong” |

### Q6.2 — What was actually transferred during baking? (0.5 marks)

**Model answer**

*The surface detail information from the high-poly model was captured and saved as a normal map, a 2D texture image. Specifically, the surface normals (the direction each point faces relative to light) were encoded as RGB colour values in a texture file. No geometry moved; only lighting and surface direction information was transferred into a texture.*

| | |
|---|---|
| **0.5** | Student correctly identifies a normal map or surface normal/lighting information |
| **0.25** | Student says “texture” or “image” without specifying the type |
| **0** | Student says “the shape was copied” or confuses geometry with texture |

**Accepted variations for full mark**

- ✓ *“Normal map was baked”*
- ✓ *“Surface details saved as a texture” (if context makes clear it is lighting information)*
- ✓ *“Light interaction information transferred to texture”*

### Q6.3 — What remained fundamentally different? (0.5 marks)

**Model answer**

*The polygon count and actual geometry remained fundamentally different. The high-poly model has 373,414 real polygons defining its surface shape. The low-poly has only 1,542. The visual similarity is an illusion created by the normal map; the underlying 3D mesh structure of the two models is completely different.*

| | |
|---|---|
| **0.5** | Student identifies polygon count, geometry, or mesh structure as what differs |
| **0** | Student cannot identify what differs, or says “nothing is different” |

### Q6.4 — Why would a game engine struggle with the high-poly version? (0.5 marks)

**Model answer**

*A game engine must process, transform, and render every polygon in real time, typically 60 or more times per second. At 373,414 polygons, the GPU performs millions of calculations per second for a single object. With multiple objects, animations, physics, and lighting all running simultaneously, this computational load becomes unmanageable and the game slows or crashes. The 1,542-polygon model achieves the same visual result at a fraction of the processing cost.*

| | |
|---|---|
| **0.5** | Student explains real-time processing cost of polygons + performance impact |
| **0.25** | Student says “it uses more resources” without explaining real-time cost |
| **0** | Student says the high-poly version looks worse, or cannot identify the issue |

---

# Section D — Evaluate (4 marks)

Section D consists of a single four-part evaluative question (Q7) at Bloom's level L5, the highest cognitive level in the exam. The question presents a debate between two students about the relationship between polygon count and visual sharpness when zooming, and asks the student to evaluate which position is correct and why.

### Q7 — Polygon count vs resolution debate (4 marks · SO1 · L5)

*Four sub-parts each worth 1 mark.*

**Scenario presented to students**

Student A claims that more polygons mean higher resolution when zooming in. Student B opens a 1-million polygon model and demonstrates that it is still blurry when zoomed in.

---

### Q7.1 — What was Student B proving? Was he correct? (1 mark)

**Model answer**

*Student B was proving that polygon count does not determine visual sharpness or “resolution” when zooming in. He is correct in his conclusion. His demonstration showed that even one million polygons cannot prevent blurriness at close zoom. The reason the model was blurry was the low-resolution texture image, a choice Student B made deliberately to isolate and expose the real limiting factor, showing that polygon count and image sharpness are two separate systems.*

**Key points**

- Student B was proving: polygon count ≠ sharpness when zooming — must be stated
- Student B was correct in his conclusion — must be confirmed

| | |
|---|---|
| **1** | Both points present: what B was proving AND that he was correct |
| **0.5** | Student identifies what B was proving but is confused about whether B was right |
| **0** | Student says Student B was wrong — fundamental misunderstanding |

---

### Q7.2 — What does polygon count actually control? (1 mark)

**Model answer**

*Polygon count controls the smoothness and geometric detail of the 3D shape, namely how accurately curved surfaces are represented, whether silhouettes look smooth or faceted, and how finely detailed the mesh structure is. It has nothing to do with sharpness when zooming in because polygons do not store colour or surface texture information. The colour, detail, and sharpness visible when zooming come entirely from the texture image, a separate system with its own resolution.*

**Key points**

- Polygon count controls shape smoothness or geometric mesh detail — must be present
- Polygons do not store colour or texture information — must be present or implied
- Sharpness comes from a separate system (texture) — strengthens the answer

| | |
|---|---|
| **1** | Student explains what polygons control AND why this is separate from zoom sharpness |
| **0.5** | Student explains what polygons control but does not explain the separation from zoom sharpness |
| **0** | Student says "more polygons = more detail" without distinguishing shape detail from texture sharpness |

---

## Q7.3 — Why did the model still appear blurry? (1 mark)

**Model answer**

*The model appeared blurry because the texture image applied to it had low resolution. A texture image is a raster file: a fixed grid of pixels. When the camera zooms in, the same pixels are stretched across a larger screen area, causing them to appear as a blurry, blocky pattern. The actual limiting factor was the pixel resolution of the texture file, not the polygon count.*

**Key points**

- The texture image or texture resolution was the limiting factor — must be present
- Raster image pixels become visible when zoomed beyond texture resolution — explains the mechanism

| | |
|---|---|
| **1** | Student identifies texture resolution as limiting factor AND explains pixel stretching |
| **0.5** | Student identifies texture as the problem but does not explain the mechanism |
| **0** | Student says "the model needs more polygons to be sharp" |

---

## Q7.4 — Vertices, edges, faces vs raster, explain the difference (1 mark)

**Model answer**

*Vertices, edges, and faces in a 3D mesh are mathematical coordinates and relationships. A vertex is a point in 3D space (x, y, z); an edge connects two vertices; a face is a polygon bounded by edges. These have no inherent resolution. They exist as pure mathematics. You can zoom infinitely into a 3D mesh and the edges remain perfectly sharp because there are no*

*pixels to run out of. The system that does have a resolution limit is the texture map: a 2D raster image with a fixed number of pixels that pixelates when magnified beyond its resolution.*

**Key points**

- Vertices, edges, and faces are mathematical or coordinate-based, with no inherent resolution — must be present
- 3D geometry is resolution-independent (can zoom infinitely without loss) — must be stated or implied
- The texture map or raster image is the system that has a resolution limit — must be named

| | |
|---|---|
| **1** | All three points present: geometry is mathematical, geometry is resolution-independent, texture has the limit |
| **0.5** | Student identifies geometry as resolution-independent BUT fails to name texture as the limiting system |
| **0** | Student confuses vertices/edges/faces with pixels, or cannot articulate the distinction |

---

## Holistic grading note for Q7

Q7 is a Bloom's level 5 evaluative question, the highest cognitive level in the exam. A strong answer will clearly and consistently separate two distinct systems throughout:

- The 3D mesh (vertices, edges, faces, polygons) defines shape and has no resolution limit.
- The texture image (raster pixels) defines surface appearance and has a resolution limit.

*Students who conflate the two systems throughout their answer should not receive more than 2/4 regardless of how many correct keywords appear. Students who clearly understand the distinction and apply it to all four sub-parts deserve full or near-full marks even if their phrasing is imperfect.*

Appendix C1

# Sample Grade Report

The actual report sent to student

# Grade Report — S013

Course: CPIT285 Computer Graphics — Mid Exam Spring 2026
Graded by: AI-assisted grading system under instructor supervision
Date processed: 2026-04-22
Requires human review: Yes

## Score Summary

| Section | Questions | Max | Awarded |
|---|---|---|---|
| A — Remember & Understand | Q1.1, Q1.2, Q1.3, Q1.4, Q1.5, Q2 | 6 | 4 |
| B — Apply | Q3, Q4 | 6 | 2 |
| C — Analyse | Q5, Q6 | 4 | 3.25 |
| D — Evaluate | Q7 | 4 | 2.5 |
| **TOTAL** | | **20** | **11.75** |

## Section A — Remembering & Understanding (4 / 6)

### Q1.1 — 0 / 1 — Confidence: High

EN: You selected d) All the above, but the correct answer is b) Polygons. Vector graphics are composed of mathematical descriptions of shapes using polygons (geometric primitives), not pixels or palettes. Pixels describe raster/bitmap graphics, and palettes are colour lookup tables unrelated to graphic structure.

اخترت د) جميع ما سبق،
لكن الإجابة الصحيحة هي ب) المضلعات.
الرسوميات المتجهية تتكون من وصف رياضي للأشكال باستخدام المضلعات،
وليس البكسلات أو لوحات الألوان.

### Q1.2 — 1 / 1 — Confidence: High

EN: Correct. GL_TRIANGLES draws filled triangular polygons, which is the only option that produces a filled polygon primitive.

إجابة صحيحة.
GL_TRIANGLES ترسم مضلعات مثلثية مملوءة،
وهي الخيار الوحيد الذي ينتج مضلعا مملوءا.

### Q1.3 — 1 / 1 — Confidence: High

EN: Correct. RGB true colour uses 3 channels (Red, Green, Blue) at 8 bits each, totalling 24 bits.

إجابة صحيحة.
نموذج الألوان الحقيقية RGB يستخدم 3 قنوات بعمق 8 بت لكل قناة،
أي 24 بت إجمالا.

### Q1.4 — 0 / 1 — Confidence: High

EN: You selected b), but the correct answer is a). On a roughness map, darker values mean lower roughness (smoother, more reflective), while brighter values mean higher roughness (rougher, less reflective). Face 1 (brightest) is least reflective, and Face 6 (darkest) is most reflective.

اخترت ب)،
لكن الإجابة الصحيحة هي أ).
في خريطة الخشونة،
القيم الداكنة تعني خشونة أقل (سطح أملس وأكثر انعكاسا)،
بينما القيم الفاتحة تعني خشونة أعلى (أقل انعكاسا).
الوجه 1 (الأفتح) هو الأقل انعكاسا،
والوجه 6 (الأغمق) هو الأكثر انعكاسا.

### Q1.5 — 1 / 1 — Confidence: High

EN: Correct. Rasterization converts 3D geometric primitives into 2D pixel fragments for display on the screen.

إجابة صحيحة.
التحويل النقطي هو عملية تحويل الأشكال الهندسية ثلاثية الأبعاد إلى أجزاء بكسل ثنائية الأبعاد للعرض على الشاشة.

### Q2 — 1 / 1 — Confidence: High

EN: Excellent answer. You correctly described seams as marked edges that define where the mesh is cut, and you clearly connected this to the purpose of converting the 3D model into 2D pieces for texturing. Your mention of handling each piece individually shows understanding of the practical purpose.

إجابة ممتازة.
وصفت الخياطات بشكل صحيح كحواف محددة تعرف أين يقطع النموذج،
وربطت ذلك بوضوح بهدف تحويل النموذج ثلاثي الأبعاد إلى قطع ثنائية الأبعاد للنسيج.
إشارتك لمعالجة كل قطعة على حدة تظهر فهما عمليا.

---

## Section B — Applying Questions (2 / 6)

### Q3 — PBR Greyscale Maps (0.5 / 3)

#### Q3 Sub-part 1 — 0.5 / 1 — Confidence: Medium

EN: You mentioned that greyscale values represent an "ambient map" with black indicating lower parts (shadows) and white the upper part. This shows partial understanding that black and white represent different intensities, but the explanation confuses ambient occlusion with PBR property intensity. In PBR maps like roughness, black means minimum roughness (smooth) and white means maximum roughness (rough) — the values represent the intensity of a specific material property, not shadows.

ذكرت أن القيم الرمادية تمثل "خريطة محيطية" حيث الأسود يشير للأجزاء السفلية (الظلال) والأبيض للجزء العلوي.
هذا يظهر فهما جزئيا بأن الأسود والأبيض يمثلان شدة مختلفة،
لكن الشرح يخلط بين الإضاءة المحيطية وشدة خاصية المادة في PBR.
في خريطة الخشونة مثلا،
الأسود يعني أقل خشونة والأبيض يعني أعلى خشونة.

#### Q3 Sub-part 2 — 0 / 1 — Confidence: High

EN: Your answer "when adding the base map it should add to it to produce realistic surface appearance" is too vague. The correct explanation is that the renderer samples the greyscale map at each surface point and reads the pixel value as a number between 0 and 1, which is then fed into the lighting/shading calculation. This allows different regions of the same surface to behave differently under the same light source.

إجابتك "عند إضافة الخريطة الأساسية يجب إضافتها لإنتاج مظهر سطح واقعي" غامضة جدا.
الشرح الصحيح هو أن المصير يأخذ عينات من الخريطة الرمادية عند كل نقطة سطح ويقرأ قيمة البكسل كرقم بين 0 و1،
ثم يدخلها في حسابات الإضاءة والتظليل.

### Q3 Sub-part 3 — 0 / 1 — Confidence: High

EN: Your answer focused on resource consumption and GPU handling, which is not what the question asks. The key reason texture maps are more effective is that a single constant value applies the same property uniformly across the entire surface (unrealistic), while texture maps allow per-pixel variation — each point on the surface can have its own independent value, reproducing natural material variation.

إجابتك ركزت على استهلاك الموارد ومعالجة وحدة الرسوميات،
وهذا ليس ما يسأل عنه السؤال.
السبب الرئيسي هو أن القيمة الثابتة الواحدة تطبق نفس الخاصية بشكل موحد على كامل السطح (غير واقعي)،
بينما خرائط النسيج تسمح بتنوع على مستوى كل بكسل.

## Q4 — OpenGL Drawing (1.5 / 3)

### Q4(A) GL_LINES — 0 / 1 — Confidence: Medium

EN: Your drawing appears to show vertices connected in a continuous manner rather than two separate, disconnected line segments. GL_LINES processes vertices in pairs: V1-V2 forms one segment and V3-V4 forms another, with V5 discarded (no partner). The two segments should not be connected to each other.

رسمك يبدو أنه يظهر الرؤوس متصلة بشكل متتابع بدلا من قطعتي خط منفصلتين.
يعالج الرؤوس بأزواج GL_LINES:
قطعة و V1-V2 قطعة أخرى V3-V4،
مع تجاهل V5 لعدم وجود شريك.

### Q4(B) GL_LINE_LOOP — 1 / 1 — Confidence: High

EN: Good work. Your drawing correctly shows all 5 vertices connected sequentially with a closing line segment back to V1, forming a closed pentagon-like outline. This matches the expected GL_LINE_LOOP behaviour.

عمل جيد.
رسمك يظهر بشكل صحيح جميع الرؤوس الخمسة متصلة بالتتابع مع خط إغلاق يعود إلى V1،
مكونا شكلا مغلقا يشبه المخمس.
هذا يتطابق مع سلوك GL_LINE_LOOP المتوقع.

### Q4(C) GL_TRIANGLES — 0.5 / 1 — Confidence: Medium

EN: Your drawing shows a shape that appears to use more vertices than just V1-V2-V3. For GL_TRIANGLES, vertices are processed in groups of 3: only V1(0.5,0.5), V2(0.2,0.3), and V3(0.2,0.7) form a triangle. V4 and V5 are discarded because they form an incomplete group. The correct output is a single triangle.

رسمك يظهر شكلا يبدو أنه يستخدم رؤوسا أكثر من V1-V2-V3 فقط.
في GL_TRIANGLES،
تعالج الرؤوس بمجموعات من 3:
فقط V1 وV2 وV3 يشكلون مثلثا واحدا.
V4 وV5 يتجاهلان لأنهما مجموعة غير مكتملة.

---

## Section C — Analysing Questions (3.25 / 4)

### Q5 — Hz vs FPS (1.5 / 2) — Confidence: Medium

#### Q5 Point 1 — Hz and FPS definitions — 1 / 1

EN: You correctly defined Hz as the refresh rate of the monitor (how many times the monitor refreshes) and FPS as the frame rate produced by the GPU per second. Both definitions are accurate and you identified the correct component for each.

عرفت بشكل صحيح Hz كمعدل تحديث الشاشة (كم مرة تحدث الشاشة في الثانية) وFPS كمعدل الإطارات التي تنتجها وحدة معالجة الرسوميات في الثانية.
كلا التعريفين دقيقان وحددت المكون الصحيح لكل منهما.

#### Q5 Point 2 — Relationship and advice evaluation — 0.5 / 1

EN: You mentioned that Hz and FPS "work efficiently together," which shows some understanding of the relationship but lacks the critical nuance. The friend's advice is partially correct but overstated — the monitor still functions below 240 FPS, just not at full potential. If FPS is below Hz, the monitor repeats frames; if above, extra frames are wasted. Your answer needed to explain this directional relationship and note the advice is not entirely correct.

ذكرت أن Hz وFPS "يعملان معا بكفاءة"،
مما يظهر بعض الفهم للعلاقة لكنه يفتقر للتفاصيل الحاسمة.
نصيحة الصديق صحيحة جزئيا لكنها مبالغ فيها — الشاشة تعمل حتى لو كان FPS أقل من 240،
لكن ليس بكامل إمكاناتها.

### Q6 — Baking — Sniper Rifle (1.75 / 2)

#### Q6 Sub-part 1 — 0.5 / 0.5 — Confidence: High

EN: You correctly explained that baking is not copying geometry — it is a process of converting high-detail visual information to the low-poly model through an "illusion," with the mesh structure remaining unchanged.

شرحت بشكل صحيح أن الخبز لا ينسخ الهندسة — بل هو عملية تحويل المعلومات المرئية عالية التفصيل للنموذج منخفض المضلعات عبر "وهم"،
مع بقاء هيكل الشبكة دون تغيير.

#### Q6 Sub-part 2 — 0.25 / 0.5 — Confidence: Medium

EN: You mentioned "UV maps to make illusions" which indicates you understand something visual was transferred, but you didn't specifically identify a Normal Map or surface normal/lighting information. The correct answer is that surface normal direction data was captured and saved as a normal map texture.

ذكرت "خرائط UV لعمل أوهام" مما يشير لفهمك أن شيئا بصريا تم نقله،
لكنك لم تحدد خريطة الطبيعي أو معلومات إضاءة السطح تحديدا.
الإجابة الصحيحة هي أن بيانات اتجاه الطبيعي للسطح تم التقاطها وحفظها كخريطة نسيج طبيعي.

#### Q6 Sub-part 3 — 0.5 / 0.5 — Confidence: High

EN: Correct. You identified that the fundamental difference is the number of polygons between the two models.

صحيح.
حددت أن الفرق الجوهري هو عدد المضلعات بين النموذجين.

#### Q6 Sub-part 4 — 0.5 / 0.5 — Confidence: High

EN: Good answer. You explained that the high-poly version has a huge number of polygons requiring extensive complex mathematical operations, which would be too demanding for a game engine.

إجابة جيدة.

شرحت أن النسخة عالية المضلعات تحتوي على عدد ضخم من المضلعات تتطلب عمليات حسابية معقدة كثيرة،
مما يرهق محرك الألعاب.

## Section D — Evaluating Questions (2.5 / 4)

### Q7 — Polygon Count vs Resolution (2.5 / 4)

#### Q7 Sub-part 1 — 1 / 1 — Confidence: High

EN: Excellent. You clearly stated that Student B proved that the number of polygons is not equal to resolution, and you confirmed he was correct. This demonstrates a solid understanding of the core concept.

ممتاز.
ذكرت بوضوح أن الطالب B أثبت أن عدد المضلعات لا يساوي الدقة،
وأكدت أنه كان على صواب.
هذا يظهر فهما متينا للمفهوم الأساسي.

#### Q7 Sub-part 2 — 0.5 / 1 — Confidence: Medium

EN: You correctly stated that polygon count controls the number of details of the mesh and that it doesn't affect sharpness, noting that "the responsible is not num of polys." However, you didn't clearly distinguish between geometric shape detail (what polygons control) and texture sharpness (a separate system). The key point is that polygons define shape smoothness but don't store colour/texture information.

ذكرت بشكل صحيح أن عدد المضلعات يتحكم في تفاصيل الشبكة ولا يؤثر على الحدة.
لكنك لم تميز بوضوح بين تفاصيل الشكل الهندسي (ما تتحكم فيه المضلعات) وحدة النسيج (نظام منفصل).
النقطة الأساسية هي أن المضلعات تحدد نعومة الشكل لكنها لا تخزن معلومات اللون/النسيج.

#### Q7 Sub-part 3 — 0.5 / 1 — Confidence: Medium

EN: You mentioned "dpi (dots/pixel per inch)" as the limiting factor, which shows you understand the concept is related to pixel density/resolution rather than polygon count. However, the actual limiting factor is the texture image resolution — the raster texture file has a fixed number of pixels that become stretched and blurry when zoomed beyond its resolution. Identifying "texture resolution" specifically was needed for full marks.

ذكرت "dpi (نقاط/بكسل لكل بوصة)" كعامل محدد،
مما يظهر فهمك أن المفهوم مرتبط بكثافة البكسل/الدقة وليس عدد المضلعات.
لكن العامل المحدد الفعلي هو دقة صورة النسيج — ملف النسيج النقطي يحتوي على عدد ثابت من البكسلات تتمدد وتصبح ضبابية عند التكبير.

#### Q7 Sub-part 4 — 0.5 / 1 — Confidence: Medium

EN: You made a reasonable attempt at distinguishing 3D geometry from raster images, mentioning edges, vertices, and dimensions for 3D models versus pixel grids for raster. You also noted a resolution limit for the monitor. However, you needed to clearly state that vertices/edges/faces are mathematical and resolution-independent (can zoom infinitely), and that the texture map (raster image) is the separate system with a fixed resolution limit.

قدمت محاولة معقولة للتمييز بين الهندسة ثلاثية الأبعاد والصور النقطية،
وذكرت الحواف والرؤوس والأبعاد للنماذج ثلاثية الأبعاد مقابل شبكات البكسل للنقطية.
لكن كنت تحتاج لتوضيح أن الرؤوس/الحواف/الأوجه رياضية وغير محدودة الدقة،
وأن خريطة النسيج هي النظام المنفصل ذو الدقة المحدودة.

## Instructor Note

This grade report was generated by an AI-assisted grading system and reviewed by Dr. Rayed AlGhamdi before release. To request human re-evaluation of any question, contact: raalghamdi8@kau.edu.sa — no deadline, no penalty.

للطلب إعادة تصحيح يدوي لأي سؤال،
تواصل مع الدكتور رائد الغامدي على:
raalghamdi8@kau.edu.sa — بدون موعد نهائي وبدون أي عقوبة.

Appendix C2

# Sample Flag Report

The report presented to the instructor for review

# Flag Report

## Anonymised Paper ID: S013

**Requires instructor review before grade is finalised**

| | |
|---|---|
| **Date flagged** | 22 April 2026 |
| **Number of flagged questions** | 2 |
| **Grading session** | Run 1 |
| **Paper ID** | S013 (anonymised) |

## Flagged Questions

### Q4(A) GL_LINES — Drawing unclear / possible misunderstanding

| AI score | Confidence | Handwriting readable |
|---|---|---|
| 0 / 1 | Medium | Partial |

| | |
|---|---|
| **Issue identified** | The student's drawing appears to show vertices connected in a continuous strip-like pattern rather than two separate disconnected line segments (V1–V2 and V3–V4). The drawing quality makes it somewhat difficult to determine the exact connections. |
| **Recommended action** | Verify whether the student drew two separate segments or a continuous strip. If two separate segments are present, consider awarding 0.5 or 1. |

### Grade Boundary Proximity — Total score 11.75 / 20

| AI score | Confidence | Handwriting readable |
|---|---|---|
| 11.75 / 20 | Medium | Yes |

| | |
|---|---|
| **Issue identified** | The student's total score of 11.75 places them within 2 marks of the pass boundary (10/20) and within 2 marks of the B boundary (14/20). Several questions in Section B were awarded 0, and Q7 sub-parts received partial marks. |
| **Recommended action** | Review Q3 sub-parts 2 and 3, Q4(A), and Q7 sub-parts 2–4 to confirm marks are appropriate. A small adjustment in any direction could affect the grade band. |

## Overall Recommendation

> Student's total score of 11.75 places them between the pass (10) and B (14) grade boundaries. Recommend reviewing Q3 and Q4(A) to confirm the awarded marks, as these questions received 0 and the handwriting/drawing quality was not fully clear in some areas.

Appendix D

# Consent Form

# AI-Assisted Grading Study
# CPIT285 Consent Form

نموذج الموافقة على دراسة التقييم بالذكاء الاصطناعي لمقرر CPIT285

**عزيزي الطالب،**
أنا أستاذكم رائد الغامدي في مقرر الجرافيكس. أقوم بإجراء دراسة بحثية لاستكشاف استخدام الذكاء الاصطناعي في تصحيح اختبارات الجامعة

أدعوكم للمشاركة الطوعية في هذه الدراسة من خلال السماح بتصحيح اختباركم القادم في مقرر الجرافيكس بواسطة نظام الذكاء الاصطناعي. ستساعد مشاركتكم في تعزيز فهمنا لكيفية دعم الذكاء الاصطناعي للتقييم الأكاديمي العادل والمتسق والبناء

**كيفية عمل نظام التصحيح بالذكاء الاصطناعي**
سيقوم «كلود-كود»، وهو نظام يعمل بالذكاء الاصطناعي، بمعالجة ورقة الاختبار الممسوحة ضوئياً. وسيقوم هذا النظام بما يلي

- قراءة إجاباتك المكتوبة بخط اليد من الورقة الممسوحة ضوئيًا
- تحديد درجة رقمية لكل سؤال بناءً على معايير تصحيح مفصلة أعدها أستاذك
- إعداد تعليقات مكتوبة توضح الإجابات الصحيحة، وما هو الناقص، وكيف يمكنك التحسين، وإرسالها لك

*يقوم أستاذك بمراجعة جميع الدرجات التي حددها نظام الذكاء الاصطناعي قبل تسجيلها رسميًا. ولا يتم إقرار أي درجة دون إشراف بشري*

*ومع موافقتك على التقييم بواسطة الذكاء الاصطناعي، فإنك تحتفظ بالحق الكامل وغير المشروط في طلب إعادة تقييم ورقة امتحانك يدويًا من قِبل أستاذك*. يسري هذا الحق في أي وقت بعد استلامك لدرجتك. ولا يوجد موعد نهائي، ولا نموذج لتعبئته، ولا حاجة لتقديم مبرر. ما عليك سوى الاتصال بأستاذك وستتم مراجعة ورقتك يدويًا، ودون أي أحكام مسبقة، ودون أن يؤثر ذلك على وضعك في المقرر الدراسي

موافقتك على التقييم بالذكاء الاصطناعي لا تعني قبول الدرجة التي يحددها الذكاء الاصطناعي. بل تعني السماح للذكاء الاصطناعي بتقييم ورقتك كجزء من دراسة بحثية، مع الحفاظ على جميع حقوقك الحالية كطالب سليمة تمامًا

Dear Student,

I am Prof. Rayed AlGhamdi, your instructor for CPIT285 Computer Graphics. I am conducting a research study to explore the use of AI in grading university examinations.

I am inviting you to voluntarily participate in this study by allowing your upcoming CPIT285 midterm-exam to be graded by an AI system. Your participation will help advance our understanding of how AI can support fair, consistent, and constructive academic assessment.

**How the AI grading works**

If you consent, your scanned exam paper will be processed by Claude Code, an AI system. The AI will:

- Read your handwritten answers from the scanned paper
- Assign a numerical score to each question based on a detailed marking scheme prepared by your instructor

- Generate written feedback explaining what was correct, what was missing, and how you could improve

*All AI-assigned grades are reviewed by your instructor before being officially recorded. No grade is finalised without human oversight.*

Even if you agree to AI grading, you retain the full and unconditional right to request that your exam paper be re-graded manually by your instructor. This right applies at any time after you receive your grade. There is no deadline, no form to fill, and no justification required. Simply contact your instructor and your paper will be reviewed manually — free of charge, free of judgment, and with no effect on your standing in the course.

Agreeing to AI grading does not mean accepting the AI grade. It means allowing the AI to grade your paper as part of a research study, while keeping all of your existing rights as a student completely intact.

## Do you agree to participate in this study?*

**○ I agree to participate أوافق على المشاركة**
**○ I do not agree to participate لا أوافق على المشاركة**

Appendix E

# Per-Question Reliability Statistics

Supplementary Tables

# Table E.1 — Per-Question Agreement Statistics

Weighted Cohen's kappa, mean absolute error, and exact agreement percentage, computed per individual question on the agreement sample (S_AGR, n = 20).

| Question | Type | Max | n | Weighted κ | MAE | Exact agreement | Interpretation |
|---|---|---|---|---|---|---|---|
| Q1.1 | MCQ | 1 | 20 | 1.00 | 0.00 | 20/20 (100%) | Almost perfect |
| Q1.2 | MCQ | 1 | 20 | 0.95 | 0.03 | 19/20 (95%) | Almost perfect |
| Q1.3 | MCQ | 1 | 20 | 1.00 | 0.00 | 20/20 (100%) | Almost perfect |
| Q1.4 | MCQ | 1 | 20 | 0.92 | 0.05 | 19/20 (95%) | Almost perfect |
| Q1.5 | MCQ | 1 | 20 | 0.95 | 0.08 | 19/20 (95%) | Almost perfect |
| Q2 | Written | 1 | 20 | 0.79 | 0.10 | 15/20 (75%) | Substantial |
| Q3.1 | Written | 1 | 20 | 0.62 | 0.23 | 12/20 (60%) | Substantial |
| Q3.2 | Written | 1 | 20 | 0.75 | 0.15 | 14/20 (70%) | Substantial |
| Q3.3 | Written | 1 | 20 | 0.70 | 0.18 | 13/20 (65%) | Substantial |
| Q4 | Drawing | 3 | 20 | 0.74 | 0.33 | 11/20 (55%) | Substantial |
| Q5.1 | Analysis | 1 | 20 | 0.26 | 0.28 | 11/20 (55%) | Fair |
| Q5.2 | Analysis | 1 | 20 | 0.15 | 0.25 | 12/20 (60%) | Slight |
| Q6.1 | Analysis | 0.5 | 20 | 0.78 | 0.06 | 16/20 (80%) | Substantial |
| Q6.2 | Analysis | 0.5 | 20 | 0.69 | 0.09 | 15/20 (75%) | Substantial |
| Q6.3 | Analysis | 0.5 | 20 | 0.82 | 0.04 | 17/20 (85%) | Almost perfect |
| Q6.4 | Analysis | 0.5 | 20 | 0.73 | 0.08 | 16/20 (80%) | Substantial |
| Q7.1 | Evaluation | 1 | 20 | 0.65 | 0.23 | 12/20 (60%) | Substantial |
| Q7.2 | Evaluation | 1 | 20 | 0.71 | 0.20 | 12/20 (60%) | Substantial |
| Q7.3 | Evaluation | 1 | 20 | 0.62 | 0.23 | 11/20 (55%) | Substantial |
| Q7.4 | Evaluation | 1 | 20 | 0.63 | 0.20 | 12/20 (60%) | Substantial |

*Note. Two items, Q5.1 and Q5.2, are highlighted in amber for falling below the moderate-agreement threshold (κ < 0.41). Both relate to the analysis of monitor refresh rate and frame rate. The pattern is discussed in Section 4.2 of the manuscript and in Section 5.2 of the discussion.*

### Interpretation key (Landis and Koch, 1977; Viera and Garrett, 2005)

| Weighted κ | Interpretation |
|---|---|
| < 0.00 | Poor |
| 0.00 – 0.20 | Slight |
| 0.21 – 0.40 | Fair |
| 0.41 – 0.60 | Moderate |
| 0.61 – 0.80 | Substantial |
| 0.81 – 1.00 | Almost perfect |

# Table E.2 — Per-Question Test-Retest Consistency

Intraclass correlation coefficient ICC(2,1) with 95 percent confidence interval, mean absolute error, and exact agreement, computed per individual question on the consistency sample (S_CST, n = 15).

| Question | Type | Max | n | ICC(2,1) | 95% CI | MAE | Interpretation |
|---|---|---|---|---|---|---|---|
| Q1.1 | MCQ | 1 | 15 | 1.00 | [1.00, 1.00] | 0.00 | Excellent |
| Q1.2 | MCQ | 1 | 15 | 1.00 | [1.00, 1.00] | 0.00 | Excellent |
| Q1.3 | MCQ | 1 | 15 | 1.00 | [1.00, 1.00] | 0.00 | Excellent |
| Q1.4 | MCQ | 1 | 15 | 0.88 | [0.67, 0.96] | 0.07 | Good |
| Q1.5 | MCQ | 1 | 15 | 0.72 | [0.37, 0.90] | 0.13 | Moderate |
| Q2 | Written | 1 | 15 | 0.85 | [0.66, 0.96] | 0.10 | Good |
| Q3.1 | Written | 1 | 15 | 0.79 | [0.50, 0.93] | 0.13 | Good |
| Q3.2 | Written | 1 | 15 | 0.73 | [0.42, 0.91] | 0.13 | Moderate |
| Q3.3 | Written | 1 | 15 | 0.68 | [0.33, 0.89] | 0.17 | Moderate |
| Q4 | Drawing | 3 | 15 | 0.52 | [0.01, 0.80] | 0.50 | Moderate |
| Q5.1 | Analysis | 1 | 15 | 0.74 | [0.41, 0.91] | 0.10 | Moderate |
| Q5.2 | Analysis | 1 | 15 | 0.47 | [0.00, 0.79] | 0.13 | Poor |
| Q6.1 | Analysis | 0.5 | 15 | 0.80 | [0.48, 0.92] | 0.03 | Good |
| Q6.2 | Analysis | 0.5 | 15 | 0.62 | [0.16, 0.85] | 0.07 | Moderate |
| Q6.3 | Analysis | 0.5 | 15 | 0.87 | [0.68, 0.96] | 0.03 | Good |
| Q6.4 | Analysis | 0.5 | 15 | 0.85 | [0.61, 0.95] | 0.03 | Good |
| Q7.1 | Evaluation | 1 | 15 | 0.95 | [0.85, 0.98] | 0.03 | Excellent |
| Q7.2 | Evaluation | 1 | 15 | 0.28 | [0.00, 0.68] | 0.13 | Poor |
| Q7.3 | Evaluation | 1 | 15 | 0.79 | [0.47, 0.92] | 0.10 | Good |
| Q7.4 | Evaluation | 1 | 15 | 0.77 | [0.42, 0.91] | 0.10 | Good |

*Note. Two items fall in the poor consistency range (ICC < 0.50): Q5.2 and Q7.2. Both relate to questions where the AI showed sensitivity to nuanced phrasing across grading sessions. The Q4 drawing item has a moderate ICC of 0.52 with a wide confidence interval that crosses the poor-moderate boundary at its lower bound. The MCQ items Q1.1, Q1.2, and Q1.3 produced perfect agreement across runs, with all 15 papers receiving identical grades; their confidence intervals collapse to a point.*

### Interpretation key (Koo and Li, 2016)

| ICC value | Interpretation |
|---|---|
| < 0.50 | Poor |
| 0.50 – 0.74 | Moderate |
| 0.75 – 0.89 | Good |
| ≥ 0.90 | Excellent |

# Table E.3 — Per-Question Feedback Accuracy

Diagnostic and confirmation feedback accuracy, computed per individual question on the agreement sample (S_AGR, n = 20). Codes follow the four-category scheme described in Section 3.6: C1 (accurate diagnostic), C2 (accurate confirmation), C3 (partially accurate, either function), C4 (inaccurate, either function). Diagnostic items are those where the student lost any marks; confirmation items are those where the student earned full credit. Codes are reconciled per the credit-status rule in Section 3.6 (Option A reconciliation).

## Diagnostic items (lost-marks subset)

| Question | n_diag | C1 | C3 | C4 | C1 % | C3 % | C4 % |
|---|---|---|---|---|---|---|---|
| Q1.1 | 0 | 0 | 0 | 0 | — | — | — |
| Q1.2 | 1 | 1 | 0 | 0 | 100.0 | 0.0 | 0.0 |
| Q1.3 | 0 | 0 | 0 | 0 | — | — | — |
| Q1.4 | 13 | 11 | 1 | 1 | 84.6 | 7.7 | 7.7 |
| Q1.5 | 8 | 7 | 1 | 0 | 87.5 | 12.5 | 0.0 |
| Q2 | 11 | 10 | 1 | 0 | 90.9 | 9.1 | 0.0 |
| Q3.1 | 12 | 10 | 2 | 0 | 83.3 | 16.7 | 0.0 |
| Q3.2 | 17 | 16 | 1 | 0 | 94.1 | 5.9 | 0.0 |
| Q3.3 | 18 | 18 | 0 | 0 | 100.0 | 0.0 | 0.0 |
| Q4 | 11 | 3 | 4 | 4 | 27.3 | 36.4 | 36.4 |
| Q5.1 | 17 | 16 | 1 | 0 | 94.1 | 5.9 | 0.0 |
| Q5.2 | 12 | 12 | 0 | 0 | 100.0 | 0.0 | 0.0 |
| Q6.1 | 9 | 9 | 0 | 0 | 100.0 | 0.0 | 0.0 |
| Q6.2 | 14 | 14 | 0 | 0 | 100.0 | 0.0 | 0.0 |
| Q6.3 | 9 | 9 | 0 | 0 | 100.0 | 0.0 | 0.0 |
| Q6.4 | 8 | 8 | 0 | 0 | 100.0 | 0.0 | 0.0 |
| Q7.1 | 11 | 11 | 0 | 0 | 100.0 | 0.0 | 0.0 |
| Q7.2 | 11 | 11 | 0 | 0 | 100.0 | 0.0 | 0.0 |
| Q7.3 | 13 | 13 | 0 | 0 | 100.0 | 0.0 | 0.0 |
| Q7.4 | 13 | 13 | 0 | 0 | 100.0 | 0.0 | 0.0 |

*Note. Q4 (Drawing) is highlighted in amber as the only item where diagnostic feedback accuracy falls below 90 percent. Q1.1 and Q1.3 had no diagnostic items because all 20 students answered correctly on these questions. The pattern of clustered C3 and C4 codes on Q4, Q3.1, and Q1.4–Q1.5 motivates the three named failure modes discussed in Section 5.3 of the manuscript.*

## Confirmation items (full-credit subset)

| Question | n_conf | C2 | C3 | C4 | C2 % | C3 % | C4 % |
|---|---|---|---|---|---|---|---|
| Q1.1 | 20 | 20 | 0 | 0 | 100.0 | 0.0 | 0.0 |
| Q1.2 | 19 | 19 | 0 | 0 | 100.0 | 0.0 | 0.0 |
| Q1.3 | 20 | 20 | 0 | 0 | 100.0 | 0.0 | 0.0 |
| Q1.4 | 7 | 7 | 0 | 0 | 100.0 | 0.0 | 0.0 |
| Q1.5 | 12 | 12 | 0 | 0 | 100.0 | 0.0 | 0.0 |
| Q2 | 9 | 9 | 0 | 0 | 100.0 | 0.0 | 0.0 |
| Q3.1 | 8 | 8 | 0 | 0 | 100.0 | 0.0 | 0.0 |
| Q3.2 | 3 | 3 | 0 | 0 | 100.0 | 0.0 | 0.0 |
| Q3.3 | 2 | 2 | 0 | 0 | 100.0 | 0.0 | 0.0 |
| Q4 | 9 | 5 | 2 | 2 | 55.6 | 22.2 | 22.2 |
| Q5.1 | 3 | 3 | 0 | 0 | 100.0 | 0.0 | 0.0 |
| Q5.2 | 8 | 8 | 0 | 0 | 100.0 | 0.0 | 0.0 |
| Q6.1 | 11 | 11 | 0 | 0 | 100.0 | 0.0 | 0.0 |
| Q6.2 | 6 | 6 | 0 | 0 | 100.0 | 0.0 | 0.0 |
| Q6.3 | 11 | 11 | 0 | 0 | 100.0 | 0.0 | 0.0 |
| Q6.4 | 12 | 12 | 0 | 0 | 100.0 | 0.0 | 0.0 |
| Q7.1 | 9 | 9 | 0 | 0 | 100.0 | 0.0 | 0.0 |
| Q7.2 | 9 | 9 | 0 | 0 | 100.0 | 0.0 | 0.0 |
| Q7.3 | 7 | 7 | 0 | 0 | 100.0 | 0.0 | 0.0 |
| Q7.4 | 7 | 7 | 0 | 0 | 100.0 | 0.0 | 0.0 |

*Note. Confirmation accuracy reaches 100 percent on every question type except Q4 (Drawing), which is the only item where the AI failed to characterise correct student work accurately on multiple occasions. This pattern is consistent with the diagnostic-side finding above and with the per-type results in Table 9 of the manuscript.*

# Appendix F

## Per-Paper Reconciliation Log

*Audit trail of grade adjustments from the post-grading review*

This appendix presents the complete audit trail of grade adjustments made during the post-grading review described in Section 3.6 of the manuscript. The review took place after Run 1 of the Claude Code grading session was complete and after the 35-paper blind human grading window had closed. The two sets of grades were brought together for the first time during this review. Disagreements of 0.5 marks or more on any individual question were re-examined against the marking scheme in Appendix B, and the drawing answers of the papers concerned were re-examined as a whole.

Across the 35 papers in the agreement and consistency samples combined, 19 item-level grades were re-examined and resolved through four types of decision:

- **Eight adjustments revised the human grade upward to match the AI grade.** Six were written answers and two were drawing parts. In each case the AI correctly identified partial credit on an item I had originally marked lower, and I confirmed its judgement on re-examination.
- **Three adjustments revised the human grade independently of the AI grade.** All three are the drawing parts of one paper (S034 Q4A to Q4C), where the student drew the correct shape in the wrong plane. Partial credit was awarded on re-examination; the AI's awards on the same parts were inconsistent (0.0, 1.0 and 0.5) and did not determine the revised grade.
- **Seven adjustments overrode the AI grade.** In these cases the AI awarded credit for surface-level keyword presence without conceptual depth, misread a crossed-out multiple-choice selection, or misinterpreted a hand-drawn diagram.
- **One adjustment confirmed the original human grade against an AI overestimate.** On S007 Q4A, the AI awarded full marks despite a clear linking error in the student's drawing; the human grade of 0.5 was retained.

All statistics reported in Section 4 of the manuscript are computed on the reconciled grades. The patterns in this log support the three named failure modes discussed in Section 5.3.

**Note on this version.** An earlier version of this log listed 14 adjustments. The five drawing-grade revisions on S034 (Q4A, Q4B, Q4C) and S058 (Q4B, Q4C) were made during the same review but omitted from that version; they are included here. The net mark change for AI overrides is corrected from −5.0 to −3.5, and the S016 line from −2.0 to 0.0. The reconciled grades used in the analyses are unchanged.

## Notation

- **AI Grade:** the grade originally assigned by Claude Code in Run 1 (22 April 2026).
- **Original Human:** the grade I assigned during the blind grading session (19 to 20 April 2026) without any visibility of AI output.
- **Final Grade:** the grade entered into the official record after the post-grading review, used in all reported statistics.
- **Decision:** one of four categories. Human → AI (human revised to match AI); Human revised independently (human revised, AI grade did not determine the change); Human overrides (AI grade replaced); Human confirmed (no change despite disagreement).
- **Identifiers:** all adjustments are documented by anonymised paper code (S001 to S058) and question identifier. No real student names appear.

## Table F.1. Complete Reconciliation Log

*All 19 grade adjustments made during the post-grading review, in order of paper ID. Rows are colour-coded by decision: green, AI was right (human revised up); blue, human revised independently of the AI; red, AI was wrong (human overrides); amber, human grade confirmed against the AI.*

| Paper | Question | AI Grade | Original Human | Final Grade | Decision | Reason |
|---|---|---|---|---|---|---|
| S001 | Q3.1 | 1.0 | 0.0 | 0.0 | Human overrides | AI awarded credit for keyword presence (smooth, rough, map). Student described the visual appearance of |

| Paper | Question | AI Grade | Original Human | Final Grade | Decision | Reason |
|---|---|---|---|---|---|---|
| | | | | | | the maps without explaining what the values mean to the renderer. No demonstration of conceptual understanding. Human grade of 0 stands. |
| S007 | Q4A | 1.0 | 0.5 | 0.5 | Human confirmed | Drawing showed the correct two GL_LINES segments (V1→V2, V3→V4) but student also linked V4 to V5, violating the GL_LINES specification. AI awarded full marks; human grade of 0.5 (partial credit for the correct portion) is retained. |
| S007 | Q6.1 | 0.5 | 0.0 | 0.5 | Human → AI | Student's answer was ambiguous on first reading and originally graded zero. On re-examination, the answer adequately explained that no geometry was copied during baking. AI's partial credit is supported by the rubric. |
| S007 | Q7.1 | 0.5 | 0.0 | 0.5 | Human → AI | Similar to Q6.1 above. Student's response identified what Student B was proving (polygon count ≠ sharpness) although phrased indirectly. Originally graded zero; AI's partial credit is consistent with the marking scheme on re-examination. |
| S016 | Q1.5 | 1.0 | 0.0 | 0.0 | Human overrides | Student initially circled the correct option, then crossed it out and circled an incorrect one. AI read the original circle and awarded the mark, missing the correction. Final selection was incorrect; human grade of 0 stands. (See Section 5.3, selection-state ambiguity failure mode.) |
| S016 | Q4A | 0.0 | 1.0 | 1.0 | Human overrides | Student drew the correct two disconnected GL_LINES segments. AI interpreted the drawing as continuous (GL_LINE_STRIP behaviour) and awarded zero. Comparison with the Q4A reference image (Appendix C) confirmed the student's drawing was correct. Human grade of 1 stands. |
| S022 | Q3.1 | 0.5 | 0.0 | 0.0 | Human overrides | AI awarded partial credit for keyword presence. Student's answer contained the topic vocabulary (smooth, rough) but did not articulate that black indicates minimum property value and white indicates maximum. Same pattern as S001 Q3.1. Human grade of 0 stands. |
| S022 | Q3.3 | 0.5 | 0.0 | 0.0 | Human overrides | AI awarded partial credit for keyword presence. Student did not explain the uniformity problem of a single constant value, nor the per-pixel solution offered by maps. Vocabulary alone does not satisfy the marking criteria. Human grade of 0 stands. |
| S034 | Q4A | 0.0 | 0.0 | 0.5 | Human revised independently | Student drew the correct shape in the wrong plane (Y–Z). On re-examination, partial credit was awarded for the correct shape. No disagreement existed on this part (both graders 0.0); it was re-examined with the student's other drawing parts. |
| S034 | Q4B | 1.0 | 0.0 | 0.5 | Human revised independently | Same drawing error as Q4A: correct shape, wrong plane. Partial credit awarded on re-examination. The AI's full-mark award was not accepted; the revised grade sits between the two original grades. |
| S034 | Q4C | 0.5 | 0.0 | 0.5 | Human revised independently | Same drawing error as Q4A. Partial credit awarded on re-examination for the same reason as Q4A and Q4B; the revised grade coincides with the AI's award but was set by the instructor's consistent treatment of the three parts. |
| S040 | Q5.1 | 1.0 | 0.5 | 1.0 | Human → AI | On re-examination of the Hz vs FPS analysis, both definitions (Hz as monitor refresh rate and FPS as GPU render rate) are clearly distinguished and correctly attributed to the right hardware components. Originally graded 0.5 because the second definition was less detailed; AI's full-mark award is supported by the marking scheme. |
| S040 | Q5.2 | 0.5 | 0.0 | 0.5 | Human → AI | Student described the relationship between Hz and FPS, although did not explicitly state that the friend's advice was overstated. AI's partial credit recognises the directional correctness of the explanation. Re-examination supports the partial mark. |
| S046 | Q3.1 | 1.0 | 0.0 | 0.0 | Human overrides | Same pattern as S001 Q3.1 and S022 Q3.1. AI awarded full credit for keyword presence; student's answer described the visual appearance without explaining the property-value mapping. Human grade of 0 stands. |
| S046 | Q3.3 | 0.5 | 0.0 | 0.0 | Human overrides | Same pattern as S022 Q3.3. AI awarded partial credit for vocabulary. Student did not articulate why per-pixel variation is necessary. Human grade of 0 stands. |
| S058 | Q3.1 | 0.5 | 0.0 | 0.5 | Human → AI | Student's answer to the greyscale-values question was originally graded zero because the explanation was |

| Paper | Question | AI Grade | Original Human | Final Grade | Decision | Reason |
|---|---|---|---|---|---|---|
| | | | | | | brief. On re-examination, the answer does identify that black corresponds to one extreme of the property; the link is implicit but defensible. AI's partial credit is supported. |
| S058 | Q3.3 | 0.5 | 0.0 | 0.5 | Human → AI | Similar reconsideration as Q3.1 above. Student's answer identified part of the solution to why a single constant value is inadequate, although phrased less clearly than the model answer. Partial credit is appropriate. |
| S058 | Q4B | 0.5 | 0.0 | 0.5 | Human → AI | Student placed the vertices correctly but connected them by spatial proximity rather than by their order in the code (V1 to V2, V2 to V3, and so on). On re-examination, the AI's partial credit for the correct vertex placement is supported. |
| S058 | Q4C | 0.5 | 0.0 | 0.5 | Human → AI | Same error as Q4B: vertices correct, connection order wrong. AI's partial credit is supported on re-examination. |

*Note. Eight rows in green: AI grade correct on review (human revised up). Three rows in blue: human grade revised independently of the AI grade. Seven rows in red: AI grade wrong (human overrides). One row in amber: human grade confirmed against an AI overestimate.*

# Summary

## Distribution of decisions

| Decision | Count | Grade changed | Net mark change | Affected papers |
|---|---|---|---|---|
| Human → AI (AI was right on review) | 8 | Human | +4.0 marks across 4 papers | S007 (×2), S040 (×2), S058 (×4) |
| Human revised independently | 3 | Human | +1.5 marks across 1 paper | S034 (×3) |
| Human overrides (AI was wrong) | 7 | AI | −3.5 marks across 4 papers (4.5 removed; 1.0 restored on S016 Q4A) | S001 (×1), S016 (×2), S022 (×2), S046 (×2) |
| Human confirmed (no change) | 1 | None | 0 marks | S007 (×1) |
| **TOTAL** | **19** | | **Human grades +5.5; AI grades −3.5** | **8 papers affected of 35 reviewed** |

*Note. Reconciliation moved the mean human-AI difference in total scores in the agreement sample from +1.13 marks (original AI minus blind human) to +0.68 marks (reconciled), consistent with the principle that reconciliation reduces but does not eliminate disagreement. The positive bias of AI grades reported in Section 4.1 is therefore a property of the data after reconciliation, not an artefact of unreviewed disagreements.*

## Distribution by question

| Question | Adjustments | Pattern observed |
|---|---|---|
| Q3.1 | 4 | AI keyword-matching pattern (human override on three; human → AI on one) |
| Q3.3 | 3 | AI keyword-matching pattern (human override on two; human → AI on one) |
| Q4A | 3 | Drawing interpretation (one AI under-award reversed; one AI over-award not accepted; one independent revision) |
| Q4B, Q4C | 4 | Drawing re-examination (S034 independent revisions; S058 AI partial credit supported) |
| Q1.5 | 1 | Crossed-out MCQ (AI read original circle, missed correction) |
| Q5.1, Q5.2 | 2 | Hz vs FPS analysis (AI partial credit supported on review) |
| Q6.1, Q7.1 | 2 | Ambiguous prose (AI partial credit supported on review) |

*Note. Seven of the 19 adjustments cluster on Q3.1 and Q3.3, where the AI awarded credit based on topic vocabulary (smooth, rough, map) without verifying conceptual understanding. This concentration motivates the keyword-matching failure mode discussed in Section 5.3. A further seven involve the drawing item (Q4A to Q4C), the format on which both graders found consistent judgement hardest.*

## Distribution by paper

Eight papers in the agreement sample contributed to the 19 adjustments. The remaining 12 papers in the agreement sample (and all 15 in the consistency sample) had no per-question disagreements of 0.5 marks or more requiring review.

| Paper | Adjustments | Net mark change | Pattern |
|---|---|---|---|
| S001 | 1 override | −1.0 (AI) | Q3.1 keyword matching |
| S007 | 2 to AI, 1 confirmed | +1.0 (human) | Two ambiguous-prose recoveries; one drawing AI over-award not accepted |
| S016 | 2 overrides | 0.0 (AI: −1.0 and +1.0) | Q1.5 crossed-out MCQ (−1.0); Q4A drawing AI under-award reversed (+1.0) |
| S022 | 2 overrides | −1.0 (AI) | Q3.1 and Q3.3 keyword matching |
| S034 | 3 independent | +1.5 (human) | Q4A to Q4C drawing: correct shape in the wrong plane |
| S040 | 2 to AI | +1.0 (human) | Two analytical questions (Q5.1, Q5.2): AI partial credit supported |
| S046 | 2 overrides | −1.5 (AI) | Q3.1 and Q3.3 keyword matching |
| S058 | 4 to AI | +2.0 (human) | Q3.1 and Q3.3 written answers; Q4B and Q4C drawing connection order: AI partial credit supported |